\documentclass[conference]{IEEEtran}

\usepackage[cmex10]{amsmath}
\usepackage{amsthm}
\usepackage{amssymb}
\usepackage{mathrsfs}
\usepackage{graphicx}
\usepackage{float}
\usepackage{array}
\usepackage{epstopdf}
\usepackage{multirow}
\usepackage{hyperref}
\usepackage{lscape}
\usepackage{longtable}
\usepackage{subcaption}
\usepackage{calrsfs}
\usepackage{booktabs}
\usepackage[table]{xcolor}
\usepackage{xcolor,colortbl}

\begin{document}

\title{Large Language Models: A Survey}

\author{Shervin Minaee$^{1}$, Tomas Mikolov$^{2}$, Narjes Nikzad$^{3}$, Meysam Chenaghlu$^{4}$ \\ 
Richard Socher$^{5}$, Xavier Amatriain$^{6}$, Jianfeng Gao$^{7}$ \\ \\
$^{1}$ Applied Scientist, Amazon Inc\\
$^{2}$ Senior Researcher, CIIRC CTU\\
$^{3}$ Cologne University of Applied Sciences\\
$^{4}$ Staff Machine Learning Scientist, Ultimate.ai\\
$^{5}$ CEO, You.com\\
$^{6}$ VP of Product, AI and Compute Enablement, Google Inc\\
$^{7}$ VP of Deep Learning Group, Microsoft Research\\
}

\maketitle

\begin{abstract}
Large Language Models (LLMs) have drawn a lot of attention due to their strong performance on a wide range of natural language tasks, since the release of ChatGPT in November 2022. 
LLMs' ability of general-purpose language understanding and generation is acquired by training billions of model's parameters on massive amounts of text data, as predicted by scaling laws \cite{kaplan2020scaling,hoffmann2022training}.
The research area of LLMs, while very recent, is evolving rapidly in many different ways.
In this paper, we review some of the most prominent LLMs, including three popular LLM families (GPT, LLaMA, PaLM), and discuss their characteristics, contributions and limitations. 
We also give an overview of techniques developed to build, and augment LLMs.
We then survey popular datasets prepared for LLM training, fine-tuning, and evaluation, review widely used LLM evaluation metrics, and compare the performance of several popular LLMs on a set of representative benchmarks. 
Finally, we conclude the paper by discussing open challenges and future research directions.
\end{abstract}

\section{Introduction}
Language modeling is a long-standing research topic, dating back to the 1950s with Shannon's application of information theory to human language, where he measured how well simple n-gram language models predict or compress natural language text \cite{shannon1951prediction}. 
Since then, statistical language modeling became fundamental to many natural language understanding and generation tasks, ranging from speech recognition, machine translation, to information retrieval \cite{jelinek1998statistical,manning1999foundations,manning2009introduction}.

The recent advances on transformer-based large language models (LMs), pretrained on Web-scale text corpora, significantly extended the capabilities of language models (LLMs). For example, OpenAI's ChatGPT and GPT-4 can be used not only for natural language processing, but also as general task solvers to power Microsoft's Co-Pilot systems, for instance, can follow human instructions of complex new tasks performing multi-step reasoning when needed. LLMs are thus becoming the basic building block for the development of general-purpose AI agents or artificial general intelligence (AGI).

As the field of LLMs is moving fast, with new findings, models and techniques being published in a matter of months or weeks \cite{zhao2023survey,zhou2023comprehensive,liu2023pre,dong2022survey,huang2022towards}, AI researchers and practitioners often find it challenging to figure out the best recipes to build LLM-powered AI systems for their tasks. This paper gives a timely survey of the recent advances on LLMs. 
We hope this survey will prove a valuable and accessible resource for students, researchers and developers.

LLMs are large-scale, pre-trained, statistical language models based on neural networks. The recent success of LLMs is an accumulation of decades of research and development of language models, which can be categorized into four waves that have different starting points and velocity: statistical language models, neural language models, pre-trained language models and LLMs.

Statistical language models (SLMs) view text as a sequence of words, and estimate the probability of text as the product of their word probabilities. The dominating form of SLMs are Markov chain models known as the n-gram models, which compute the probability of a word conditioned on its immediate proceeding $n-1$ words. Since word probabilities are estimated using word and n-gram counts collected from text corpora, the model needs to deal with data sparsity (i.e., assigning zero probabilities to unseen words or n-grams) by using \emph{smoothing}, where some probability mass of the model is reserved for unseen n-grams \cite{chen1999empirical}. N-gram models are widely used in many NLP systems. However, these models are incomplete in that they cannot fully capture the diversity and variability of natural language due to data sparsity.  

Early neural language models (NLMs) \cite{bengio2000neural,schwenk2006continuous,mikolov2010recurrent,graves2013generating} deal with data sparsity by mapping words to low-dimensional continuous vectors (embedding vectors) and predict the next word based on the aggregation of the embedding vectors of its proceeding words using neural networks. The embedding vectors learned by NLMs define a hidden space where the semantic similarity between vectors can be readily computed as their distance. 
This opens the door to computing semantic similarity of any two inputs regardless their forms (e.g., queries vs. documents in Web search \cite{huang2013learning, gao2023neural}, sentences in different languages in machine translation \cite{sutskever2014sequence, cho2014properties}) or modalities (e.g., image and text in image captioning \cite{fang2015captions, vinyals2015show}). Early NLMs are task-specific models, in that they are trained on task-specific data and their learned hidden space is task-specific.



Pre-trained language models (PLMs), unlike early NLMs, are task-agnostic. This generality also extends to the learned hidden embedding space. The training and inference of PLMs follows the \emph{pre-training and fine-tuning} paradigm, where language models with recurrent neural networks \cite{peters2018deep} or transformers \cite{devlin2018bert,liu2019roberta,he2020deberta} are pre-trained on Web-scale unlabeled text corpora for general tasks such as word prediction, and then finetuned to specific tasks using small amounts of (labeled) task-specific data. Recent surveys on PLMs include \cite{zhou2023comprehensive, han2021pre,qiu2020pre}. 

\begin{figure*}
\begin{center}
    \includegraphics [scale=0.57] {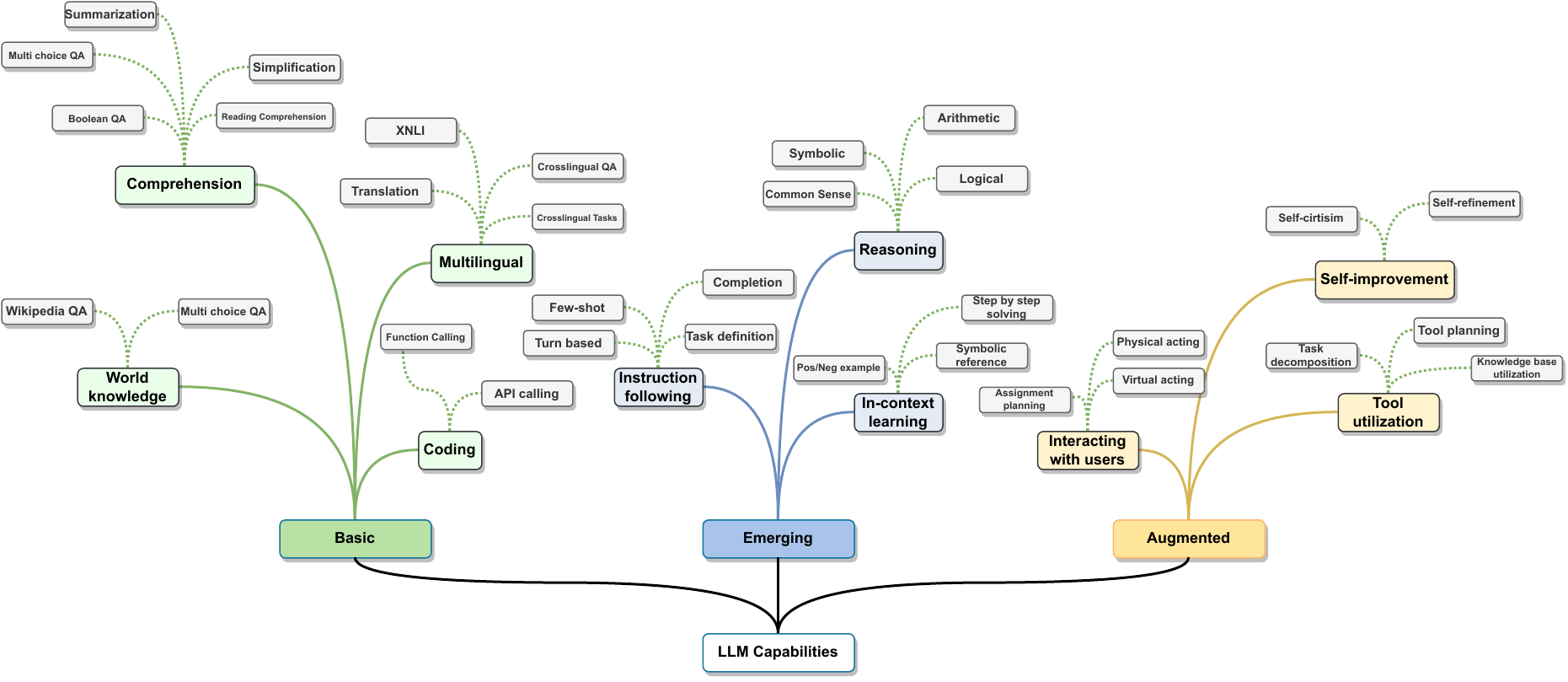}
\end{center}
  \caption{LLM Capabilities.}
  \label{fig:llm_cap}
\end{figure*}

Large language models mainly refer to transformer-based neural language models \footnote{Recently, several very promising non-transformer LLMs have been proposed, such as the LLMs based on structured state space models \cite{gu2022S4,gu2023mamba}. See Section \ref{sec:LLM_challenges} for more details.} that contain tens to hundreds of billions of parameters, which are pre-trained on massive text data, such as PaLM \cite{chowdhery2022palm}, LLaMA \cite{touvron2023llama}, and GPT-4 \cite{gpt4}, as summarized in Table III.
Compared to PLMs, LLMs are not only much larger in model size, but also exhibit stronger language understanding and generation abilities, and more importantly, emergent abilities that are not present in smaller-scale language models. As illustrated in Fig.~\ref{fig:llm_cap}, these emergent abilities include (1) in-context learning, where LLMs learn a new task from a small set of examples presented in the prompt at inference time, (2) instruction following, where LLMs, after instruction tuning, can follow the instructions for new types of tasks without using explicit examples, and (3) multi-step reasoning, where LLMs can solve a complex task by breaking down that task into intermediate reasoning steps as demonstrated in the chain-of-thought prompt \cite{Wei2022COT}. 
LLMs can also be augmented by using external knowledge and tools \cite{mialon2023augmented,peng2023check} so that they can effectively interact with users and environment \cite{yao2022react}, and continually improve itself using feedback data collected through interactions (e.g. via reinforcement learning with human feedback (RLHF)).

Through advanced usage and augmentation techniques, LLMs can be deployed as so-called AI agents: artificial entities that sense their environment, make decisions, and take actions. Previous research has focused on developing agents for specific tasks and domains. The emergent abilities demonstrated by LLMs make it possible to build general-purpose AI agents based on LLMs. 
While LLMs are trained to produce responses in static settings, AI agents need to take actions to interact with dynamic environment. Therefore, LLM-based agents often need to augment LLMs to e.g., obtain updated information from external knowledge bases, verify whether a system action produces the expected result, and cope with when things do not go as expected, etc. We will discuss in detail LLM-based agents in Section \ref{sec:LLM_used}.

In the rest of this paper,
Section \ref{sec:LLM_models} presents an overview of state of the art of LLMs, focusing on three LLM families (GPT, LLaMA and PaLM) and other representative models. 
Section \ref{sec:LLM_built} discusses how LLMs are built.
Section \ref{sec:LLM_used} discusses how LLMs are used, and augmented for real-world applications
Sections \ref{sec:llm_datasets} and \ref{sec:llm_performance} review popular datasets and benchmarks for evaluating LLMs, and summarize the reported LLM evaluation results.
Finally, Section \ref{sec:LLM_challenges} concludes the paper  by summarizing the challenges and future research directions. 

\begin{figure*}[h]
\begin{center}
    \includegraphics [scale=0.52] {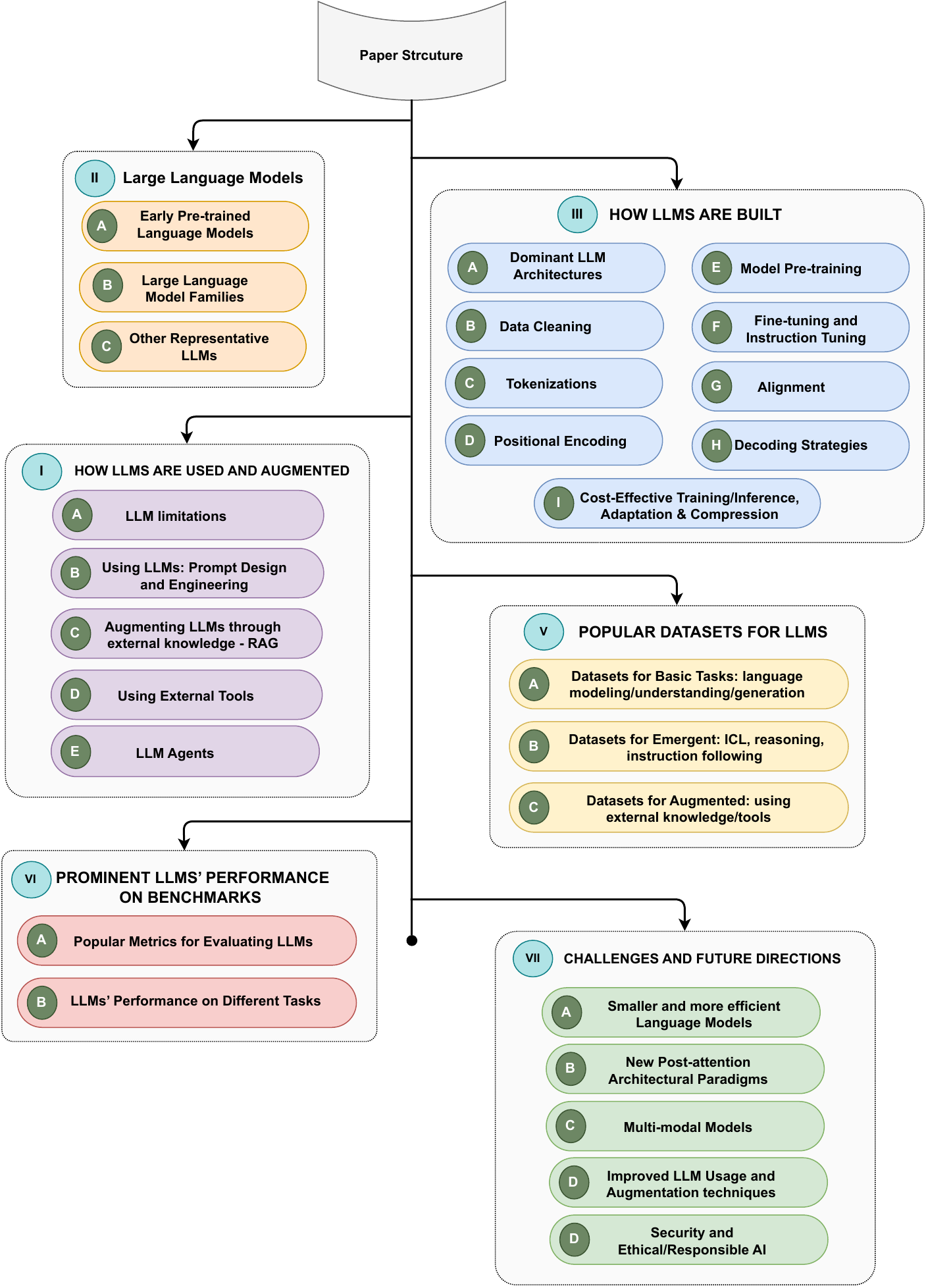}
\end{center}
  \caption{The paper structure.}
\label{fig:paper_structure}
\end{figure*}

\section{Large Language Models} 
\label{sec:LLM_models}

%

In this section we start with a review of early pre-trained neural language models as they are the base of LLMs, and then focus our discussion on three families of LLMs: GPT, LlaMA, and PaLM. 
Table \ref{tab:language_models} provides an overview of some of these models and their characteristics.

\begin{table*}
  \centering
  \caption{High-level Overview of Popular Language Models}
  \label{tab:language_models}
  \begin{tabular}{p{2.4cm}p{1.7cm}p{2cm}p{1cm}p{1.5cm}p{0.8cm}p{1cm}p{5cm}}
    \toprule
   \textbf{Type} &  \textbf{Model Name} & \textbf{\#Parameters} & \textbf{Release} & \textbf{Base Models} & \textbf{Open Source} & \textbf{\#Tokens} & \textbf{Training dataset} \\
    \midrule
    & BERT & 110M, 340M & 2018 & - & \checkmark &  137B &  BooksCorpus, English Wikipedia \\
    & RoBERTa & 355M & 2019& -  & \checkmark & 2.2T & BooksCorpus, English Wikipedia, CC-NEWS, STORIES (a subset of Common Crawl), Reddit \\
   \multirow{2}{*}{\textbf{Encoder-Only}} & ALBERT & 12M, 18M, 60M, 235M & 2019 & -  & \checkmark & 137B &  BooksCorpus, English Wikipedia \\
    & DeBERTa &  - & 2020 & - & \checkmark & - & BooksCorpus, English Wikipedia, STORIES, Reddit content\\
    & XLNet & 110M, 340M & 2019 & - & \checkmark & 32.89B & BooksCorpus, English Wikipedia, Giga5, Common Crawl, ClueWeb 2012-B \\  
      \hline
  \multirow{2}{*}{\textbf{Decoder-only}} & GPT-1 & 120M & 2018 & - & \checkmark & 1.3B & BooksCorpus \\
      & GPT-2 & 1.5B & 2019 & - & \checkmark &  10B & Reddit outbound \\
      \hline
    & T5 (Base) & 223M & 2019 & - & \checkmark & 156B & Common Crawl \\
   \multirow{2}{*}{\textbf{Encoder-Decoder}}&  MT5 (Base) & 300M & 2020 & - & \checkmark & - & New Common Crawl-based dataset in 101 languages (m Common Crawl) \\
    & BART (Base) & 139M & 2019 & - & \checkmark &  - & Corrupting text  \\
 \hline
    & GPT-3 &  125M, 350M, 760M, 1.3B, 2.7B, 6.7B, 13B, 175B  & 2020 & & $\times$ &  300B& Common Crawl (filtered), WebText2, Books1, Books2, Wikipedia\\
    \multirow{2}{*}{\textbf{GPT Family}}& CODEX & 12B & 2021 & GPT & \checkmark  & - & Public GitHub software repositories  \\
     & WebGPT & 760M, 13B, 175B& 2021& GPT-3 & $\times$ & -& ELI5  \\
    & GPT-4 & 1.76T & 2023 & - & $\times$& 13T & - \\
 \hline
  &LLaMA1& 7B, 13B, 33B, 65B& 2023 & -& \checkmark & 1T, 1.4T& Online sources\\
   & LLaMA2 & 7B, 13B, 34B, 70B& 2023 & - & \checkmark & 2T &  Online sources\\
   &  Alpaca & 7B& 2023 & LLaMA1& \checkmark & - & GPT-3.5 \\
  & Vicuna-13B & 13B & 2023 & LLaMA1 & \checkmark & - & GPT-3.5 \\
  \multirow{2}{*}{\textbf{LLaMA Family}}  & Koala & 13B & 2023 & LLaMA & \checkmark & - & Dialogue data\\
    & Mistral-7B & 7.3B & 2023 & & \checkmark & - & - \\
    & Code Llama & 34 & 2023 & LLaMA2 & \checkmark & 500B & Publicly available code\\
    & LongLLaMA & 3B, 7B & 2023 & OpenLLaMA & \checkmark & 1T & -\\
    & LLaMA-Pro-8B & 8.3B & 2024 & LLaMA2-7B & \checkmark & 80B& Code and math corpora\\
    & TinyLlama-1.1B & 1.1B & 2024 & LLaMA1.1B  & \checkmark & 3T& SlimPajama, Starcoderdata\\
 \hline
  & PaLM & 8B, 62B, 540B & 2022 & - & $\times$ & 780B & Web documents, books, Wikipedia, conversations, GitHub code \\
    &U-PaLM & 8B, 62B, 540B & 2022 & - & $\times$ & 1.3B& Web documents, books, Wikipedia, conversations, GitHub code\\
 \multirow{2}{*}{\textbf{PaLM Family}}  & PaLM-2 & 340B & 2023 & - & \checkmark & 3.6T &  Web documents, books, code, mathematics, conversational data\\
  & Med-PaLM & 540B & 2022 & PaLM & $\times$ & 780B & HealthSearchQA, MedicationQA, LiveQA \\
 & Med-PaLM 2 & - & 2023 & PaLM 2 & $\times$ & - & MedQA, MedMCQA, HealthSearchQA, LiveQA, MedicationQA\\
 \hline
    & FLAN & 137B & 2021 & LaMDA-PT & \checkmark & - & Web documents, code, dialog data, Wikipedia\\
    & Gopher & 280B & 2021 & - & $\times$  & 300B & MassiveText\\
    & ERNIE 4.0 & 10B & 2023 & - & $\times$ & 4TB&  Chinese text\\
      &Retro & 7.5B & 2021 & - & $\times$ & 600B & MassiveText \\
    &LaMDA & 137B & 2022 & - & $\times$ & 168B & public dialog data and web documents \\
   &ChinChilla & 70B & 2022 & - & $\times$  & 1.4T & MassiveText \\
   & Galactia-120B & 120B & 2022 & - &  & 450B&  \\
   \multirow{2}{*}{\textbf{Other Popular LLMs}} & CodeGen & 16.1B & 2022 & - & \checkmark & - & THE PILE, BIGQUERY, BIGPYTHON\\
    & BLOOM & 176B & 2022 & - & \checkmark & 366B & ROOTS \\ 
    & Zephyr & 7.24B & 2023 & Mistral-7B & \checkmark & 800B &  Synthetic data \\
    &Grok-0 & 33B & 2023 & - & $\times$ & - & Online source \\
    &ORCA-2 & 13B & 2023 & LLaMA2 & - & 2001B& - \\
    & StartCoder & 15.5B & 2023 & - & \checkmark & 35B & GitHub \\
    &MPT & 7B & 2023 & - & \checkmark & 1T & RedPajama, m Common Crawl, S2ORC, Common Crawl \\
    & Mixtral-8x7B & 46.7B & 2023& - & \checkmark & - & Instruction dataset \\
   & Falcon 180B & 180B & 2023 & - & \checkmark & 3.5T &  RefinedWeb\\
    & Gemini &  1.8B, 3.25B & 2023 & & \checkmark & - & Web documents, books, and code, image data, audio data, video data\\
    & DeepSeek-Coder& 1.3B, 6.7B, 33B & 2024& - & \checkmark & 2T& GitHub’s Markdown and StackExchange\\
     & DocLLM & 1B,7B & 2024 & - & $\times$  & 2T& IIT-CDIP Test Collection 1.0, DocBank \\
    \bottomrule
  \end{tabular}
\end{table*}

\subsection{Early Pre-trained Neural Language Models}


Language modeling using neural networks was pioneered by \cite{rumelhart1985learning, elman1990finding, mahoney2000fast}.  
Bengio et al. \cite{bengio2000neural} developed one of the first neural language models (NLMs) that are comparable to n-gram models. 
Then, \cite{schwenk2006continuous} successfully applied NLMs to machine translation. 
The release of RNNLM (an open source NLM toolkit) by Mikolov  \cite{mikolov2011strategies, rnnlm} helped significantly popularize NLMs. 
Afterwards, NLMs based on recurrent neural networks (RNNs) and their variants, such as long short-term memory (LSTM) \cite{sutskever2014sequence} and gated recurrent unit (GRU) \cite{cho2014properties}, were widely used for many natural language applications including machine translation, text generation and text classification \cite{minaee2021deep}.

Then, the invention of the Transformer architecture \cite{vaswani2017attention} marks another milestone in the development of NLMs. By applying self-attention to compute in parallel for every word in a sentence or document an ``attention score'' to model the influence each word has on another, Transformers allow for much more parallelization than RNNs, which makes it possible to efficiently pre-train very big language models on large amounts of data on GPUs. These pre-trained language models (PLMs) can be fine-tuned for many downstream tasks.    

We group early popular Transformer-based PLMs, based on their neural architectures, into three main categories: encoder-only, decoder-only, and encoder-decoder models. 
Comprehensive surveys of early PLMs are provided in \cite{minaee2021deep,qiu2020pre}.

\subsubsection{Encoder-only PLMs}
As the name suggests, the encoder-only models only consist of an encoder network. These models are originally developed for language understanding tasks, such as text classification, where the models need to predict a class label for an input text.
Representative encoder-only models include BERT and its variants, e.g., RoBERTa, ALBERT, DeBERTa, XLM, XLNet, UNILM, as to be described below. 

\begin{figure}[h]
\begin{center}
    \includegraphics [scale=0.4] {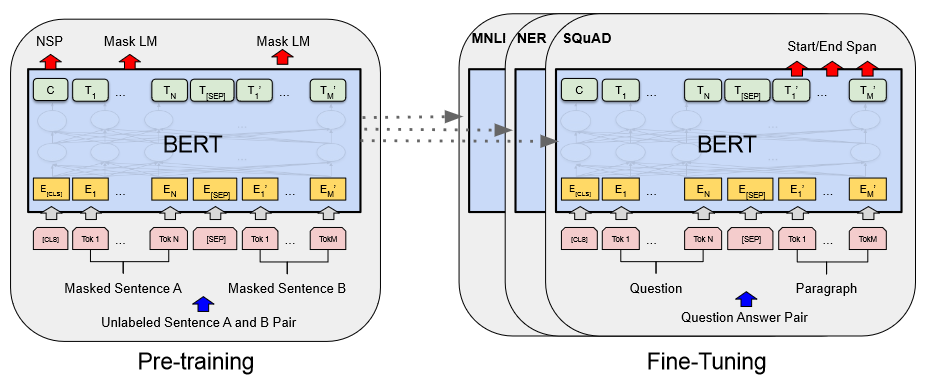}
\end{center}
  \caption{Overall pre-training and fine-tuning procedures for BERT. Courtesy of \cite{devlin2018bert}}
\label{fig:bert_arch}
\end{figure}

BERT (Birectional Encoder Representations from Transformers) \cite{devlin2018bert} is one of the most widely used encoder-only language models. BERT consists of three modules: (1) an embedding module that converts input text into a sequence of embedding vectors, (2) a stack of Transformer encoders that converts embedding vectors into contextual representation vectors, and (3) a fully connected layer that converts the representation vectors (at the final layer) to one-hot vectors. 
BERT is pre-trained uses two objectives:
masked language modeling (MLM) and next sentence prediction.
The pre-trained BERT model can be fine-tuned by adding a classifier layer for many language understanding tasks, ranging from text classification, question answering to language inference.
A high-level overview of BERT framework is shown in Fig \ref{fig:bert_arch}.
As BERT significantly improved state of the art on a wide range of language understanding tasks when it was published, the AI community was inspired to develop many similar encoder-only language models based on BERT. 

\begin{figure}[h]
\begin{center}
    \includegraphics [scale=0.4] {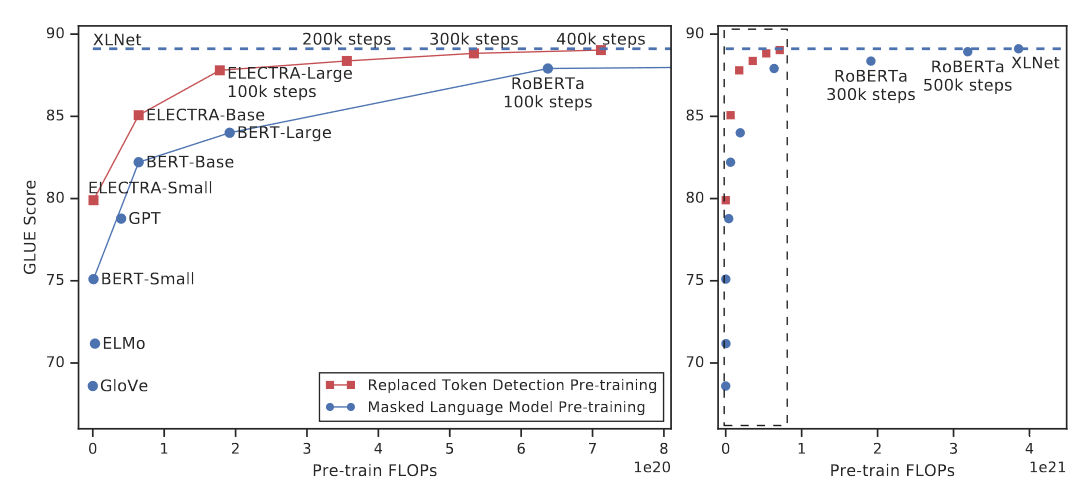}
\end{center}
  \caption{A comparison between replaced token detection and masked language modeling. Courtesy of \cite{clark2020electra}.}
\label{fig:electra}
\end{figure}

RoBERTa \cite{liu2019roberta} significantly improves the robustness of BERT using a set of model design choices and training strategies, such as modifying a few key hyperparameters, removing the next-sentence pre-training objective and training with much larger mini-batches and learning rates.
ALBERT \cite{lan2019albert} uses two parameter-reduction techniques to lower memory consumption and increase the training speed of BERT: (1) splitting the embedding matrix into two smaller matrices, and (2) using repeating layers split among groups.   
DeBERTa (Decoding-enhanced BERT with disentangled attention) \cite{he2020deberta} improves the BERT and RoBERTa models using two novel techniques.
The first is the disentangled attention mechanism, where
each word is represented using two vectors that encode its content and position, respectively, and the attention weights among words are computed using disentangled matrices on their contents and relative positions, respectively. Second, an enhanced mask decoder is used to incorporate absolute positions in the decoding layer to predict the masked tokens in model pre-training.
In addition, a novel virtual adversarial training method is used for fine-tuning to improve models’ generalization.
ELECTRA \cite{clark2020electra} uses a new pre-training task, known as replaced token detection (RTD), which is empirically proven to be more sample-efficient than MLM.
Instead of masking the input, RTD corrupts it by replacing some tokens with plausible alternatives sampled from a small generator network. Then, instead of training a model that predicts the original identities of the corrupted tokens, a discriminative model is trained to predict whether a token in the corrupted input was replaced by a generated sample or not.
RTD is more sample-efficient than MLM because the former is defined over all input tokens rather than just the small subset being masked out, as illustrated in Fig \ref{fig:electra}.

\begin{figure}[h]
\begin{center}
    \includegraphics [scale=0.4] {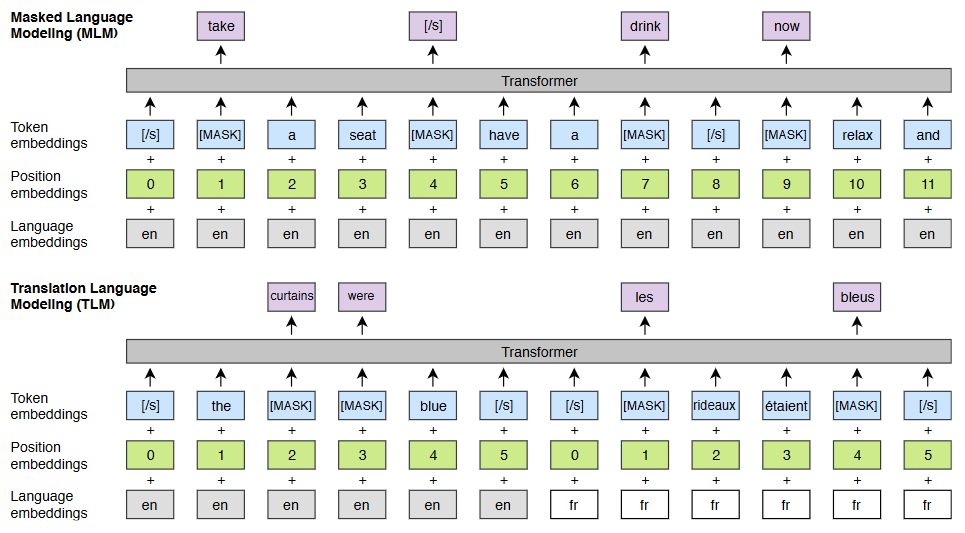}
\end{center}
  \caption{ Cross-lingual language model pretraining. The MLM objective is similar to BERT, but with continuous streams of text as opposed to sentence pairs. The TLM objective extends MLM to pairs of parallel sentences. To
  predict a masked English word, the model can attend to both the English sentence and its French translation, and is encouraged to align English and French representations. Courtesy of \cite{lample2019cross}.}
\label{fig:XLM}
\end{figure}

XLMs \cite{lample2019cross} extended BERT to cross-lingual language models using two methods: (1) a unsupervised method that only relies on monolingual data, and (2) a supervised method that leverages parallel data with a new cross-lingual language model objective, as illustrated in Fig \ref{fig:XLM}.
XLMs had obtained state-of-the-art results on cross-lingual classification, unsupervised and supervised machine translation, at the time they were proposed.

\begin{figure}[h]
\begin{center}
    \includegraphics [scale=0.4] {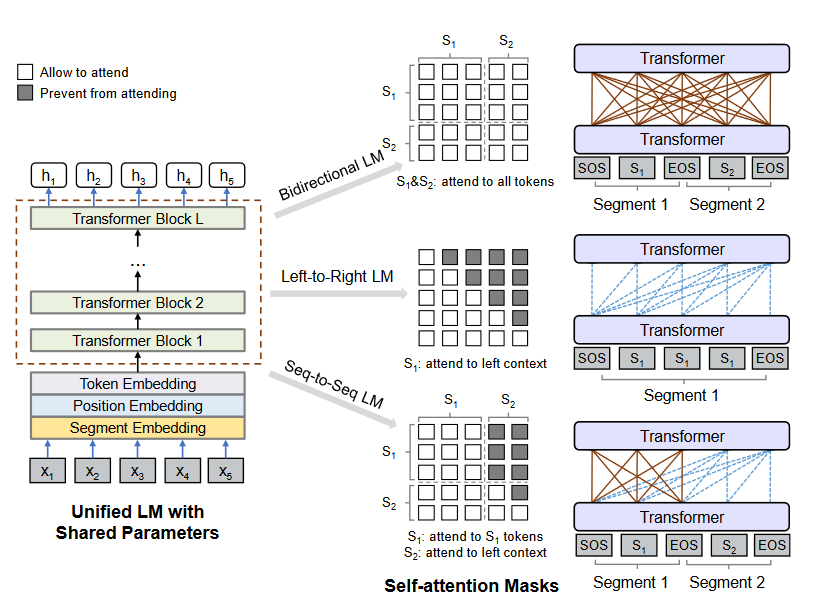}
\end{center}
  \caption{Overview of unified LM pre-training. The model parameters are shared across the LM
objectives (i.e., bidirectional LM, unidirectional LM, and sequence-to-sequence LM). Courtesy of \cite{dong2019unified}.}
\label{fig:UNILM}
\end{figure}

There are also encoder-only language models that leverage the advantages
of auto-regressive (decoder) models for model training and inference. 
Two examples are XLNet and UNILM.
XLNet \cite{yang2019xlnet} is based on Transformer-XL, pre-trained using a generalized autoregressive method that enables learning bidirectional contexts by maximizing the expected likelihood over all permutations of the factorization order.
UNILM (UNIfied pre-trained Language Model) \cite{dong2019unified} is pre-trained using three types of language modeling tasks: unidirectional, bidirectional, and sequence-to-sequence prediction. This is achieved by employing a shared Transformer network and utilizing specific self-attention masks to control what context the prediction is conditioned on, as illustrated in Fig \ref{fig:UNILM}. 
The pre-trained model can be fine-tuned for both natural language understanding and generation tasks. 

\subsubsection{Decoder-only PLMs}
Two of the most widely used decoder-only PLMs are GPT-1 and GPT-2, developed by OpenAI. These models lay the foundation to more powerful LLMs subsequently, i.e., GPT-3 and GPT-4. 

\begin{figure}[h]
\begin{center}
    \includegraphics [scale=0.4] {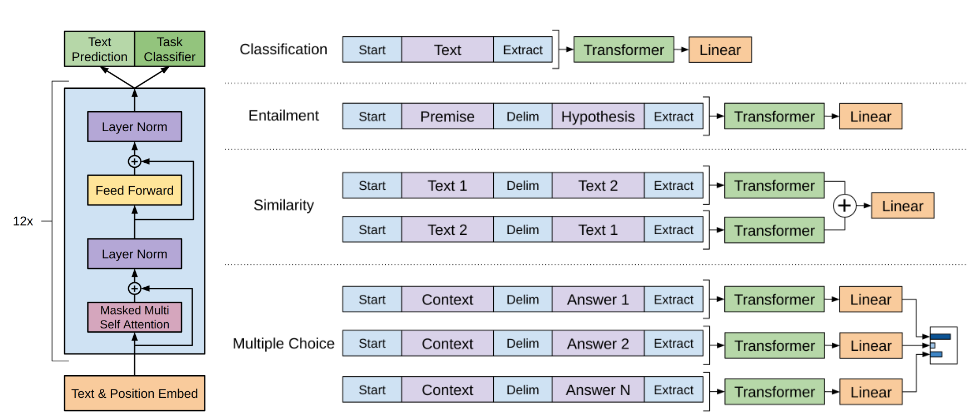}
\end{center}
  \caption{High-level overview of GPT pretraining, and fine-tuning steps. Courtesy of OpenAI.}
\label{fig:GPT}
\end{figure}

GPT-1 \cite{radford2018improving} demonstrates for the first time that good performance over a wide range of natural language tasks can be obtained by Generative Pre-Training (GPT) of a decoder-only Transformer model on a diverse corpus of unlabeled text in a self-supervised learning fashion (i.e., next word/token prediction), followed by discriminative fine-tuning on each specific downstream task (with much fewer samples),
as illustrated in Fig \ref{fig:GPT}. 
GPT-1 paves the way for subsequent GPT models, with each version improving upon the architecture and achieving better performance on various language tasks.

GPT-2 \cite{radford2019language} shows that language models are able to learn to perform specific natural language tasks without any explicit supervision when trained on a large WebText dataset consisting of millions of webpages.
The GPT-2 model follows the model designs of 
GPT-1 with a few modifications: Layer normalization is moved to the input of each sub-block, additional layer normalization is added after the final self-attention block, initialization is modified to account for the accumulation on the residual path and scaling the weights of residual layers, vocabulary size is expanded to 50,25, and context size is increased from 512 to 1024 tokens.

\subsubsection{Encoder-Decoder PLMs}

In \cite{raffel2020exploring}, Raffle et al. shows that almost all NLP tasks can be cast as a sequence-to-sequence generation task. Thus, an encoder-decoder language model, by design, is a unified model in that it can perform all natural language understanding and generation tasks.
Representative encoder-decoder PLMs we will review below are T5, mT5, MASS, and BART.

T5 \cite{raffel2020exploring} is a Text-to-Text Transfer Transformer (T5) model, where transfer learning is effectively exploited for NLP via an introduction of a unified framework in which all NLP tasks are cast as a text-to-text generation task.
mT5 \cite{xue2020mt5} is a multilingual variant of T5, which is pre-trained on a new Common Crawl-based dataset consisting of texts in 101 languages.

MASS (MAsked Sequence to Sequence pre-training) \cite{song2019mass} adopts the encoder-decoder framework to reconstruct a sentence fragment given the remaining part of the sentence. The encoder takes a sentence with randomly masked fragment (several consecutive tokens) as input, and the decoder predicts the masked fragment. In this way, MASS jointly trains the encoder and decoder for language embedding and generation, respectively. 

BART \cite{lewis2019bart} uses a standard sequence-to-sequence translation model architecture. It is pre-trained by corrupting text with an arbitrary noising function, and then learning to reconstruct the original text. 



\subsection{Large Language Model Families}
\begin{figure*}[h]
\begin{center}
    \includegraphics [scale=0.75] {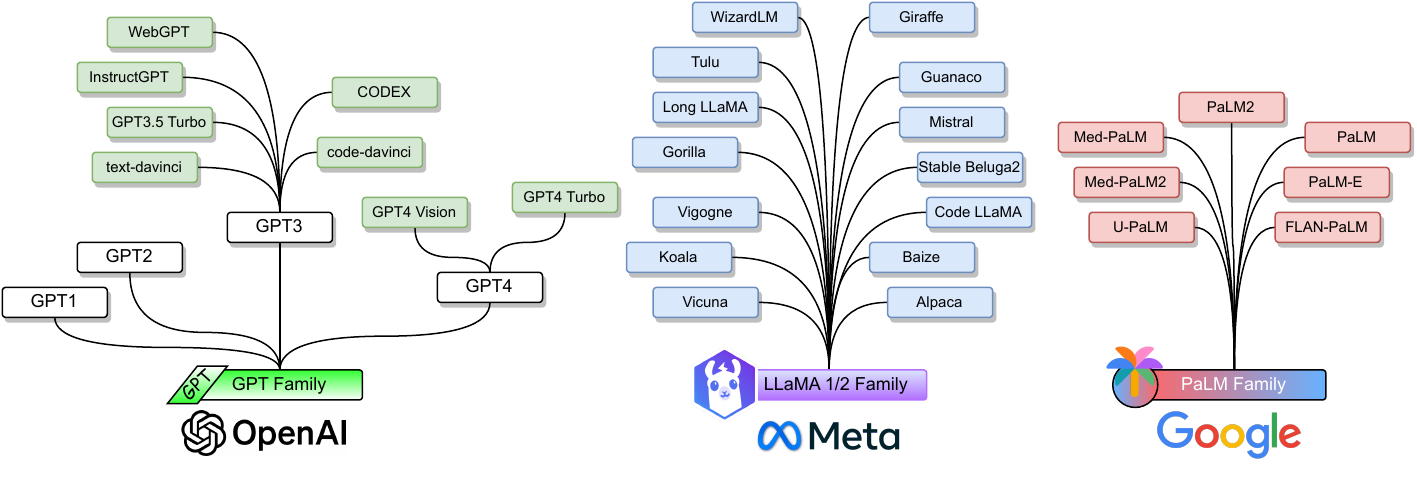}
\end{center}
  \caption{Popular LLM Families.}
\label{fig:llm_family}
\end{figure*}

Large language models (LLMs) mainly refer to transformer-based PLMs that contain tens to hundreds of billions of parameters. Compared to PLMs reviewed above,
LLMs are not only much larger in model size, but also exhibit stronger language understanding and generation and emergent abilities that are not present in smaller-scale models.
In what follows, we review three LLM families: GPT, LLaMA, and PaLM, as illustrated in Fig \ref{fig:llm_family}. 


\subsubsection{\textbf{The GPT Family}}

Generative Pre-trained Transformers (GPT) are a family of decoder-only Transformer-based language models, developed by OpenAI. This family consists of
GPT-1, GPT-2, GPT-3, InstrucGPT, ChatGPT, GPT-4, CODEX, and WebGPT.
Although early GPT models, such as GPT-1 and GPT-2, are open-source, recent models, such as GPT-3 and GPT-4, are close-source and can only be accessed via APIs.
GPT-1 and GPT-2 models have been discussed in the early PLM subsection. We start with GPT-3 below.

\begin{figure}[h]
\begin{center}
    \includegraphics [scale=0.4] {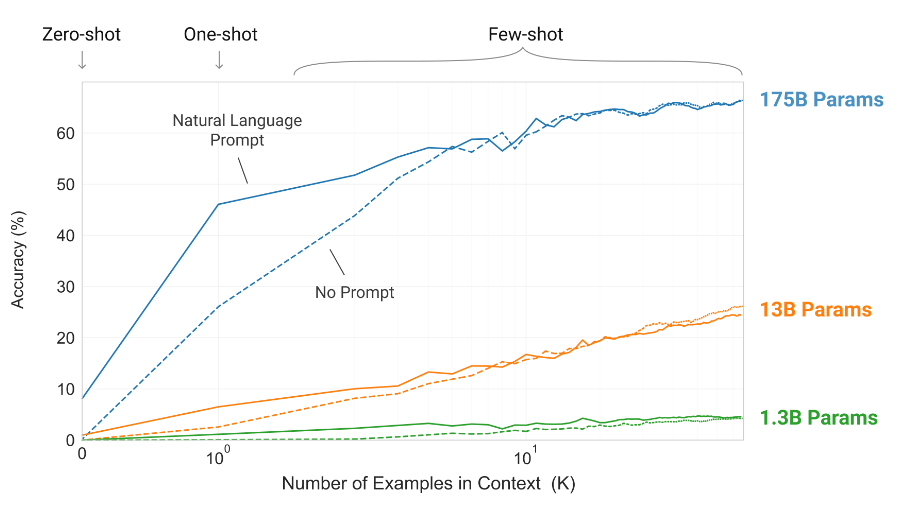}
\end{center}
  \caption{GPT-3 shows that larger models make increasingly efficient use of in-context information. It shows in-context learning performance on a simple task requiring the model to remove random symbols from a word, both with and without a
natural language task description. Courtesy of \cite{brown2020language}.}
\label{fig:gpt3_size}
\end{figure}

GPT-3 \cite{brown2020language} is a pre-trained autoregressive language model with 175 billion parameters. 
GPT-3 is widely considered as the first LLM in that not only it is much larger than previous PLMs, but also for the first time demonstrates emergent abilities that are not observed in previous smaller PLMs.
GPT-3 shows the emergent ability of in-context learning, which means GPT-3 can be applied to any downstream tasks without any gradient updates or fine-tuning, with tasks and few-shot demonstrations specified purely via text interaction with the model.
GPT-3 achieved strong performance on many NLP tasks, including translation, question-answering, and the cloze tasks, as well as several ones that require on-the-fly reasoning or domain adaptation, such as unscrambling words, using a novel word in a sentence, 3-digit arithmetic.
Fig \ref{fig:gpt3_size} plots the performance of GPT-3 as a function of the number of examples in in-context prompts.

CODEX \cite{chen2021evaluating}, released by OpenAI in March 2023, is a general-purpose programming model that can parse natural language and generate code in response. 
CODEX is a descendant of GPT-3, fine-tuned for programming applications on code corpora collected from GitHub. 
CODEX powers Microsoft's GitHub Copilot.

WebGPT \cite{nakano2021webgpt} is another descendant of GPT-3, fine-tuned to answer open-ended questions using a text-based web browser, facilitating users to search and navigate the web.
Specifically, WebGPT is trained in three steps. The first is for WebGPT to learn to mimic human browsing behaviors using human demonstration data. Then, a reward function is learned to predict human preferences. Finally, WebGPT is refined to optimize the reward function via reinforcement learning and rejection sampling. 

\begin{figure}[h]
\begin{center}
    \includegraphics [scale=0.4] {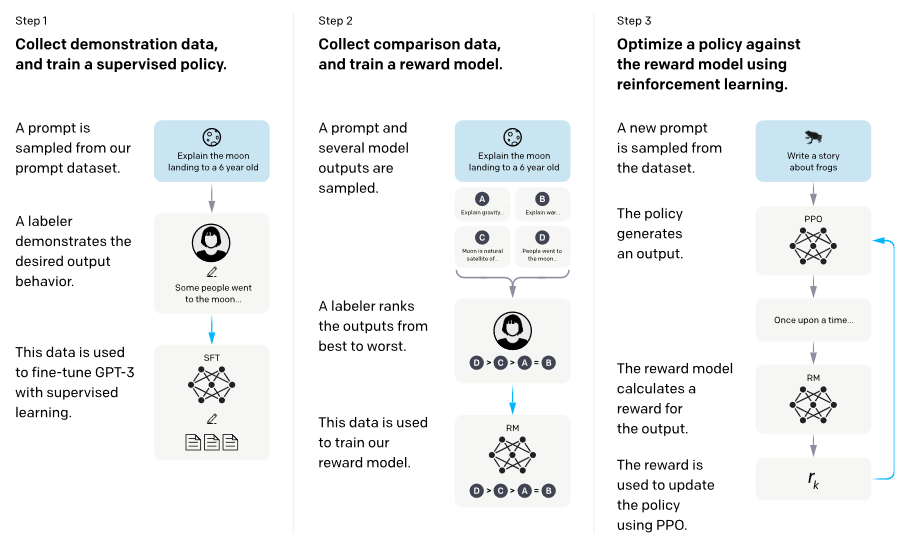}
\end{center}
  \caption{The high-level overview of RLHF. Courtesy of \cite{ouyang2022training}.}
\label{fig:sft}
\end{figure}

To enable LLMs to follow expected human instructions, InstructGPT \cite{ouyang2022training} is
proposed to align language models with user intent on a wide range of tasks by fine-tuning with human feedback.
Starting with a set of labeler-written prompts and prompts submitted through the OpenAI API, a dataset of labeler demonstrations of the desired model behavior is collected. Then GPT-3 is fine-tuned on this dataset. Then, a dataset of human-ranked model outputs is collected to further fine-tune the model using reinforcement learning. The method is known Reinforcement Learning from Human Feedback (RLHF), as shown in \ref{fig:sft}.
The resultant InstructGPT models have shown improvements in truthfulness and reductions in toxic output generation while having minimal performance regressions on public NLP datasets.

The most important milestone of LLM development is the launch of ChatGPT (Chat Generative Pre-trained Transformer) \cite{chatgpt} on November 30, 2022. 
ChatGPT is chatbot that enables users to steer a conversation to complete a wide range of tasks such as question answering, information seeking, text summarization, and more.
ChatGPT is powered by GPT-3.5 (and later by GPT-4), a sibling model to InstructGPT, which is trained to follow an instruction in a prompt and provide a detailed response.



\begin{figure}[h]
\begin{center}
    \includegraphics [scale=0.4] {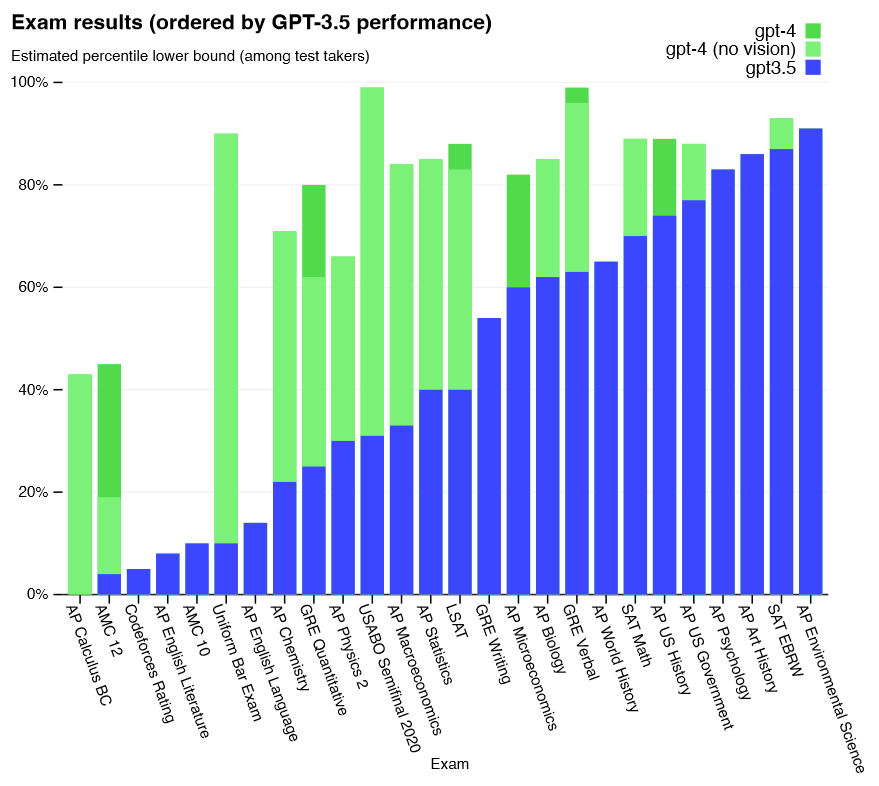}
\end{center}
  \caption{GPT-4 performance on academic and professional exams, compared with GPT 3.5. Courtesy of \cite{gpt4}.}
\label{fig:gpt4}
\end{figure}

GPT-4 \cite{gpt4} is the latest and most powerful LLM in the GPT family.  Launched in March, 2023, GPT-4 is a multimodal LLM in that it can take image and text as inputs and produce text outputs. 
While still less capable than humans in some of the most challenging real-world scenarios, GPT-4 exhibits human-level performance on various professional and academic benchmarks, including passing a simulated bar exam with a score around the top 10\% of test takers, as shown in Fig \ref{fig:gpt4}.
Like early GPT models, GPT-4 was first pre-trained to predict next tokens on large text corpora, and then fine-tuned with RLHF to align model behaviors with human-desired ones. 

\subsubsection{\textbf{The LLaMA Family}}
LLaMA is a collection of foundation language models, released by Meta. 
Unlike GPT models, LLaMA models are open-source, i.e., model weights are released to the research community under a noncommercial license. 
Thus, the LLaMA family grows rapidly as these models are widely used by many research groups to develop better open-source LLMs to compete the closed-source ones or to develop task-specific LLMs for mission-critical applications. 

The first set of LLaMA models \cite{touvron2023llama} was released in February 2023, ranging from 7B to 65B parameters.
These models are pre-trained on trillions of tokens, collected from publicly available datasets.
LLaMA uses the transformer architecture of GPT-3, with a few minor architectural modifications, including (1) using a SwiGLU activation function instead of ReLU,
(2) using rotary positional embeddings instead of absolute positional embedding, and (3) using root-mean-squared layer-normalization instead of standard layer-normalization.
The open-source LLaMA-13B model outperforms the proprietary GPT-3 (175B) model on most benchmarks, making it a good baseline for LLM research.

\begin{figure}[h]
\begin{center}
    \includegraphics [scale=0.4] {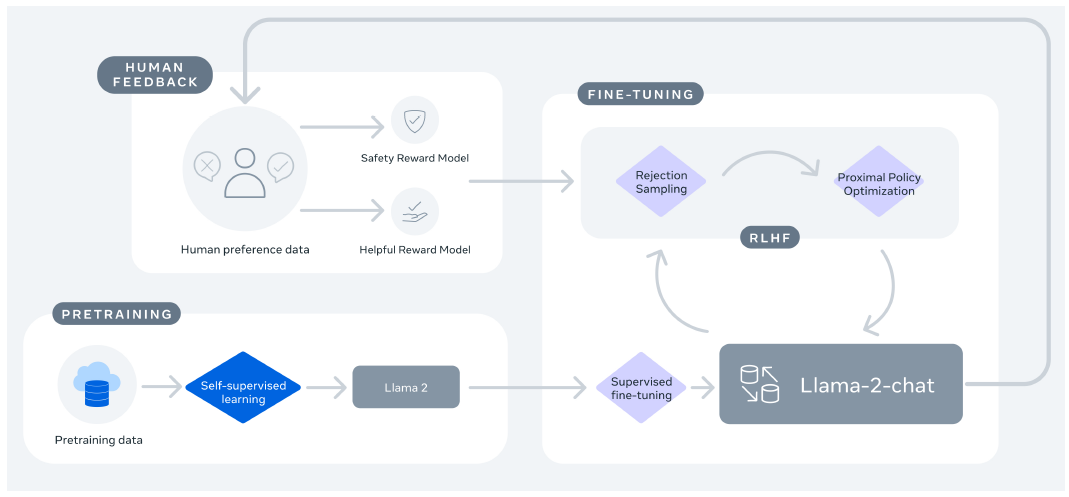}
\end{center}
  \caption{Training of LLaMA-2 Chat. Courtesy of \cite{touvron2023llama2}.}
\label{fig:llama2}
\end{figure}

In July 2023, Meta, in partnership with Microsoft, released the LLaMA-2 collection \cite{touvron2023llama2}, which include both foundation language models and Chat models finetuned for dialog, known as LLaMA-2 Chat.
The LLaMA-2 Chat models were reported to outperform other open-source models on many public benchmarks. 
Fig \ref{fig:llama2} shows the training process of LLaMA-2 Chat. The process begins with pre-training LLaMA-2 using publicly available online data. Then, an initial version of LLaMA-2 Chat is built via supervised fine-tuning. Subsequently, the model is iteratively refined using RLHF, rejection sampling and proximal policy optimization. In the RLHF stage, the accumulation of human feedback for revising the reward model 
is crucial to prevent the reward model from being changed too much, which could hurt the stability of LLaMA model training.

Alpaca \cite{taori2023alpaca} is fine-tuned from the LLaMA-7B model using 52K instruction-following demonstrations generated in the style of self-instruct using GPT-3.5 (text-davinci-003). 
Alpaca is very cost-effective for training, especially for academic research.
On the self-instruct evaluation set, Alpaca performs similarly to GPT-3.5, despite that Alpaca is much smaller. 

\begin{figure}[h]
\begin{center}
    \includegraphics [scale=0.4] {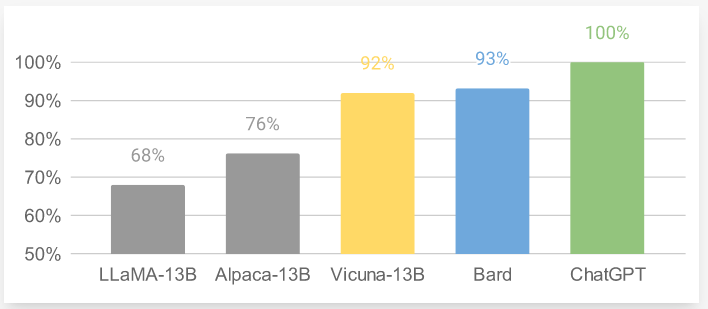}
\end{center}
  \caption{Relative Response Quality of Vicuna and a few other well-known models by GPT-4. Courtesy of Vicuna Team.}
\label{fig:Vicuna}
\end{figure}

The Vicuna team has developed a 13B chat model, Vicuna-13B, by fine-tuning LLaMA on user-shared conversations collected from ShareGPT. Preliminary evaluation using GPT-4 as a evaluator shows that Vicuna-13B achieves more than 90\% quality of OpenAI's ChatGPT, and Google's Bard while outperforming other models like LLaMA and Stanford Alpaca in more than 90\% of cases. 
\ref{fig:Vicuna} shows the relative response quality of Vicuna and a few other well-known models by GPT-4.
Another advantage of Vicuna-13B is its relative limited computational demand for model training. The training cost of Vicuna-13B is merely \$300.

Like Alpaca and Vicuna, the Guanaco models \cite{dettmers2023qlora} are also finetuned LLaMA models using instruction-following data. But the finetuning is done very efficiently using QLoRA such that 
finetuning a 65B parameter model can be done on a single 48GB GPU.
QLoRA back-propagates gradients through a frozen, 4-bit quantized pre-trained language model into Low Rank Adapters (LoRA). 
The best Guanaco model outperforms all previously released models on the Vicuna benchmark, reaching 99.3\% of the performance level of ChatGPT while only requiring 24 hours of fine-tuning on a single GPU.

Koala \cite{geng2023koala} is yet another instruction-following language model built on LLaMA, 
but with a specific focus on interaction data that include user inputs and responses generated by highly capable closed-source chat models such as ChatGPT. 
The Koala-13B model performs competitively with state-of-the-art chat models according to human evaluation based on real-world user prompts.




Mistral-7B \cite{jiang2023mistral} is a 7B-parameter language model engineered for superior performance and efficiency. Mistral-7B outperforms the best open-source 13B model (LLaMA-2-13B) across all evaluated benchmarks, and the best open-source 34B model (LLaMA-34B) in reasoning, mathematics, and code generation. This model leverages grouped-query attention for faster inference, coupled with sliding window attention to effectively handle sequences of arbitrary length with a
reduced inference cost.

The LLaMA family is growing rapidly, as more instruction-following models have been built on LLaMA or LLaMA-2, including
Code LLaMA \cite{roziere2023code}, Gorilla \cite{patil2023gorilla}, Giraffe \cite{pal2023giraffe}, Vigogne \cite{vigogne}, Tulu 65B \cite{wang2023far}, Long LLaMA \cite{tworkowski2023focused}, and Stable Beluga2 \cite{StableBelugaModels}, just to name a few.

\subsubsection{\textbf{The PaLM Family}}
The PaLM (Pathways Language Model) family are developed by Google. 
The first PaLM model \cite{chowdhery2022palm} was announced in April 2022 and remained private until March 2023. It is a 540B parameter transformer-based LLM.
The model is pre-trained on a high-quality text corpus consisting of 780 billion tokens that comprise a wide range of natural language tasks and use cases.
PaLM is pre-trained on 6144 TPU v4 chips using the Pathways system, which enables highly efficient training across multiple TPU Pods. 
PaLM demonstrates continued benefits of scaling by achieving state-of-the-art few-shot learning results on hundreds of language understanding and generation benchmarks. 
PaLM-540B outperforms not only state-of-the-art fine-tuned models on a suite of multi-step reasoning tasks, but also on par with humans on the recently released BIG-bench benchmark.

The U-PaLM models of 8B, 62B, and 540B scales are continually trained on PaLM with UL2R, a method of continue training LLMs on a few steps with UL2’s mixture-of-denoiser objective \cite{tay2022transcending}. An approximately 2x computational savings rate is reported.

\begin{figure}[h]
\begin{center}
    \includegraphics [scale=0.4] {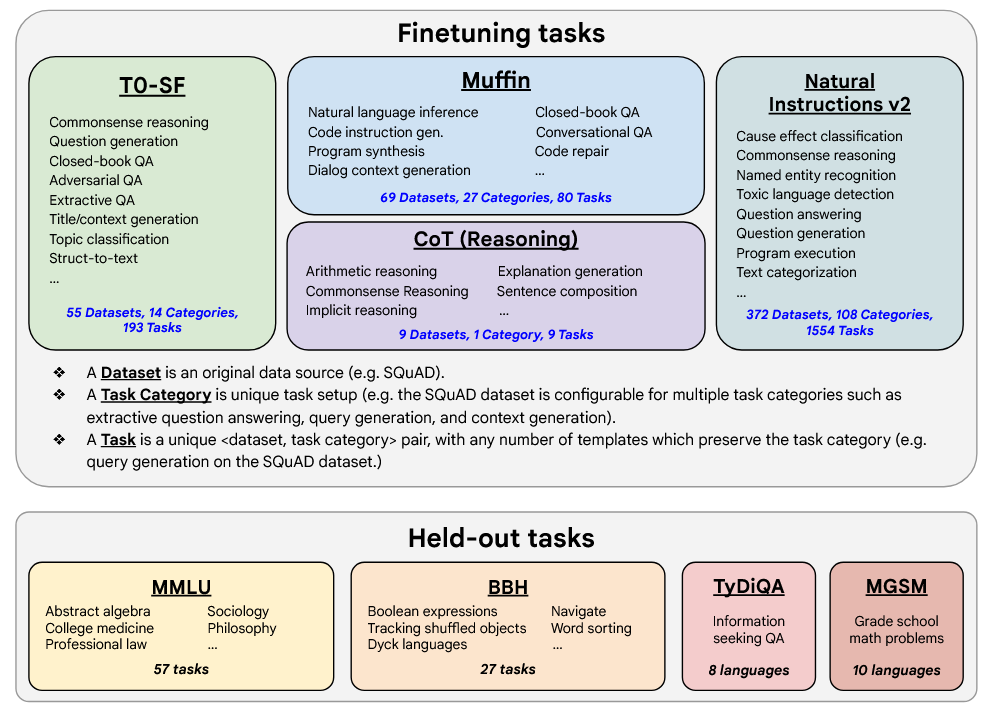}
\end{center}
  \caption{Flan-PaLM finetuning consist of 473 datasets in above task categories. Courtesy of \cite{chung2022scaling}.}
\label{fig:FlanPaLM}
\end{figure}

U-PaLM is later instruction-finetuned as Flan-PaLM \cite{chung2022scaling}. Compared to other instruction finetuning work mentioned above, Flan-PaLM's finetuning is performed using a much larger number of tasks, larger model sizes, and chain-of-thought data. As a result, Flan-PaLM substantially outperforms previous instruction-following models.
For instance, Flan-PaLM-540B, which is instruction-finetuned on 1.8K tasks, outperforms PaLM-540B by a large margin (+9.4\% on average).
The finetuning data comprises 473 datasets, 146 task categories, and 1,836 total tasks, as illustrated in Fig \ref{fig:FlanPaLM}.

PaLM-2 \cite{anil2023palm} is a more compute-efficient LLM with better multilingual and reasoning capabilities, compared to its predecessor PaLM. 
PaLM-2 is trained using a mixture of objectives.
Through extensive evaluations on English, multilingual, and reasoning tasks, PaLM-2 significantly improves the model performance on downstream tasks across different model sizes, while simultaneously exhibiting faster and more efficient inference than PaLM.


Med-PaLM \cite{singhal2022large} is a domain-specific PaLM, and is designed to provide high-quality answers to medical questions.
Med-PaLM is finetuned on PaLM using instruction prompt tuning, a parameter-efficient method for aligning LLMs to new domains using a few exemplars.
Med-PaLM obtains very encouraging results on many healthcare tasks, although it is still inferior to human clinicians.
Med-PaLM 2 improves Med-PaLM via med-domain finetuning and ensemble prompting \cite{singhal2023towards}. Med-PaLM 2 scored up to 86.5\% on the MedQA dataset (i.e., a benchmark combining six existing open question answering datasets spanning professional medical exams, research, and consumer queries), improving upon Med-PaLM by over 19\% and setting a new state-of-the-art. 



\subsection{Other Representative LLMs}
In addition to the models discussed in the previous subsections, there are other popular LLMs which do not belong to those three model families, yet they have achieved great performance and have pushed the LLMs field forward.
We briefly describe these LLMs in this subsection.

\begin{figure}[h]
\begin{center}
    \includegraphics [scale=0.45] {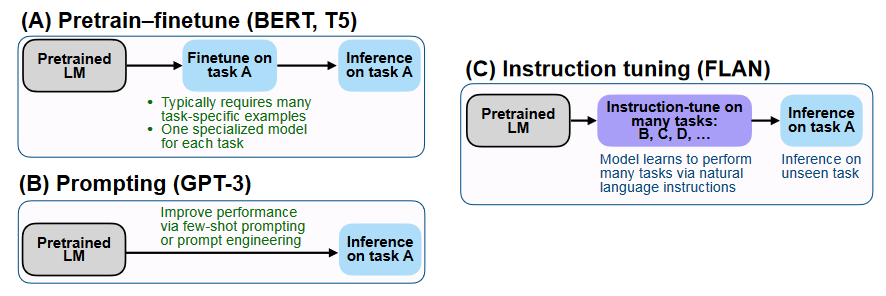}
\end{center}
  \caption{comparison of instruction tuning with pretrain–finetune and prompting. Courtesy of \cite{wei2021finetuned}.}
\label{fig:flan}
\end{figure}

\textbf{FLAN:} In \cite{wei2021finetuned}, Wei et al. explored a simple method for improving the zero-shot learning abilities of language models. They showed that instruction tuning language models on a collection of datasets described via instructions substantially improves zero-shot performance on unseen tasks.
They take a 137B parameter pretrained language model and instruction tune it on over 60 NLP datasets verbalized via natural language instruction templates.
They call this instruction-tuned model FLAN.
Fig \ref{fig:flan} provides a comparison of instruction tuning with pretrain–finetune and prompting.

\begin{figure}[h]
\begin{center}
    \includegraphics [scale=0.4] {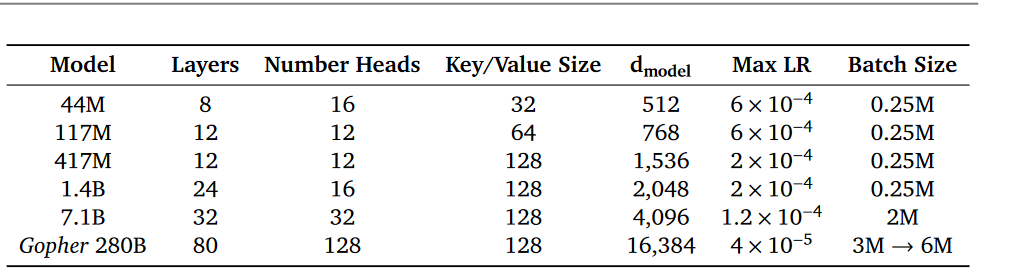}
\end{center}
  \caption{Model architecture details of Gopher with different number of parameters. Courtesy of \cite{wei2021finetuned}.}
\label{fig:gopher}
\end{figure}

\textbf{Gopher:} In \cite{rae2021scaling}, Rae et al. presented an analysis of Transformer-based language model performance across a wide range of model scales — from models with tens of millions of parameters up to a 280 billion parameter model called Gopher.
These models were evaluated on 152 diverse tasks, achieving state-of-the-art performance across the majority. 
The number of layers, the key/value size, and other hyper-parameters of different model sizes are shown in Fig \ref{fig:gopher}.

\textbf{T0:} In \cite{sanh2021multitask}, Sanh et al. developed T0, a system for easily mapping any natural language tasks into a human-readable prompted form. They converted a large set of supervised datasets, each with multiple prompts with diverse wording. These prompted datasets allow for benchmarking the ability of a model to perform completely held-out tasks. 
Then, a T0 encoder-decoder model is developed to consume textual
inputs and produces target responses. The model is trained on a multitask mixture of NLP datasets partitioned into different tasks.

\begin{figure}[h]
\begin{center}
    \includegraphics [scale=0.4] {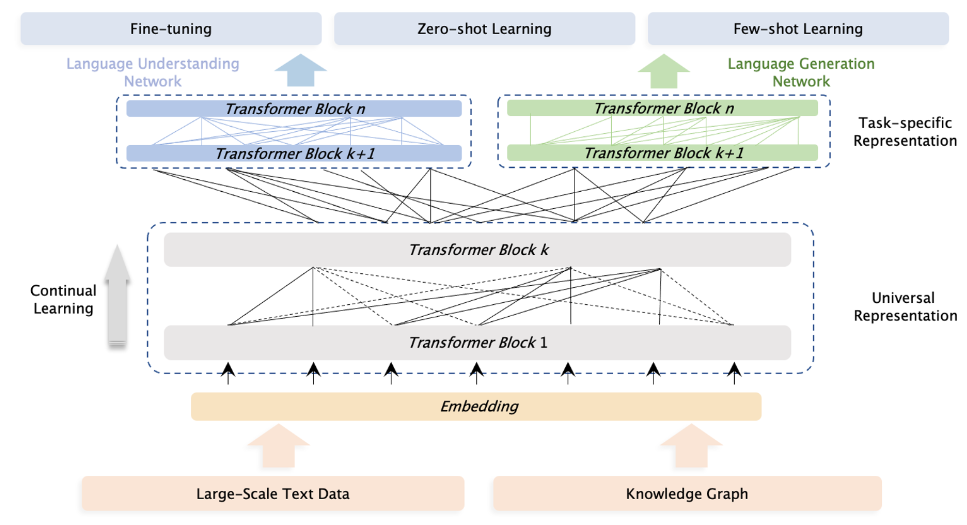}
\end{center}
  \caption{High-level model architecture of ERNIE 3.0. Courtesy of \cite{sun2021ernie}.}
\label{fig:ernie3}
\end{figure}

\textbf{ERNIE 3.0:} In \cite{sun2021ernie}, Sun et al.  proposed a unified framework named ERNIE 3.0 for pre-training large-scale knowledge enhanced models. It fuses auto-regressive network and auto-encoding network, so that the trained model can be easily tailored for both natural language understanding and generation tasks using zero-shot learning, few-shot learning or fine-tuning. They have trained ERNIE 3.0 with 10 billion parameters on a 4TB corpus consisting of plain texts and a large-scale knowledge graph.
Fig \ref{fig:ernie3} illustrates the model architecture of  Ernie 3.0.

\begin{figure}[h]
\begin{center}
    \includegraphics [scale=0.37] {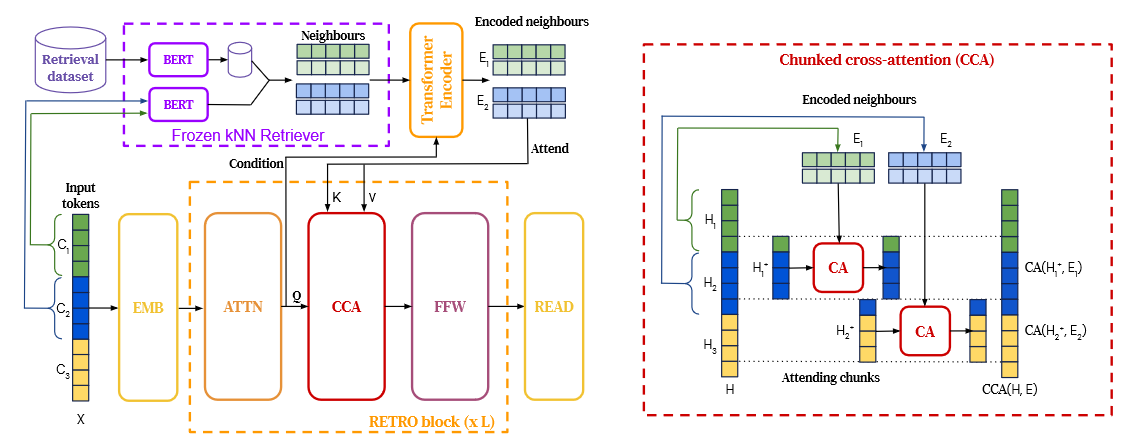}
\end{center}
  \caption{Retro architecture. Left: simplified version where a sequence of length n = 12 is split into l = 3 chunks of size m = 4. For each chunk, we retrieve k = 2 neighbours of r = 5 tokens each. The retrieval pathway is shown on top. Right: Details of the interactions in the CCA operator. Causality is maintained as neighbours of the first chunk only affect the last token of the first chunk and tokens from the second chunk. Courtesy of \cite{borgeaud2022improving}.}
\label{fig:retro}
\end{figure}

\textbf{RETRO:} In \cite{borgeaud2022improving}, Borgeaud et al. enhanced auto-regressive language models by conditioning on document chunks retrieved from a large corpus, based on local similarity with preceding tokens. Using a 2-trillion-token database, the
Retrieval-Enhanced Transformer (Retro) obtains comparable performance to GPT-3 and Jurassic-1 \cite{lieber2021jurassic} on the Pile, despite using 25\% fewer parameters.
As shown in Fig \ref{fig:retro}, Retro combines a frozen Bert retriever, a differentiable encoder and a chunked cross-attention mechanism to predict tokens based on an order of magnitude more data than what is typically consumed during training.

\begin{figure}[h]
\begin{center}
    \includegraphics [scale=0.45] {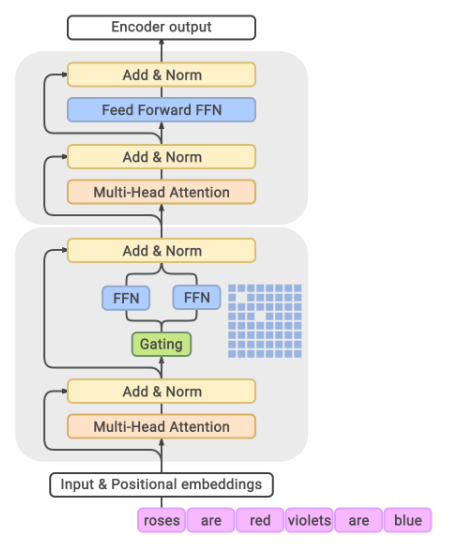}
\end{center}
  \caption{GLaM model architecture. Each MoE layer (the bottom
block) is interleaved with a Transformer layer (the upper block). Courtesy of \cite{du2022glam}.}
\label{fig:GLAM}
\end{figure}

\textbf{GLaM:} In \cite{du2022glam}, Du et al. proposed a family of LLMs named GLaM (Generalist Language Model),
which use a sparsely activated mixture-of-experts
architecture to scale the model capacity while also
incurring substantially less training cost compared
to dense variants. 
The largest GLaM has 1.2 trillion parameters, which is approximately 7x larger than GPT-3. It consumes only 1/3 of the energy used to train GPT-3 and requires half of the computation flops for inference, while still achieving better overall zero, one and few-shot performance across 29 NLP tasks.
Fig \ref{fig:GLAM} shows the high-level architecture of GLAM.

\textbf{LaMDA:} In \cite{thoppilan2022lamda}, Thoppilan et al. presented LaMDA, a family of Transformer-based neural language models specialized for dialog, which have up to 137B parameters and are pre-trained on 1.56T words of public dialog data and web text.
They showed that fine-tuning with annotated data and enabling the model to consult external knowledge sources can lead to significant improvements towards the two key challenges of safety and factual grounding.

\begin{figure}[h]
\begin{center}
    \includegraphics [scale=0.4] {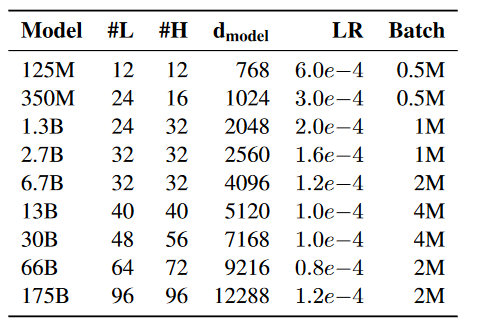}
\end{center}
  \caption{Different OPT Models' architecture details. Courtesy of \cite{zhang2022opt}.}
\label{fig:opt}
\end{figure}

\textbf{OPT:} In \cite{zhang2022opt}, Zhang et al. presented Open Pre-trained Transformers (OPT), a suite of decoder-only pre-trained transformers ranging from 125M to 175B parameters, which they share with researchers.
The OPT models' parameters are shown in \ref{fig:opt}

\textbf{Chinchilla:} In \cite{hoffmann2022training}, Hoffmann et al. investigated the optimal model size and number of tokens for training a transformer language model under a given compute budget.
By training over 400 language models ranging from 70 million to over 16 billion parameters on 5 to 500 billion tokens, they found that for compute-optimal training, the model size and the number of training tokens should be scaled equally: for every doubling of model size the number of training tokens should also be doubled.
They tested this hypothesis by training a predicted compute-optimal model, Chinchilla, that uses the same compute budget as Gopher but with 70B parameters and 4\% more more data.

\textbf{Galactica:} In \cite{taylor2022galactica}, Taylor et al. introduced Galactica, a large language model that can store, combine and reason about scientific knowledge. They trained on a large scientific corpus of papers, reference material, knowledge bases and many other sources.
Galactica performed well on reasoning, outperforming Chinchilla on mathematical MMLU by 41.3\% to 35.7\%, and PaLM 540B on MATH with a score of 20.4\% versus 8.8\%.

\textbf{CodeGen:} In \cite{nijkamp2022codegen}, Nijkamp et al.  trained and released a family of large
language models up to 16.1B parameters, called CODEGEN, on natural language and programming language data, and open sourced the training library JAXFORMER. They showed the utility of the trained model by demonstrating that it is competitive with the previous state-of-the-art on zero-shot Python code generation on HumanEval. They further investigated the multi-step paradigm for program synthesis, where a single program is factorized into multiple prompts specifying sub-problems. They also  constructed an open benchmark, Multi-Turn Programming Benchmark (MTPB), consisting of 115 diverse problem sets that are factorized into multi-turn prompts.

\textbf{AlexaTM:} In \cite{soltan2022alexatm}, Soltan et al. demonstrated that multilingual large-scale sequence-to-sequence (seq2seq) models, pre-trained on a mixture of denoising and Causal Language Modeling (CLM) tasks, are more efficient few-shot learners than decoder-only models on various task.
They trained a 20 billion parameter multilingual seq2seq model called Alexa Teacher Model (AlexaTM 20B) and showed that it achieves state-of-the-art (SOTA) performance on 1-shot summarization tasks, outperforming a much larger 540B PaLM decoder model. AlexaTM consist of 46 encoder layers, 32 decoder layers, 32 attention heads, and $d_{model}= 4096$.

\begin{figure}[h]
\begin{center}
    \includegraphics [scale=0.4] {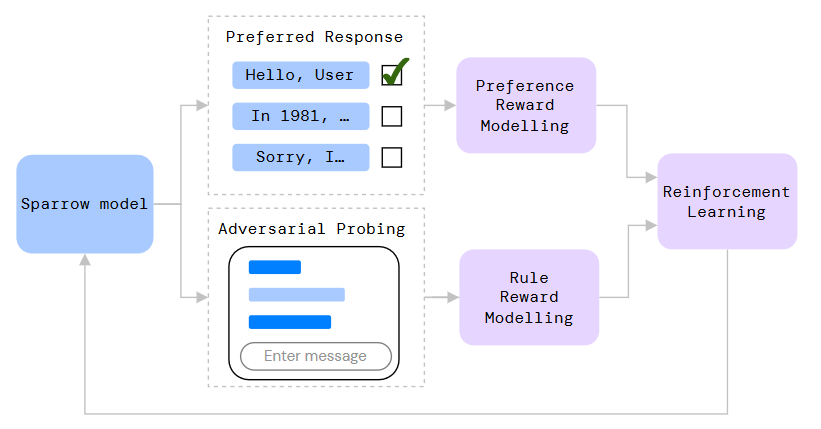}
\end{center}
  \caption{Sparrow pipeline relies on human participation to continually expand a training set. Courtesy of \cite{glaese2022improving}.}
\label{fig:sparrow}
\end{figure}

\textbf{Sparrow:} In \cite{glaese2022improving}, Glaese et al. presented Sparrow, an information-seeking dialogue agent trained to be more helpful, correct, and harmless compared to prompted language model baselines. They used reinforcement learning from human feedback to train their models with two new additions to help human raters judge agent behaviour.
The high-level pipeline of Sparrow model is shown in Fig \ref{fig:sparrow}.

\textbf{Minerva:} In \cite{lewkowycz2022solving}, Lewkowycz et al. introduced Minerva, a large language model pretrained on general natural language data and further trained on technical content, to tackle previous LLM struggle with quantitative reasoning (such as solving mathematics, science, and engineering problems).

\textbf{MoD:} In \cite{tay2022unifying}, Tay et al. presented a generalized and unified perspective for self-supervision in NLP and show how different pre-training objectives can be cast as one another and how interpolating between different objectives can be effective. They proposed Mixture-of-Denoisers (MoD), a pre-training objective that combines diverse pre-training paradigms together.
This framework is known as Unifying Language Learning (UL2).
An overview of UL2 pretraining paradigm is shown in Fig \ref{fig:sparrow}.
\begin{figure}[h]
\begin{center}
    \includegraphics [scale=0.4] {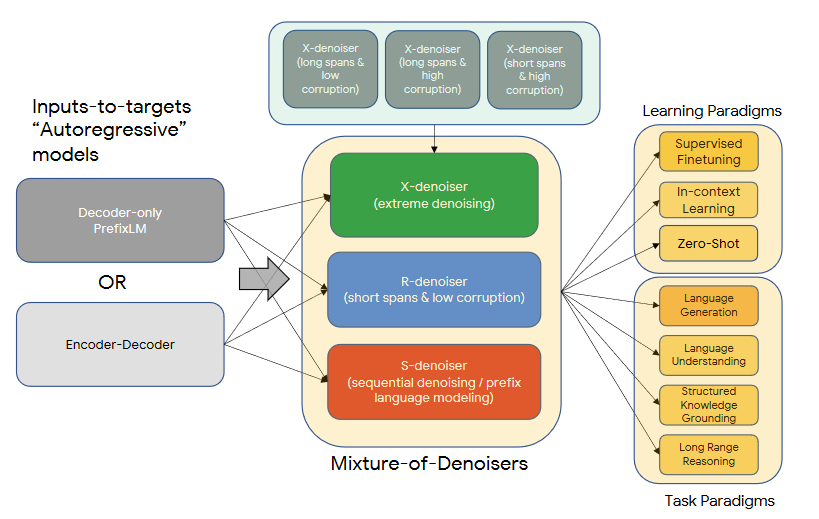}
\end{center}
  \caption{An overview of UL2 pretraining paradigm. Courtesy of \cite{tay2022unifying}.}
\label{fig:ul2}
\end{figure}

\begin{figure}[h]
\begin{center}
    \includegraphics [scale=0.4] {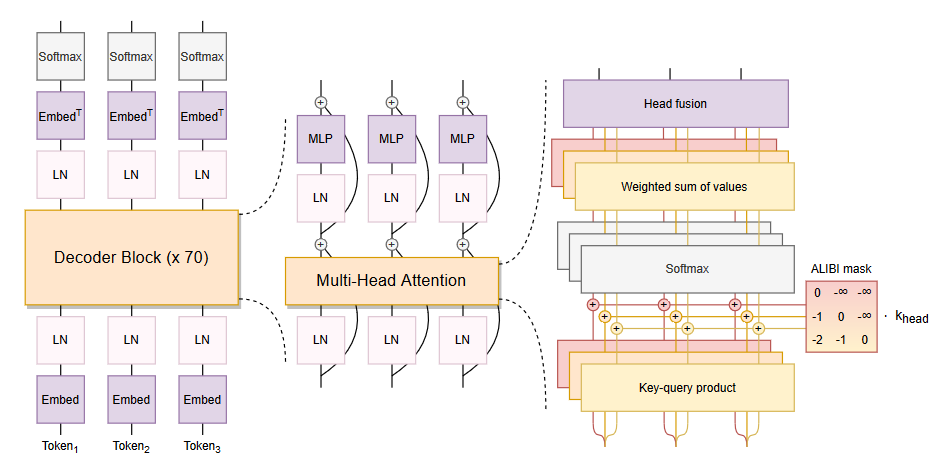}
\end{center}
  \caption{An overview of BLOOM architecture. Courtesy of \cite{scao2022bloom}.}
\label{fig:bloom}
\end{figure}

\textbf{BLOOM:} In \cite{scao2022bloom}, Scao et al. presented BLOOM, a 176B-parameter open-access language model designed and built thanks to a collaboration of hundreds of researchers. BLOOM is a decoder-only Transformer language model trained on the ROOTS corpus, a dataset comprising hundreds of sources in 46 natural and 13 programming languages (59 in total).
An overview of BLOOM architecture is shown in Fig \ref{fig:bloom}.


\textbf{GLM:} In \cite{zeng2022glm}, Zeng et al. introduced GLM-130B, a bilingual (English and Chinese) pre-trained language model with 130 billion parameters. It was an attempt to open-source a 100B-scale model at least as good as GPT-3 (davinci) and unveil how models of such a scale
can be successfully pre-trained.

\textbf{Pythia:} In \cite{biderman2023pythia}, Biderman et al. introduced Pythia, a suite of 16 LLMs all trained on public data seen in the exact same order and ranging in size from 70M to 12B parameters. We provide public access to 154 checkpoints for each one of the 16 models, alongside tools to download and reconstruct their exact training dataloaders for further study.

\textbf{Orca:} In \cite{mukherjee2023orca}, Mukherjee et al.  develop Orca, a 13-billion parameter model that learns to imitate the reasoning process of large foundation models. Orca learns from rich signals from GPT-4 including explanation traces; step-by-step thought processes; and other complex instructions, guided by teacher assistance from ChatGPT.

\textbf{StarCoder:} In \cite{li2023starcoder}, Li et al. introduced
StarCoder and StarCoderBase. They are 15.5B parameter models with 8K context length, infilling capabilities and fast large-batch inference enabled by multi-query attention.
StarCoderBase is trained on one trillion tokens sourced from The Stack, a large collection of permissively licensed GitHub repositories with inspection tools and an opt-out process.
They fine-tuned StarCoderBase on 35B Python tokens, resulting in the creation of StarCoder. They performed the most comprehensive
evaluation of Code LLMs to date and showed that StarCoderBase outperforms every open Code LLM that supports multiple programming languages and matches or outperforms the OpenAI code-cushman-001 model.

\textbf{KOSMOS:} In \cite{huang2023language}, Huang et al. introduced
KOSMOS-1, a Multimodal Large Language Model (MLLM) that can perceive general modalities, learn in context (i.e., few-shot), and follow instructions (i.e. zero-shot). Specifically, they trained KOSMOS-1 from scratch on web-scale multi-modal corpora, including arbitrarily interleaved text and images, image-caption
pairs, and text data.
Experimental results show that KOSMOS-1 achieves impressive performance on (i) language understanding, generation, and
even OCR-free NLP (directly fed with document images), (ii) perception-language tasks, including multimodal dialogue, image captioning, visual question answering, and (iii) vision tasks, such as image recognition with descriptions (specifying classification via text instructions).

\textbf{Gemini:} In \cite{team2023gemini}, Gemini team introduced a new family of multimodal models, that exhibit promising capabilities across image, audio, video, and text understanding. Gemini family includes three versions: Ultra for highly-complex tasks, Pro for enhanced performance and deployability at scale, and Nano for on-device applications.
Gemini architecture is built on top of Transformer decoders, and is trained to support 32k context length (via using efficient attention mechanisms).

\begin{figure*}[t]
\begin{center}
    \includegraphics [scale=0.55] {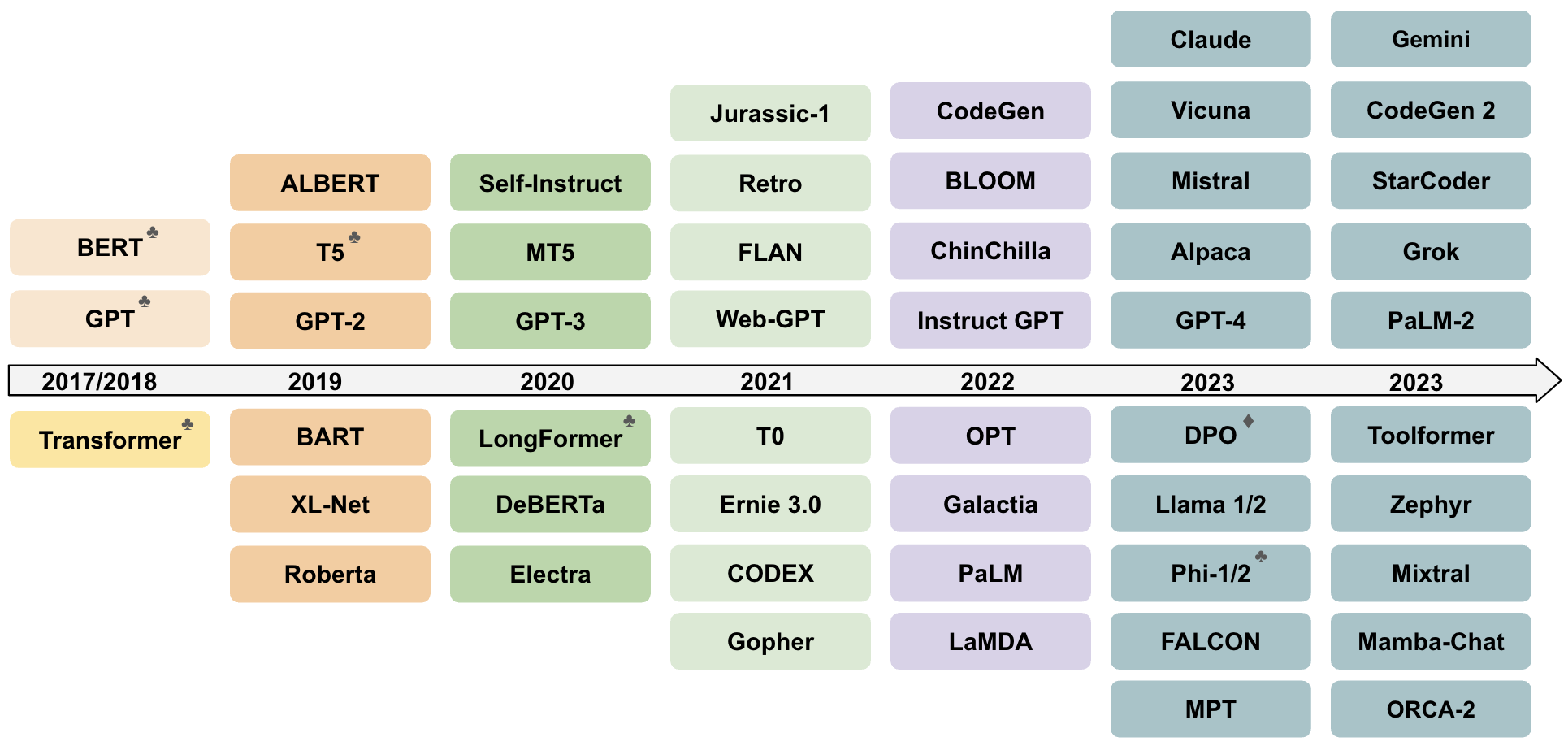}
\end{center}
  \caption{Timeline of some of the most representative LLM frameworks (so far). In addition to large language models with our \#parameters threshold, we included a few representative works, which pushed the limits of language models, and paved the way for their success (e.g. vanilla Transformer, BERT, GPT-1), as well as some small language models.
  $\clubsuit$ shows entities that serve not only as models but also as approaches. $\blacklozenge$ shows only approaches.}
  \label{fig:timeline}
\end{figure*}

Some of the other popular LLM frameworks (or techniques used for efficient developments of LLMs) includes 
Inner-Monologue \cite{huang2022inner},  
Megatron-Turing NLG \cite{smith2022using},  
LongFormer \cite{beltagy2020longformer},
OPT-IML \cite{iyer2022opt}, 
MeTaLM  \cite{hao2022language},
Dromedary \cite{sun2023principle}, 
Palmyra \cite{Palmyra}, 
Camel \cite{Camel}, 
Yalm \cite{yalm}, 
MPT \cite{team2023introducing},
ORCA-2 \cite{mitra2023orca},
Gorilla \cite{patil2023gorilla}, 
PAL \cite{gao2023pal}, 
Claude \cite{claude},
CodeGen 2 \cite{nijkamp2023codegen2},
Zephyr \cite{tunstall2023zephyr}, 
Grok \cite{Grok},  
Qwen \cite{bai2023qwen}, 
Mamba \cite{gu2023mamba}, 
Mixtral-8x7B \cite{mixtral},
DocLLM \cite{wang2023docllm},
DeepSeek-Coder \cite{guo2024deepseekcoder},
FuseLLM-7B \cite{wan2024knowledge},
TinyLlama-1.1B \cite{zhang2024tinyllama},
LLaMA-Pro-8B \cite{wu2024llama}.

Fig \ref{fig:timeline} provides an overview of some of the most representative LLM frameworks, and the relevant works that have contributed to the success of LLMs and helped to push the limits of LLMs.

\section{How LLMs Are Built}
\label{sec:LLM_built}
In this section, we first review the popular architectures used for LLMs, and then discuss data and modeling techniques ranging from data preparation, tokenization, to pre-training, instruction tuning, and alignment.

\begin{figure*}
    \centering
    \includegraphics[scale=0.8]{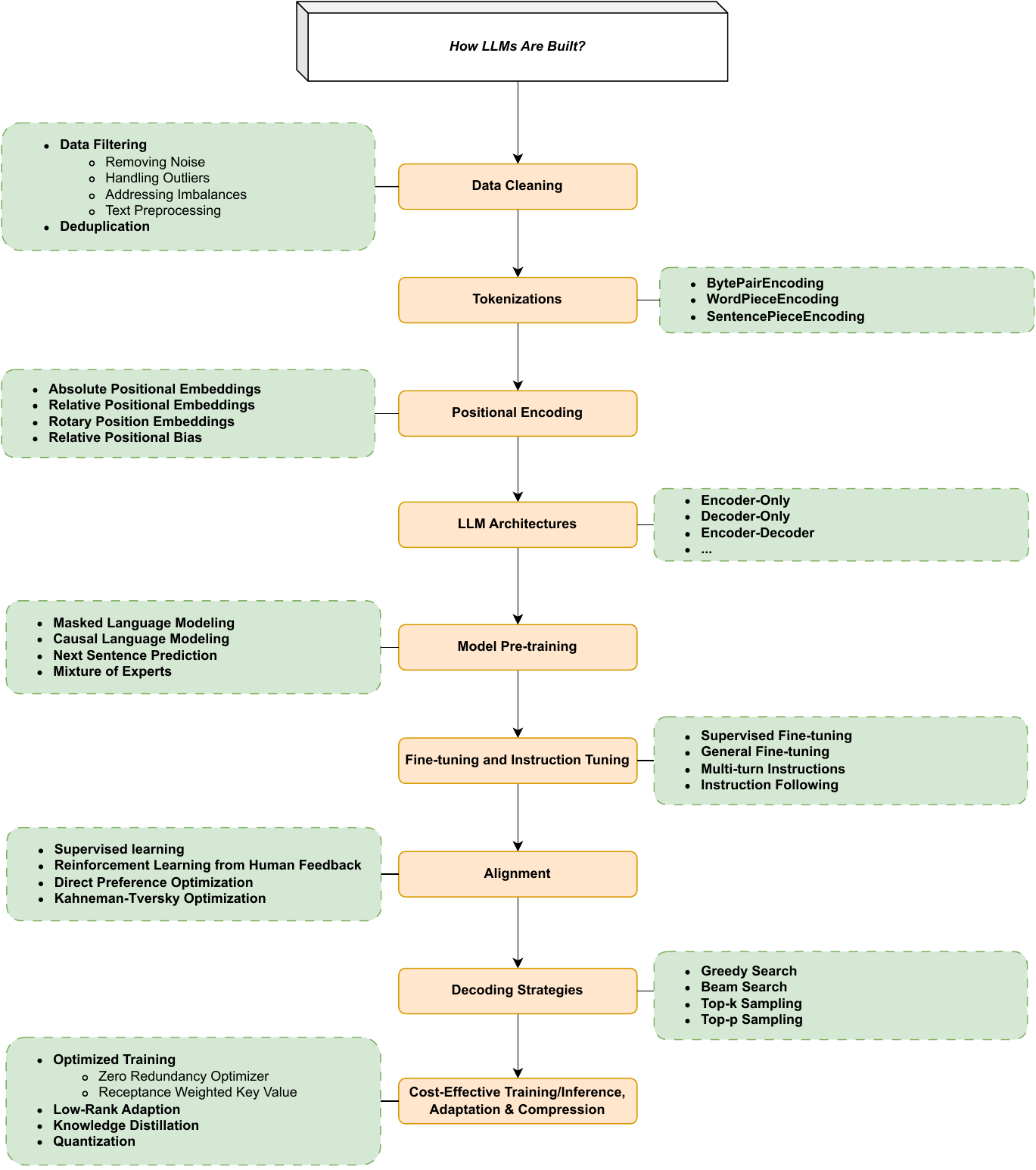}
    \caption{This figure shows different components of LLMs.}
    \label{fig:LLMs_components}
\end{figure*}

Once the model architecture is chosen, the major steps involved in training an LLM includes: data preparation (collection, cleaning, deduping, etc.), tokenization, model pre-training (in a self-supervised learning fashion), instruction tuning, and alignment. 
We will explain each of them in a separate subsection below. 
These steps are also illustrated in Fig \ref{fig:LLMs_components}.

\subsection{Dominant LLM Architectures}
The most widely used LLM architectures are encoder-only, decoder-only, and encoder-decoder. Most of them are based on Transformer (as the building block). Therefore we also review the Transformer architecture here.

\subsubsection{\textbf{Transformer}} 
\begin{figure}[h]
\begin{center}
    \includegraphics [scale=0.49] {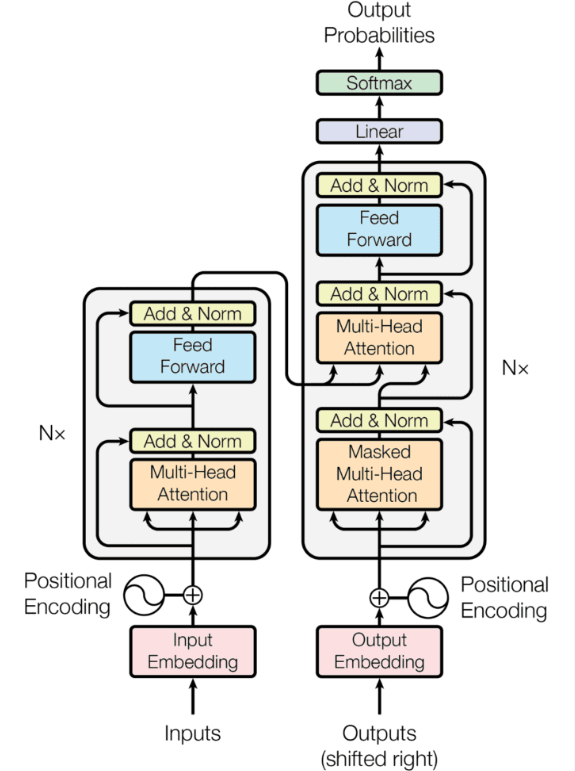}
\end{center}
  \caption{High-level overview of transformer work. Courtesy of \cite{vaswani2017attention}.}
  \label{fig:transformer}
\end{figure}

in  a ground-breaking work \cite{vaswani2017attention}, Vaswani et al. proposed the Transformer framework, which was originally designed for effective parallel computing using GPUs. 
The heart of Transformer is the (self-)attention mechanism, which can capture long-term contextual information much more effectively using GPUs than the recurrence and convolution mechanisms.
Fig \ref{fig:transformer} provides a high-level overview of transformer work. In this section we provide an overview of the main elements and variants, see \cite{vaswani2017attention,amatriain2023transformer} for more details.

The Transformer language model architecture, originally proposed for machine translation, consists of an encoder and a decoder. 
The encoder is composed of a stack of N = 6 identical Transformer layers. Each layer has two sub-layers. The first one is a multi-head self-attention layer, and the other one is a simple position-wise fully connected feed-forward network.
The decoder is composed of a stack of 6 identical layers. In addition to the two sub-layers in each encoder layer, the decoder has a third sub-layer, which performs multi-head attention over the output of the encoder stack.
The attention function can be described as mapping a query and a set of key-value pairs to an output, where the query, keys, values, and output are all vectors.
The output is computed as a weighted sum of the values, where the weight assigned to each value is computed by a compatibility function of the query with the corresponding key.
Instead of performing a single attention function with $d_{model}$ dimensional keys, values and queries,
it is found to be beneficial to linearly project the queries, keys and values $h$ with different, learned linear projections to $d_k$, $d_k$ and $d_v$ dimensions, respectively.
Positional encoding is incorporated to fuse information about the relative or absolute position of the tokens in the sequence.

\subsubsection{\textbf{Encoder-Only}} 
For this family, at each stage, the attention layers can access all the words in the initial sentence. The pre-training of these models usually consist of somehow corrupting a given sentence (for instance, by masking random words in it) and tasking the model with finding or reconstructing the initial sentence. Encoder models are great for tasks requiring an understanding of the full sequence, such as sentence classification, named entity recognition, and extractive question answering.
One prominent encoder only model is BERT (Bidirectional Encoder Representations from Transformers), proposed in \cite{devlin2018bert}.

\subsubsection{\textbf{Decoder-Only}} 
For these models, at each stage, for any word, the attention layers can only access the words positioned before that in the sentence. These models are also sometimes called auto-regressive models.
The pretraining of these models is usually formulated as predicting the next word (or token) in the sequence. The decoder-only models are best suited for tasks involving text generation. GPT models are prominent example of this model category.

\subsubsection{\textbf{Encoder-Decoder}}
These models use both encoder and decoder, and are sometimes called sequence-to-sequence models.  At each stage, the attention layers of the encoder can access all the words in the initial sentence, whereas the attention layers of the decoder only accesses the words positioned before a given word in the input.
These models are usually pre-trained using the objectives of encoder or decoder models, but usually involve something a bit more complex. For instance, some models are pretrained by replacing random spans of text (that can contain several words) with a single mask special word, and the objective is then to predict the text that this mask word replaces.
Encoder-decoder models are best suited for tasks about generating new sentences conditioned on a given input, such as summarization, translation, or generative question answering.

\subsection{Data Cleaning}
Data quality is crucial to the performance of language models trained on them. Data cleaning techniques such as filtering, deduplication, are shown to have a big impact on the model performance. 

\begin{figure}[h]
\begin{center}
    \includegraphics [scale=0.4] {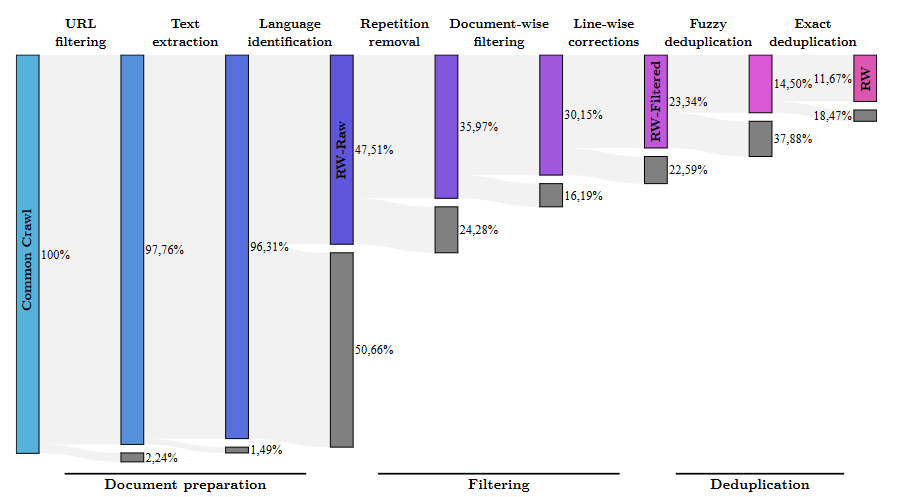}
\end{center}
  \caption{Subsequent stages of Macrodata Refinement remove nearly 90\% of the documents originally in CommonCrawl. Courtesy of \cite{penedo2023refinedweb}.}
\label{fig:Falcon40B_filtering}
\end{figure}

As an example, in \textbf{Falcon40B} \cite{penedo2023refinedweb}, Penedo et al. showed that properly filtered and deduplicated web data alone can lead to powerful models; even significantly outperforming models from the state-of-the-art trained on The Pile. Despite extensive filtering, they were able to obtain five trillion tokens from CommonCrawl. They also released an extract of 600 billion tokens from our REFINEDWEB dataset, and 1.3/7.5B parameters language models trained on it. \ref{fig:Falcon40B_filtering} shows the Refinement process of CommonCrawl data by this work.

\subsubsection{Data Filtering}
Data filtering aims to enhance the quality of training data and the effectiveness of the trained LLMs. 
Common data filtering techniques include:

\textbf{Removing Noise:} refers to eliminating irrelevant or noisy data that might impact the model's ability to generalize well. As an example, one can think of removing false information from the training data, to lower the chance of model generating false responses.
Two mainstream approaches for quality filtering includes: classifier-based, and heuristic-based frameworks.

\textbf{Handling Outliers:} Identifying and handling outliers or anomalies in the data to prevent them from disproportionately influencing the model.

\textbf{Addressing Imbalances:} Balancing the distribution of classes or categories in the dataset to avoid biases and ensure fair representation. This is specially useful for responsible model training and evaluation.

\textbf{Text Preprocessing:} Cleaning and standardizing text data by removing stop words, punctuation, or other elements that may not contribute significantly to the model's learning.

\textbf{Dealing with Ambiguities:} Resolving or excluding ambiguous or contradictory data that might confuse the model during training. This can help the model to provide more definite and reliable answers.

\subsubsection{Deduplication}
De-duplication refers to the process of removing duplicate instances or repeated occurrences of the same data in a dataset. 
Duplicate data points can introduce biases in the model training process and reduce the diversity, as the model may learn from the same examples multiple times, potentially leading to overfitting on those particular instances.
Some works \cite{hernandez2022scaling} have shown that de-duplication improves models' ability to generalize to new, unseen data.

The de-duplication process is particularly important when dealing with large datasets, as duplicates can unintentionally inflate the importance of certain patterns or characteristics. This is especially relevant in NLP tasks, where diverse and representative training data is crucial for building robust language models.

The specific de-duplication method can vary based on the nature of the data and the requirements of the particular language model being trained. It may involve comparing entire data points or specific features to identify and eliminate duplicates.
At the document level, existing works mainly rely on the overlap ratio of high-level features (e.g. n-grams overlap) between documents to detect duplicate samples.

\subsection{Tokenizations}
Tokenization referes to the process of converting a sequence of text into smaller parts, known as tokens.  
While the simplest tokenization tool simply chops text into tokens based on white space, most tokenization tools rely on a word dictionary. 
However, out-of-vocabulary (OOV) is a problem in this case because the tokenizer only knows words in its dictionary.
To increase the coverage of dictionaries, popular tokenizers used for LLMs are based on sub-words, which can be combined to form a large number of words, including the words unseen in training data or words in different languages.
In what follows, we describe three popular tokenizers.

\subsubsection{\textbf{BytePairEncoding}}
\textit{BytePairEncoding} is originally a type of data compression algorithm that uses frequent patterns at byte level to compress the data. By definition, this algorithm mainly tries to keep the frequent words in their original form and break down ones that are not common. This simple paradigm keeps the vocabulary not very large, but also good enough to represent common words at the same time. Also morphological forms of the frequent words can be represented very well if suffix or prefix is also commonly presented in the training data of the algorithm.

\subsubsection{\textbf{WordPieceEncoding}}
This algorithm is mainly used for very well-known models such as BERT and Electra. At the beginning of training, the algorithm takes all the alphabet from the training data to make sure that nothing will be left as UNK or \textit{unknown} from the training dataset. This case happens when the model is given an input that can not be tokenized by the tokenizer. It mostly happens in cases where some characters are not tokenizable by it. Similar to BytePairEncoding, it tries to maximize the likelihood of putting all the tokens in vocabulary based on their frequency.

\subsubsection{\textbf{SentencePieceEncoding}}
Although both tokenizers described before are strong and have many advantages compared to white-space tokenization, they still take assumption of words being always separated by white-space as granted. This assumption is not always true, in fact in some languages, words can be corrupted by many noisy elements such as unwanted spaces or even invented words. SentencePieceEncoding tries to address this issue.

\subsection{\textbf{Positional Encoding}}

\subsubsection{\textbf{Absolute Positional Embeddings}} (APE) \cite{vaswani2017attention} has been used in the original Transformer model to preserve the information of sequence order. Therefore, the positional information of words is added to the input embeddings at the bottom of both the encoder and decoder stacks. There are various options for positional encodings, either learned or fixed. In the vanilla Transformer, sine and cosine functions are employed for this purpose. The main drawback of using APE in Transformers is the restriction to a certain number of tokens. Additionally,  APE fails to account for the relative distances between tokens.

\subsubsection{\textbf{Relative Positional Embeddings}} (RPE) \cite{shaw2018self} involves extending self-attention to take into account the pairwise links between input elements. RPE is added to the model at two levels: first as an additional component to the keys, and subsequently as a sub-component of the values matrix.  This approach looks at the input as a fully-connected graph with labels and directed edges. In the case of linear sequences, edges can capture information about the relative position differences between input elements.  A clipping distance, represented as k $2 \leq k \leq n - 4$, specifies the maximum limit on relative locations. This allows the model to make reasonable predictions for sequence lengths that are not part of the training data.

\subsubsection{\textbf{Rotary Position Embeddings}}
Rotary Positional Embedding (RoPE) \cite{rope-paper} tackles problems with existing approaches. 
Learned absolute positional encodings can lack generalizability and meaningfulness, particularly when sentences are short. Moreover, current methods like T5's positional embedding face challenges with constructing a full attention matrix between positions.  RoPE uses a rotation matrix to encode the absolute position of words and simultaneously includes explicit relative position details in self-attention. RoPE brings useful features like flexibility with sentence lengths, a decrease in word dependency as relative distances increase, and the ability to improve linear self-attention with relative position encoding. GPT-NeoX-20B, PaLM, CODEGEN, and LLaMA are among models that take advantage of RoPE in their architectures.

\subsubsection{\textbf{Relative Positional Bias}}
The concept behind this type of positional embedding is to facilitate extrapolation during inference for sequences longer than those encountered in training.
In \cite{press2021train} Press et al.  proposed Attention with Linear Biases (ALiBi). Instead of simply adding positional embeddings to word embeddings, they introduced a bias to the attention scores of query-key pairs, imposing a penalty proportional to their distance. In the BLOOM model, ALiBi is leveraged.

\begin{figure*}[t]
    \centering
    \begin{subfigure}[t]{0.35\textwidth}
        \includegraphics[width=0.5 \textwidth]{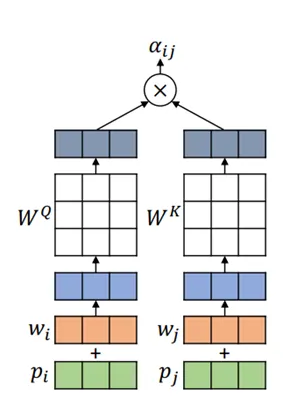}
        \caption{Absolute Positional Embeddings \cite{ke2020rethinking}}
        \label{fig:APE}        
    \end{subfigure}%
    \begin{subfigure}[t]{0.35\textwidth}
        \includegraphics[width=0.9\textwidth]{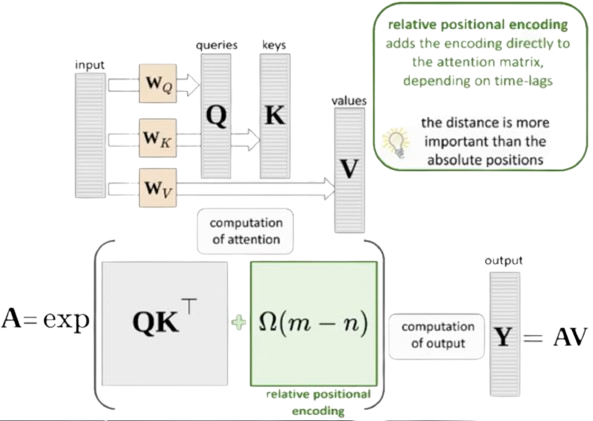}
        \caption{Relative Positional Embeddings}
        \label{fig:RPE}        
    \end{subfigure}%
        \\
    \begin{subfigure}[t]{0.35\textwidth}
        \includegraphics[width=0.9 \textwidth]{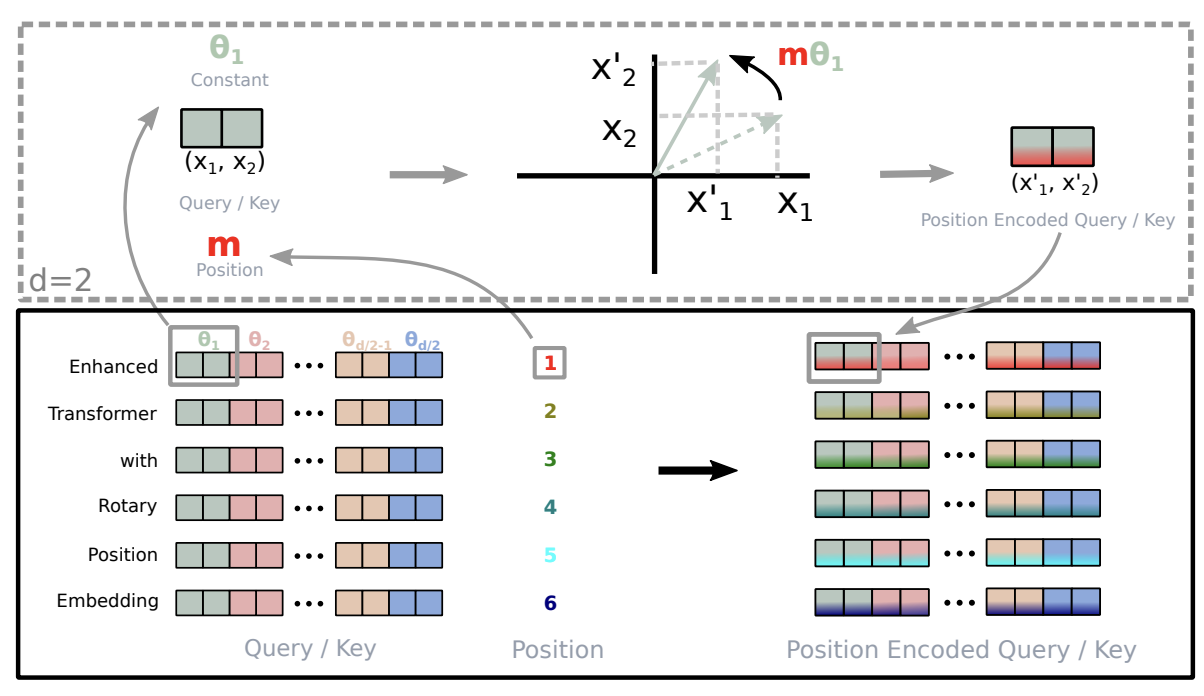}
        \caption{Rotary Positional Embedding \cite{rope-paper}}
        \label{fig:RoPE}        
    \end{subfigure}%
    \begin{subfigure}[t]{0.35\textwidth}
        \includegraphics[width=0.7 \textwidth]{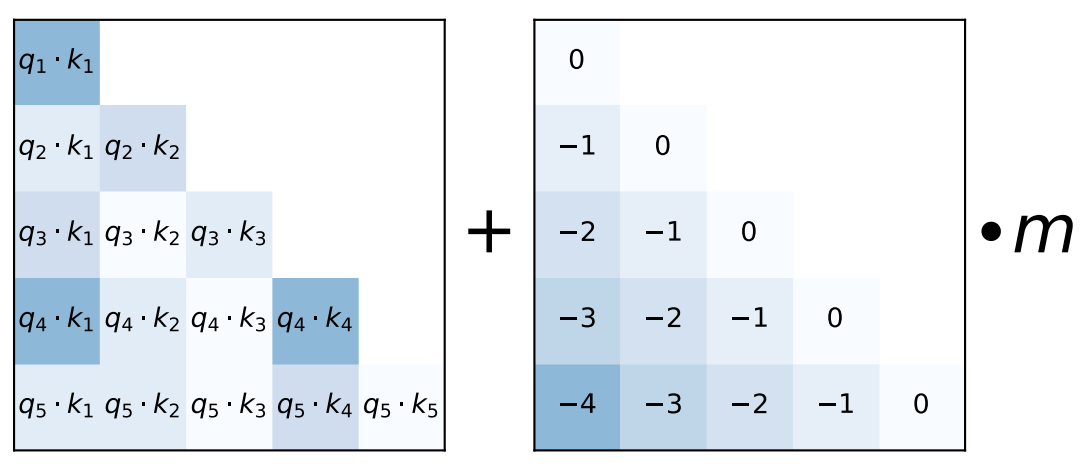}
        \caption{Relative Positional Bias \cite{press2021train}}
        \label{fig:RPB}        
    \end{subfigure}  
    \caption{Various positional encodings are employed in LLMs.}
    \label{fig:positional_encodings}
\end{figure*}

\subsection{Model Pre-training}\label{sub:pre-training}
Pre-training is the very first step in large language model training pipeline, and it helps LLMs to acquire fundamental language understanding capabilities, which can be useful in a wide range of language related tasks. 
During pre-training, the LLM is trained on a massive amount of (usually) unlabeled texts, usually in a self-supervised manner.
There are different approaches used for pre-training like next sentence prediction \cite{devlin2018bert}, two most common ones include, next token prediction (autoregressive language modeling), and masked language modeling.

In \textbf{Autoregressive Language Modeling} framework, given a sequence of $n$ tokens {$x_1$, ..., $x_n$}, the model tries to predict next token $x_{n+1}$ (and sometimes next sequence of tokens) in an auto-regressive fashion.
One popular loss function in this case is the log-likelihood of predicted tokens as shown in Eq \ref{eq:lm}
\begin{equation}
\mathcal{L}_{ALM}(x)= \sum_{i=1}^{N} p(x_{i+n} | x_i, ..., x_{i+n-1})
    \label{eq:lm}
\end{equation}
Given the auto-regressive nature of this framework, the decoder-only models are naturally better suited to learn how to accomplish these task.

In \textbf{Masked Language Modeling}, some words are masked in a sequence and the model is trained to predict the masked words based on the surrounding context. Sometimes people refer to this approach as denoising autoencoding, too.
If we denote the masked/corrupted samples in the sequence $x$, as $\tilde{x}$, then the training objective of this approach can be written as:
\begin{equation}
\mathcal{L}_{MLM}(x)= \sum_{i=1}^{N} p( \tilde{x} | x \backslash  \tilde{x}  )
    \label{eq:lm}
\end{equation}

\begin{figure}[h]
\begin{center}
    \includegraphics [scale=0.44] {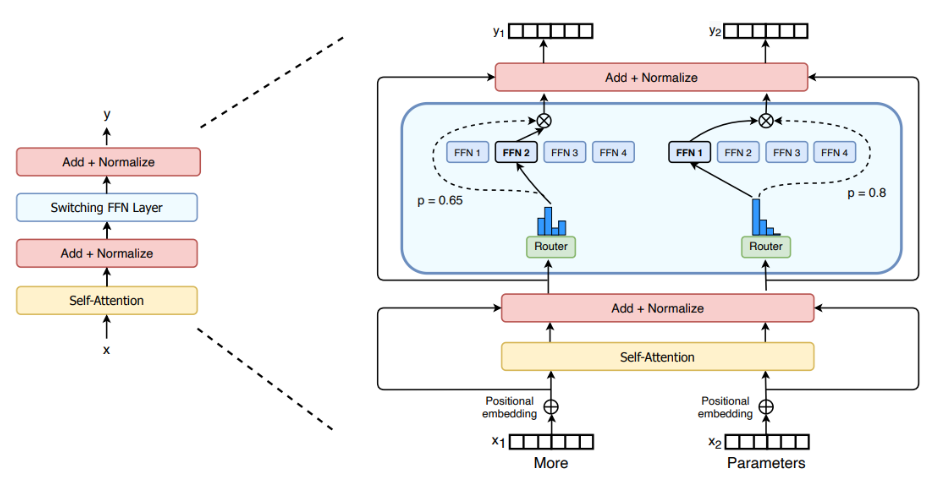}
\end{center}
  \caption{: Illustration of a Switch Transformer encoder block. They replaced the dense feed forward network (FFN) layer present in the Transformer with a sparse Switch FFN layer (light blue). . Courtesy of \cite{fedus2022switch}.}
\label{fig:moe}
\end{figure}

And more recently, \textbf{Mixture of Experts (MoE)} \cite{shazeer2017outrageously, fedus2022switch} have become very popular in LLM space too. MoEs enable models to be pre-trained with much less compute, which means one can dramatically scale up the model or dataset size with the same compute budget as a dense model. 
MoE consists of two main elements:
\textbf{Sparse MoE layers}, which are used instead of dense feed-forward network (FFN) layers, and have a certain number of “experts” (e.g. 8), in which each expert is a neural network. In practice, the experts are FFNs, but they can also be more complex networks.
\textbf{A gate network or router}, that determines which tokens are sent to which expert. It is worth noting that, one can send a token to more than one expert. How to route a token to an expert is one of the big decisions when working with MoEs - the router is composed of learned parameters and is pretrained at the same time as the rest of the network.
Fig \ref{fig:moe} provides an illustration of a Switch Transformer encoder block, which are used in MoE.

\subsection{Fine-tuning and Instruction Tuning}
Early language models such as BERT trained using self-supervision as explained in section \ref{sub:pre-training} were not able to perform specific tasks. In order for the foundation model to be useful it needed to be fine-tuned to a specific task with labeled data (so-called supervised fine-tuning or SFT for short). For example, in the original BERT paper \cite{devlin2018bert}, the model was fine-tuned to 11 different tasks. While more recent LLMs no longer require fine-tuning to be used, they can still benefit from task or data-specific fine-tuning. For example, OpenAI reports that the much smaller GPT-3.5 Turbo model can outperform GPT-4 when fine-tuned with task specific data \footnote{https://platform.openai.com/docs/guides/fine-tuning}.

Fine-tuning does not need to be performed to a single task though, and there are different approaches to multi-task fine-tuning (see e.g. Mahabi et al. \cite{mahabadi2021parameterefficient}). Fine-tuning to one or more tasks is known to improve results and reduce the complexity of prompt engineering, and it can serve as an alternative to retrieval augmented generation. Furthermore, there are other reasons why it might be advisable to fine-tune. For example, one might want to fine-tune to expose the model to new or proprietary data that it has not been exposed to during pre-training.

An important reason to fine-tune LLMs is to align the responses to the expectations humans will have when providing instructions through prompts. This is the so-called \textbf{instruction tuning} \cite{zhang2023instruction}. We dive into the details of how to design and engineer prompts in section \ref{sub:Prompts}, but in the context of instruction tuning, it is important to understand that the instruction is a prompt that specifies the task that the LLM should accomplish. Instruction tuning datasets such as Natural Instructions \cite{mishra2021cross} include not only the task definition but other components such as positive/negative examples or things to avoid.

The specific approach and instruction datasets used to instruction-tune an LLM varies, but, generally speaking, instruction tuned models outperform their original foundation models they are based on. For example, InstructGPT \cite{ouyang2022training} outperforms GPT-3 on most benchmarks. The same is true for Alpaca \cite{taori2023alpaca} when compared to LLaMA.




\textbf{Self-Instruct} \cite{wang2022self}, proposed by Wang et al. is also a popular approach along this line, in which they introduced a framework for improving the instruction-following capabilities of pre-trained language models by bootstrapping their own generations. Their pipeline generates instructions, input, and output samples from a language model, then filters invalid or similar ones before using them to fine tune the original model.

\subsection{Alignment}

AI Alignment is the process of steering AI systems towards human goals, preferences, and principles. LLMs, pre-trained for word prediction, often exhibit unintended behaviors. For example, they might generate contents that are toxic, harmful, misleading and biased.

Instruction tuning, discussed above, gets LLMs a step closer to being aligned. However, in many cases, it is important to include further steps to improve the alignment of the model and avoid unintended behaviors \footnote{According to very recent research by Ethayarajh et al. \cite{KTO}, further alignment besides SFT mainly improves models of at least 7B parameters. For smaller models, SFT is sufficient.}. We review the most popular approaches to alignment in this subsection.



\textbf{RLHF} (reinforcement learning from human feedback) and  \textbf{RLAIF} (reinforcement learning from AI feedback) are two popular approaches. RLHF uses a reward model to learn alignment from human feedback. This reward model, after being tuned, is able to rate different outputs and score them according to their alignment preferences given by humans. The reward model gives feedback to the original LLM and this feedback is used to tune the LLM further \cite{christiano2017deep}. Reinforcement learning from AI feedback on the other hand, directly connects a pretrained and well-aligned model to the LLM and helps it to learn from larger and more aligned models \cite{lee2023rlaif}.

\begin{figure}[h]
\begin{center}
    \includegraphics [scale=0.38] {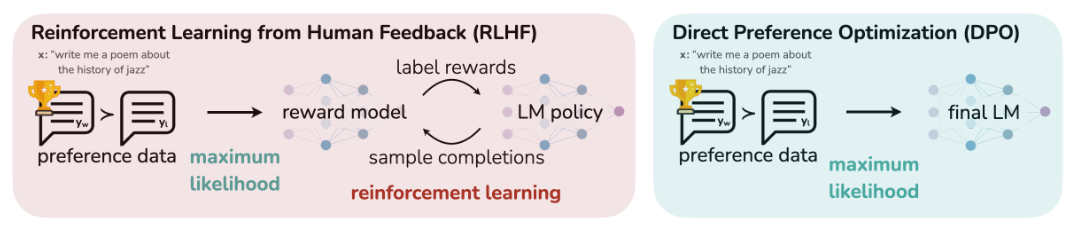}
\end{center}
  \caption{DPO optimizes for human preferences while avoiding reinforcement learning. Existing methods for fine-tuning language models with human feedback first fit a reward model to a dataset of prompts and human preferences over pairs of responses, and then use RL to find a policy that maximizes the learned reward. In contrast, DPO directly optimizes for the policy best satisfying the preferences with a simple classification objective, without an explicit reward function or RL. Courtesy of \cite{rafailov2023direct}.}
\label{fig:dpo}
\end{figure}

In another recent work (known as \textbf{DPO}) \cite{rafailov2023direct}, Rafailov et al. discussed that RLHF is a complex and often unstable procedure, and tried to address this with a new approach. They leveraged a mapping between reward functions and optimal policies to show that this constrained reward maximization problem can be optimized exactly with a single stage of policy training, essentially solving a classification problem on the human preference data. The resulting algorithm, which they called Direct Preference Optimization (DPO), is stable, performant, and computationally lightweight, eliminating the need for fitting a reward model, sampling from the LM during fine-tuning, or performing significant
hyperparameter tuning.
They observed that fine-tuning with DPO exceeds RLHF’s ability to control sentiment of generations and improves response quality in summarization.
Fig \ref{fig:dpo} shows the high-level comparison between DPO vs RLHF.

\begin{figure}[h]
\begin{center}
    \includegraphics [scale=0.5] {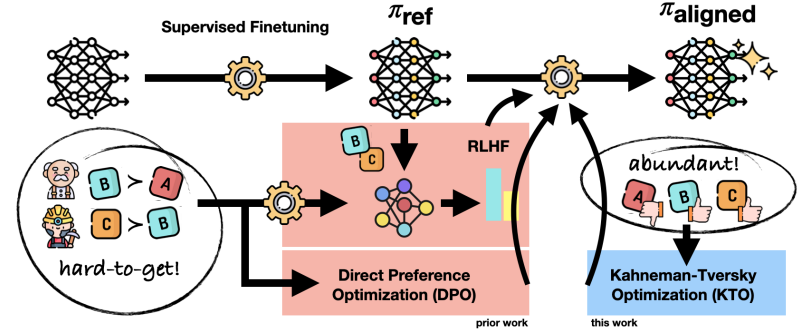}
\end{center}
  \caption{LLM alignment involves supervised finetuning followed by optimizing a human-centered loss (HALO). However, the paired preferences that existing approaches need are hard-to-obtain. In contrast, KTO uses a far more abundant kind of data, making it much easier to use in the real world. Courtesy of \cite{KTO}.}
\label{fig:KTO}
\end{figure}

Even more recently Ethayarajh et al. proposed a new alignment approach called the Kahneman-Tversky Optimization (KTO) \cite{KTO}. Unlike existing state-of-the-art approaches,  KTO does not require paired preference data ($x$, $y_w$, $y_l$), and it only needs (x,y) and knowledge of whether $y$ is desirable or undesirable. KTO-aligned models are shown to be good or better than DPO-aligned models at scales from 1B to 30B, despite not using paired preferences.
KTO is also far easier to use in the real world than preference optimization methods, as the kind of data it needs is far more abundant. As an example, every retail company has a lot of customer interaction data and whether that interaction was successful (e.g., purchase made) or unsuccessful (e.g., no purchase made).
However,  they have little to no counterfactual data (i.e., what would have made an unsuccessful customer interaction $y_l$
into a successful one $y_w$).
Fig \ref{fig:KTO} shows a high-level comparison between KTO and other alignment approaches discussed above.

\subsection{Decoding Strategies}
Decoding refers to the process of text generation using pre-trained LLMs.
Given an input prompt, the tokenizer translates each token in the input text into a corresponding token ID. 
Then, the language model uses these token IDs as input and predicts the next most likely token (or a sequence of tokens). 
Finally, the model generates logits, which are converted to probabilities using a softmax function.
Different decoding strategies have been proposed. Some of the most popular ones are greedy search, beam search, as well as different sample techniques such as top-K, top-P (Nucleus sampling).

\subsubsection{\textbf{Greedy Search}}
Greedy search takes the most probable token at each step as the next token in the sequence, discarding all other potential options. As you can imagine, this is a simple approach and can loose a lot of temporal consistency and coherency.
It only considers the most probable token at each step, without considering the overall effect on the sequence. This property makes it fast, but it also means that it can miss out on better sequences that might have appeared with slightly less probable next tokens.

\subsubsection{\textbf{Beam Search}}
Unlike greedy search that only considers the next most probable token, beam search takes into account the \textbf{N} most likely tokens, where \textbf{N} denotes the number of beams. This procedure is repeated until a predefined maximum sequence length is reached or an end-of-sequence token appears. At this point, the sequence of tokens (AKA “beam”) with the highest overall score is chosen as the output.
For example for beam size of 2 and maximum length of 5, the beam search needs to keep track of $2^5=32$ possible sequences. So it is more computationally intensive than greedy search.

\subsubsection{\textbf{Top-k Sampling}}
Top-k sampling is a technique that uses the probability distribution generated by the language model to select a token randomly from the k most likely options. 

Suppose we have 6 tokens (A, B, C, D, E, F) and k=2, and P(A)= 30\%, and P(B)= 20\%, P(C)= P(D)= P(E)= P(F)= 12.5\%. In top-k sampling, tokens C, D, E, F are disregarded, and the model outputs A 60\% of the time, and B, 40\% of the time.
This approach ensures that we prioritize the most probable tokens while introducing an element of randomness in the selection process.

The randomness is usually introduced via the concept of temperature. The temperature T is a parameter that ranges from 0 to 1, which affects the probabilities generated by the softmax function, making the most likely tokens more influential. In practice, it simply consists of dividing the input logits by temperature value: 
\begin{equation}
    softmax(x_i)= \frac{e^{x_i/T}}{\sum_j e^{x_j/T}}
\end{equation}

A low temperature setting significantly alters the probability distribution (and is commonly used in text generation to control the level of “creativity” in the generated output), while a large temperature prioritizes the tokens with higher probabilities. 
Top-k is a creative way of sampling, and can be used along with beam search.
The sequence chosen by top-k sampling may not be the sequence with highest probability in beam search. But it’s important to remember that highest scores do not always lead to more realistic or meaningful sequences.

\subsubsection{\textbf{Top-p Sampling}}
Top-p sampling, also known as Nucleus sampling, takes a slightly different approach from top-k sampling. Instead of selecting the top k most probable tokens, nucleus sampling chooses a cutoff value p such that the sum of the probabilities of the selected tokens exceeds p. This forms a “nucleus” of tokens from which to randomly choose the next token. In other words, in top-p sampling the language model examines the most probable tokens in descending order and keeps adding them to the list until the sum of probabilities surpasses the threshold p.
As you can imagine, this could be better specially for scenarios in which top-k tokens do not have a large probability mass.
Unlike top-k sampling, the number of tokens included in the nucleus sampling is not fixed. This variability often results in a more diverse and creative output, making nucleus sampling popular for text generation related tasks.

\subsection{Cost-Effective Training/Inference/Adaptation/Compression}\label{sub:cost-effective}
In this part, we review some of the popular approaches used for more cost-friendly (and compute-friendly) training and usage of LLMs.

\subsubsection{\textbf{Optimized Training}}
There are many frameworks developed for optimized training of LLMs, here we introduce some of the prominent ones.

\textbf{ZeRO: } In \cite{rajbhandari2020zero}, Rajbhandari et al. developed a novel solution, Zero Redundancy Optimizer (ZeRO), to optimize memory, vastly improving training speed of LLMs while increasing the model size that can be efficiently trained. ZeRO eliminates memory redundancies in data- and model-parallel training while retaining low communication volume and high computational granularity, allowing one to scale the model size proportional to the number of devices with sustained high efficiency. 

\begin{figure}[h]

\begin{center}
    \includegraphics [scale=0.7] {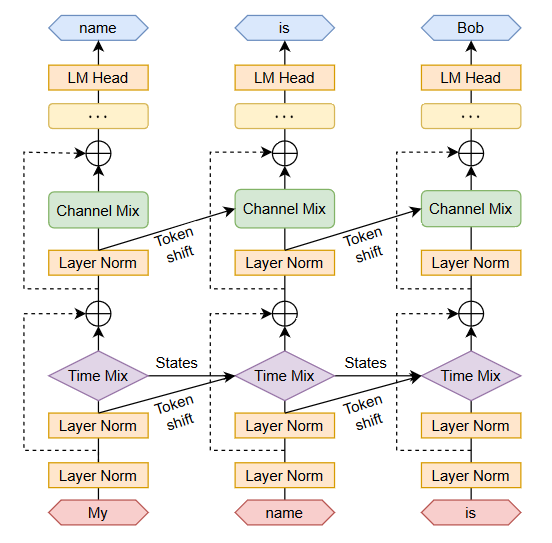}
\end{center}
  \caption{RWKV architecture. Courtesy of \cite{peng2023rwkv}.}
\label{fig:rwkv}
\end{figure}

\begin{figure}[h]
\begin{center}
    \includegraphics [scale=0.6] {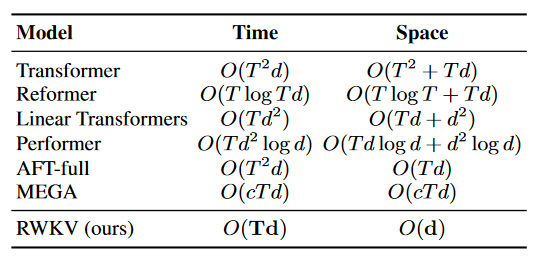}
\end{center}
  \caption{Time Complexity comparison of RWKV with different Transformers. Here T denotes the sequence length, d the feature dimension, and c is MEGA’s chunk size of quadratic attention. Courtesy of \cite{peng2023rwkv}.}
\label{fig:rwkv_time}
\end{figure}

\textbf{RWKV:} In \cite{peng2023rwkv}, Peng et al. proposed a novel model architecture, Receptance Weighted Key Value (RWKV), that combines the efficient parallelizable training of Transformers with the efficient inference of RNNs. Their approach leverages a linear attention mechanism and allows them to formulate the model as either a Transformer or an RNN, which parallelizes computations during training and maintains constant computational and memory complexity during inference, leading to the first non-transformer architecture to be scaled to tens of billions of parameters.
RWKV architecture is shown in Fig \ref{fig:rwkv}.
The Time Complexity comparison of RWKV with different Transformers are provided in Fig \ref{fig:rwkv_time}.

\subsubsection{\textbf{Low-Rank Adaption (LoRA)}}
Low-Rank Adaptation is a popular and lightweight training technique that significantly reduces the number of trainable parameters, and is based on a crucial insight that the difference between the fine-tuned weights for a specialized task and the initial pre-trained weights often exhibits “low intrinsic rank” - meaning that it can be approximated well by a low rank matrix \cite{hu2021lora}.
Training with LoRA is much faster, memory-efficient, and produces smaller model weights (a few hundred MBs), that are easier to store and share.
One property of low-rank matrices is that they can be represented as the product of two smaller matrices. This realization leads to the hypothesis that this delta between fine-tuned weights and initial pre-trained weights can be represented as the matrix product of two much smaller matrices. By focusing on updating these two smaller matrices rather than the entire original weight matrix, computational efficiency can be substantially improved. 

\begin{figure}
    \centering
    \includegraphics[scale=0.6]{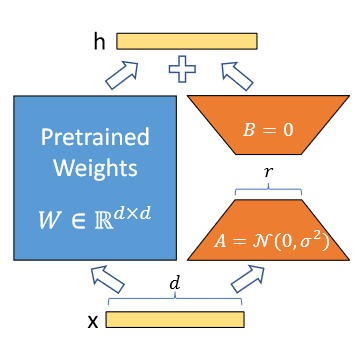}
    \caption{An illustration of LoRA reparametrizan. Only $A$ and $B$ trained during this process. Courtesy of \cite{hu2021lora}.}
    \label{fig:lora}
\end{figure}

Specifically, for a pre-trained weight matrix $W_0 \in R^{d \times k}$, LoRA constrains its update by representing the latter with a low-rank decomposition $W_0 + \Delta W = W_0 + BA$, where $B \in R^{d \times r}$ , $A \in R^{r \times k}$, and the rank $r \ll min(d, k)$.
During training, $W_0$ is frozen and does not receive gradient updates, while $A$ and $B$ contain trainable parameters. 
It is worth mentioning that both $W_0$ and $\Delta W = BA$ are multiplied with the same input, and their respective
output vectors are summed coordinate-wise. For $h = W_0 x$, their modified forward pass yields:
$h = W_0 x + \Delta W x = W_0 x + BA x$.
Usually a random Gaussian initialization is used for $A$, and zero initialization for $B$, so $\Delta W = BA$ is zero at the beginning of training. 
They then scale $\Delta Wx$ by $\alpha r$, where $\alpha$ is a constant in r.
This reparametrization is illustrated in Figure \ref{fig:lora}

It is worth mentioning that  LoRA can be applied to a subset of weight matrices in a neural network to reduce the number of trainable parameters. 
In the Transformer architecture, there are four weight matrices in the self-attention module ($W_q$ , $W_k$, $W_v$ , $W_o$), and two in the MLP module. Most of the time, LoRA is focused on adapting the attention weights only for downstream tasks, and freezes the MLP modules,  so they are not trained in downstream tasks both for simplicity and parameter-efficiency.

\subsubsection{\textbf{Knowledge Distillation}}
\begin{figure}[h]
    \centering
    \includegraphics[scale=0.33]{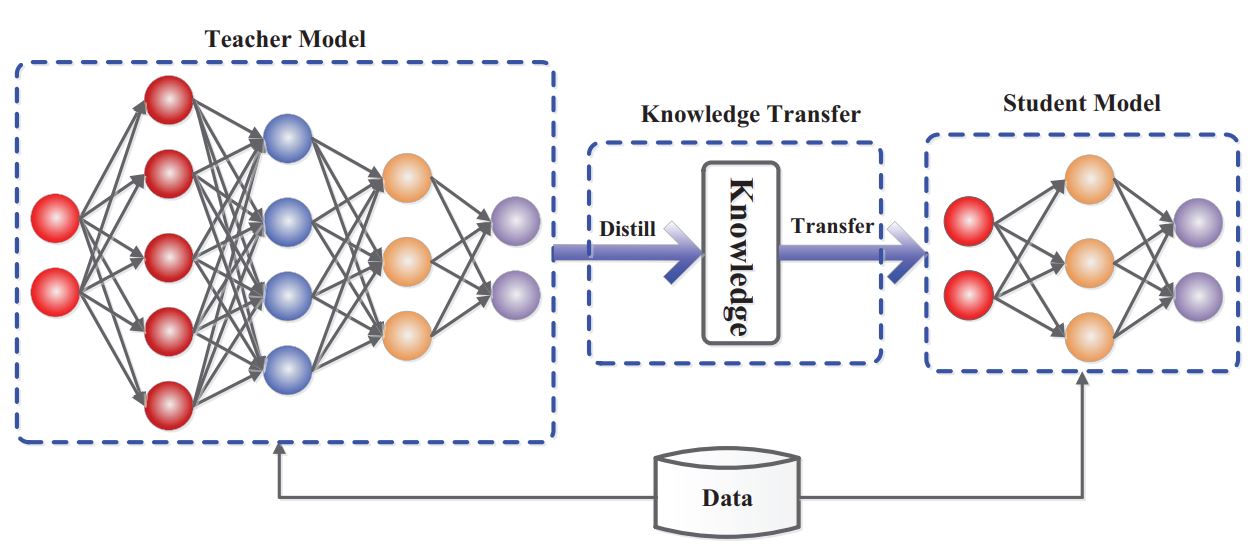}
    \caption{A generic knowledge distillation framework with student and teacher (Courtesy of \cite{gou2021knowledge}).}
    \label{fig:distill}
\end{figure}

Knowledge distillation is the process of learning from a larger model \cite{hinton2015distilling}. Earlier days of best-performing models release have proven that this approach is very useful even if it is used in an API distillation approach. It is also referred to as an approach to distill the knowledge of not a single model but in fact multiple models into a smaller one. Creating smaller models by this approach yields smaller model sizes that can be used even on edge devices. Knowledge distillation as shown in Fig \ref{fig:distill}, illustrates a general setup of this training scheme.

Knowledge can be transferred by different forms of learning: response distillation, feature distillation, and API distillation. Response distillation is concerned only with the outputs of the teacher model and tries to teach the student model how to exactly or at least similarly perform (in the sense of prediction) as the teacher. Feature distillation not only uses the last layer but also intermediate layers as well to create a better inner representation for the student model. This helps the smaller model to have a similar representation as the teacher model.

API distillation is the process of using an API (typically from an LLM provider such as OpenAI) to train smaller models. In the case of LLMs, it is used to train the model from the direct output of the larger model which makes it very similar to response distillation. Many concerns are raised by this type of distillation because in cases where the model itself is not openly available, a (usually) paid API is exposed for end users. On the other hand, while users pay for each call, how to use the predictions is limited, for example, OpenAI prohibits usage of its API to create LLMs that later will be used to compete with it. The main value in such case is training data. 

\subsubsection{\textbf{Quantization}}
deep learning in its core, is a set of mathematical functions applied to matrices, with a specific precision for model weights. Reducing the precision of the weights can be used to reduce the size of the model and also make it faster. As an example, Float-32 operations compared to Int-8 operations are slower. This process, which is called quantization, can be applied in different phases.
Main approaches for model quantization can be categorized as: post training quantization and quantization-aware training.  Post-training quantization is concerned with quantized trained models in two well-known methods: dynamic and static. Dynamic post-training quantization computes the range of quantization on the runtime and is slower compared to static. Quantization-aware training adds quantization criteria into training, and a quantized model is trained and optimized during training process. This approach ensures that the end model will have good performance and also does not need to be quantized after training.

\section{How LLMs Are Used and Augmented}
\label{sec:LLM_used}
Once the LLMs are trained, we can use them to generate desired outputs for a variety of tasks.
LLMs can be used directly through basic prompting. However, in order to exploit their full potential or to address some of the shortcomings, we need to augment the models through some external means. 
In this section we first provide a brief overview of the main shortcoming of LLMs, with a deeper look at the issue of hallucination. We then describe how prompting and some augmentation approaches can not only address those limitations but also be used to augment the capabilities of LLMs going as far as turning an LLM into a full-blown AI agent with the ability to interface with the external world.

\begin{figure*}
    \centering
    \includegraphics[scale=0.44]{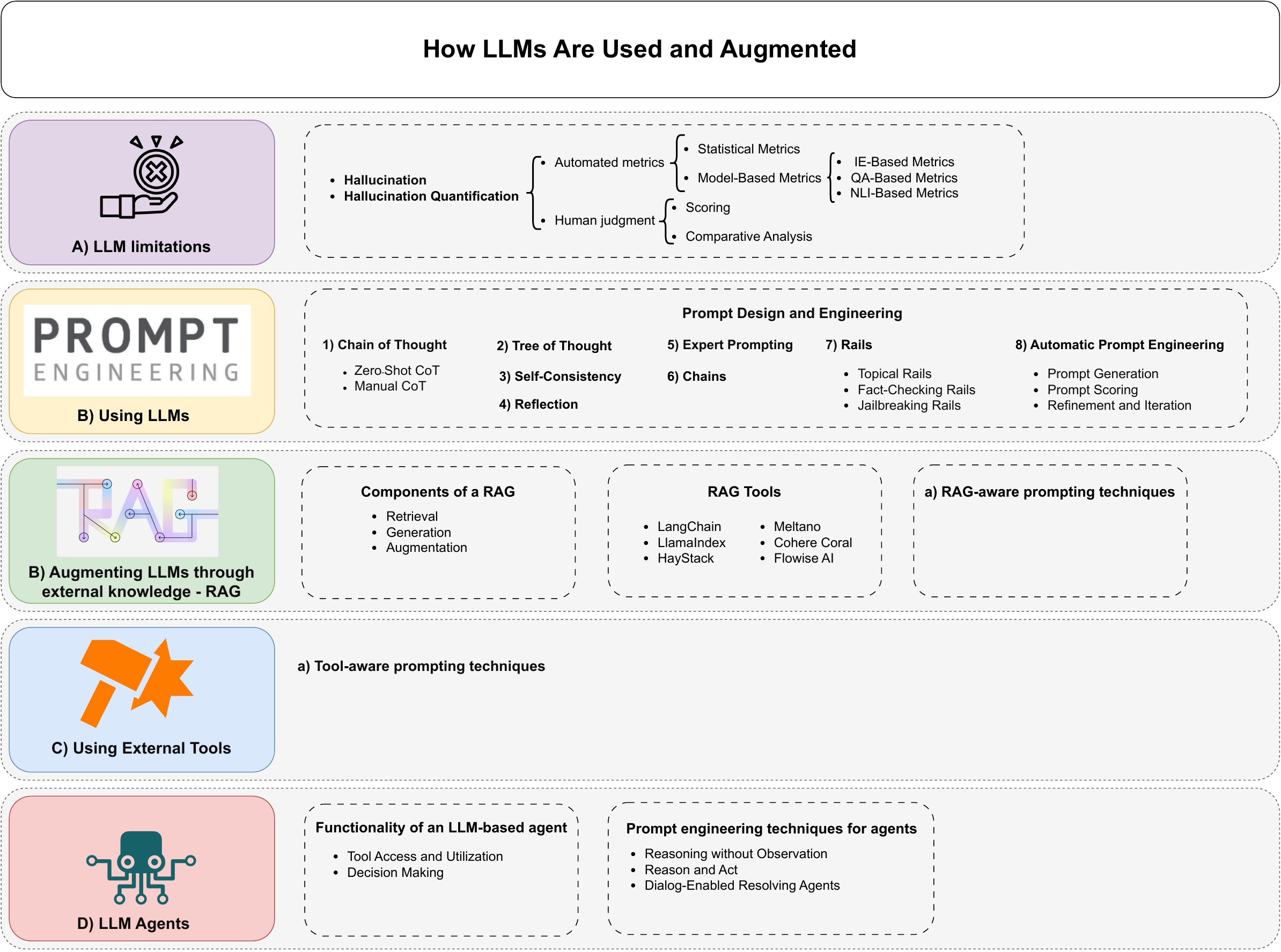}
    \caption{How LLMs Are Used and Augmented.}
    \label{fig:how_llms_are_used}
\end{figure*}

\subsection{LLM limitations}

It is important to remember that LLMs are trained to predict a token. While fine-tuning and alignment improves their performance and adds different dimensions to their abilities, there are still some important limitations that come up, particularly if they are used naively. Some of them include the following:

\begin{itemize}
    \item They don't have state/memory. LLMs on their own cannot remember even what was sent to them in the previous prompt. That is an important limitation for many of the use cases that require some form of state.
    \item They are stochastic/probabilistic. If you send the same prompt to an LLM several times, you are likely to get different responses. While there are parameters, and in particular the temperature, to limit the variability in the response, this is an inherent property of their training that can create issues.
    \item They have stale information and, on their own, don't have access to external data. An LLM on its own does not even know about the current time or day and does not have access to any information that was not present in its training set.
    \item They are generally very large. This means that many costly GPU machines are needed for training and serving. In some cases, largest models have poor SLAs, particularly in terms of latency.
    \item They hallucinate. LLMs do not have a notion of "truth" and they have usually been trained on a mix of good and bad content. They can produce very plausible but untruthful answers.
\end{itemize}

While the previous limitations can all become important for some applications, it is worth for us to dive a bit into the last one, hallucinations, since it has gathered a lot of interest over the past few months and it has also sparked many of the prompt approaches and LLM augmentation methods we will later describe.

\textbf{Hallucination:}
In the realm of Large Language Models (LLMs), the phenomenon of "hallucinations" has garnered significant attention. Defined in the literature, notably in the "Survey of Hallucination in Natural Language Generation" paper \cite{ziweiHallucinations}, hallucination in an LLM is characterized as "the generation of content that is nonsensical or unfaithful to the provided source." This terminology, although rooted in psychological parlance, has been appropriated within the field of artificial intelligence.

Hallucinations in LLMs can be broadly categorized into two types:

\begin{enumerate}
    \item \textbf{Intrinsic Hallucinations}: These directly conflict with the source material, introducing factual inaccuracies or logical inconsistencies.
    \item \textbf{Extrinsic Hallucinations}: These, while not contradicting, are unverifiable against the source, encompassing speculative or unconfirmable elements.
\end{enumerate}

The definition of 'source' in LLM contexts varies with the task. In dialogue-based tasks, it refers to 'world knowledge', whereas in text summarization, it pertains to the input text itself. This distinction plays a crucial role in evaluating and interpreting hallucinations. The impact of hallucinations is also highly context-dependent. For instance, in creative endeavors like poem writing, hallucinations might be deemed acceptable or even beneficial.

LLMs, trained on diverse datasets including the internet, books, and Wikipedia, generate text based on probabilistic models without an inherent understanding of truth or falsity. Recent advancements like instruct tuning and Reinforcement Learning from Human Feedback (RLHF) have attempted to steer LLMs towards more factual outputs, but the fundamental probabilistic nature and its inherent limitations remain. A recent study, “Sources of Hallucination by Large Language Models on Inference Tasks” \cite{mckenna2023sources}, highlights two key aspects contributing to hallucinations in LLMs: the veracity prior and the relative frequency heuristic, underscoring the complexities inherent in LLM training and output generation.

Effective automated measurement of hallucinations in LLMs requires a combination of statistical and model-based metrics.

\emph{Statistical Metrics}:
\begin{itemize}
    \item Metrics like ROUGE \cite{lin-2004-rouge} and BLEU \cite{papineni-etal-2002-bleu} are common for assessing text similarity, focusing on intrinsic hallucinations.
    \item Advanced metrics such as PARENT \cite{dhingra-etal-2019-PARENT}, PARENT-T \cite{wang-etal-2020-PARENT-T}, and Knowledge F1 \cite{Song_Zhang_Hu_Liu_2020} are utilized when structured knowledge sources are available. These metrics, while effective, have limitations in capturing syntactic and semantic nuances.
\end{itemize}

\emph{Model-Based Metrics}:
\begin{itemize}
    \item \textbf{IE-Based Metrics}: Utilize Information Extraction models to simplify knowledge into relational tuples, then compare these with the source.
    \item \textbf{QA-Based Metrics}: Assess the overlap between generated content and the source through a question-answering framework (see \cite{honovich-etal-2021-q2}).
    \item \textbf{NLI-Based Metrics}: Use Natural Language Inference datasets to evaluate the truthfulness of a generated hypothesis based on a given premise (see \cite{dziri-etal-2022-evaluating}).
    \item \textbf{Faithfulness Classification Metrics}: Offer a refined assessment by creating task-specific datasets for a nuanced evaluation (see \cite{Santhanam2021RomeWB}).
\end{itemize}

Despite advances in automated metrics, human judgment remains a vital piece. It typically involves two methodologies:

\begin{enumerate}
    \item \textbf{Scoring}: Human evaluators rate the level of hallucination within a predefined scale.
    \item \textbf{Comparative Analysis}: Evaluators compare generated content against baseline or ground-truth references, adding an essential layer of subjective assessment.
\end{enumerate}

FactScore \cite{min2023factscore} is a recent example of a metric that can be used both for human and model-based evaluation. The metric breaks an LLM generation into “atomic facts”. The final score is computed as the sum of the accuracy of each atomic fact, giving each of them equal weight. Accuracy is a binary number that simply states whether the atomic fact is supported by the source. The authors implement different automation strategies that use LLMs to estimate this metric.

Finally, mitigating hallucinations in LLMs is a multifaceted challenge, requiring tailored strategies to suit various applications.  Those include:
\begin{itemize}
    \item Product Design and User Interaction Strategies such as use case design, structuring the input/output, or providing mechanisms for user feedback.
    \item Data Management and Continuous Improvement. Maintaining and analyzing a tracking set of hallucinations is essential for ongoing model improvement.
    \item Prompt Engineering and Metaprompt Design. Many of the advanced prompt techniques described in \ref{sub:Prompts} such as Retrieval Augmented Generation directly address hallucination risks.
    \item Model Selection and Configuration for Hallucination Mitigation. For exemple, larger models with lower temperature settings usually perform better. Also, techniques such as RLHF or domain-sepcific fine-tuning can mitigate hallucination risks.
\end{itemize}

\subsection{Using LLMs: \textbf{Prompt Design and Engineering}}\label{sub:Prompts}

A prompt in generative AI models is the textual input provided by users to guide the model's output. This could range from simple questions to detailed descriptions or specific tasks. Prompts generally consist of instructions, questions, input data, and examples. In practice, to elicit a desired response from an AI model, a prompt must contain either instructions or questions, with other elements being optional. Advanced prompts involve more complex structures, such as "chain of thought" prompting, where the model is guided to follow a logical reasoning process to arrive at an answer.

Prompt engineering is a rapidly evolving discipline that shapes the interactions and outputs of LLMs and other generative AI models. The essence of prompt engineering lies in crafting the optimal prompt to achieve a specific goal with a generative model. This process is not only about instructing the model but also involves some understanding of the model's capabilities and limitations, and the context within which it operates. 

Prompt engineering transcends the mere construction of prompts; it requires a blend of domain knowledge, understanding of the AI model, and a methodical approach to tailor prompts for different contexts. This might involve creating templates that can be programmatically modified based on a given dataset or context. For example, generating personalized responses based on user data might use a template that is dynamically filled with relevant user information. 

Furthermore, prompt engineering is an iterative and exploratory process, akin to traditional machine learning practices such as model evaluation or hyperparameter tuning. The rapid growth of this field suggests its potential to revolutionize certain aspects of machine learning, moving beyond traditional methods like feature or architecture engineering. On the other hand, traditional engineering practices such as version control and regression testing need to be adapted to this new paradigm just like they were adapted to other machine learning approaches \cite{Sculley2014CreditCard}.

In the following paragraphs we detail some of the most interesting and popular prompt engineering approaches.

\subsubsection{Chain of Thought (CoT)}

The Chain of Thought (CoT) technique, initially described in the paper ``Chain-of-Thought Prompting Elicits Reasoning in Large Language Models''\cite{Wei2022COT} by Google researchers, represents a pivotal advancement in prompt engineering for Large Language Models (LLMs). This approach hinges on the understanding that LLMs, while proficient in token prediction, are not inherently designed for explicit reasoning. CoT addresses this by guiding the model through essential reasoning steps.

CoT is based on making the implicit reasoning process of LLMs explicit. By outlining the steps required for reasoning, the model is directed closer to a logical and reasoned output, especially in scenarios demanding more than simple information retrieval or pattern recognition.

CoT prompting manifests in two primary forms:
\begin{enumerate}
    \item \textbf{Zero-Shot CoT:} This form involves instructing the LLM to ``think step by step'', prompting it to deconstruct the problem and articulate each stage of reasoning.
    \item \textbf{Manual CoT:} A more complex variant, it requires providing step-by-step reasoning examples as templates for the model. While yielding more effective results, it poses challenges in scalability and maintenance.
\end{enumerate}

Manual CoT is more effective than zero-shot. However, the effectiveness of this example-based CoT depends on the choice of diverse examples, and constructing prompts with such examples of step by step reasoning by hand is hard and error prone. That is where automatic CoT \cite{zhang2022automaticCOT} comes into play.

\subsubsection{Tree of Thought (ToT)}
The Tree of Thought (ToT) \cite{yao2023tree} prompting technique is inspired by the concept of considering various alternative solutions or thought processes before converging on the most plausible one.
ToT is based on the idea of branching out into multiple "thought trees" where each branch represents a different line of reasoning. This method allows the LLM to explore various possibilities and hypotheses, much like human cognitive processes where multiple scenarios are considered before determining the most likely one.

A critical aspect of ToT is the evaluation of these reasoning paths. As the LLM generates different branches of thought, each is assessed for its validity and relevance to the query. This process involves real-time analysis and comparison of the branches, leading to a selection of the most coherent and logical outcome.

ToT is particularly useful in complex problem-solving scenarios where a single line of reasoning might not suffice. It allows LLMs to mimic a more human-like problem-solving approach, considering a range of possibilities before arriving at a conclusion. This technique enhances the model's ability to handle ambiguity, complexity, and nuanced tasks, making it a valuable tool in advanced AI applications.

\subsubsection{Self-Consistency}
Self-Consistency \cite{manakul2023selfcheckgpt} utilizes an ensemble-based method, where the LLM is prompted to generate multiple responses to the same query. The consistency among these responses serves as an indicator of their accuracy and reliability.

The Self-Consistency approach is grounded in the principle that if an LLM generates multiple, similar responses to the same prompt, it is more likely that the response is accurate. This method involves asking the LLM to tackle a query multiple times, each time analyzing the response for consistency. This technique is especially useful in scenarios where factual accuracy and precision are paramount.

The consistency of responses can be measured using various methods. One common approach is to analyze the overlap in the content of the responses. Other methods may include comparing the semantic similarity of responses or employing more sophisticated techniques like BERT-scores or n-gram overlaps. These measures help in quantifying the level of agreement among the responses generated by the LLM.

Self-Consistency has significant applications in fields where the veracity of information is critical. It is particularly relevant in scenarios like fact-checking, where ensuring the accuracy of information provided by AI models is essential. By employing this technique, prompt engineers can enhance the trustworthiness of LLMs, making them more reliable for tasks that require high levels of factual accuracy.

\subsubsection{Reflection}
Reflection \cite{shinn2023reflexion} involves prompting LLMs to assess and potentially revise their own outputs based on reasoning about the correctness and coherence of their responses. The concept of Reflection centers on the ability of LLMs to engage in a form of self-evaluation. After generating an initial response, the model is prompted to reflect on its own output, considering factors like factual accuracy, logical consistency, and relevance. This introspective process can lead to the generation of revised or improved responses.

A key aspect of Reflection is the LLM's capacity for self-editing. By evaluating its initial response, the model can identify potential errors or areas of improvement. This iterative process of generation, reflection, and revision enables the LLM to refine its output, enhancing the overall quality and reliability of its responses.

\subsubsection{Expert Prompting}
Expert Prompting \cite{zhang2023exploring} enhances the capabilities of Large Language Models (LLMs) by simulating the responses of experts in various fields. This method involves prompting the LLMs to assume the role of an expert and respond accordingly, providing high-quality, informed answers. A key strategy within Expert Prompting is the multi-expert approach. The LLM is prompted to consider responses from multiple expert perspectives, which are then synthesized to form a comprehensive and well-rounded answer. This technique not only enhances the depth of the response but also incorporates a range of viewpoints, reflecting a more holistic understanding of the subject matter.

\subsubsection{Chains}
Chains refer to the method of linking multiple components in a sequence to handle complex tasks with Large Language Models (LLMs). This approach involves creating a series of interconnected steps or processes, each contributing to the final outcome. The concept of Chains is based on the idea of constructing a workflow where different stages or components are sequentially arranged. Each component in a Chain performs a specific function, and the output of one serves as the input for the next. This end-to-end arrangement allows for more complex and nuanced processing, as each stage can be tailored to handle a specific aspect of the task. Chains can vary in complexity and structure, depending on the requirements. In “PromptChainer: Chaining Large Language Model Prompts through Visual Programming” \cite{wu2022promptchainer}, the authors not only describe the main challenges in designing chains, but also describe a visual tool to support those tasks.

\subsubsection{Rails}
Rails in advanced prompt engineering refer to a method of guiding and controlling the output of Large Language Models (LLMs) through predefined rules or templates. This approach is designed to ensure that the model's responses adhere to certain standards or criteria, enhancing the relevance, safety, and accuracy of the output. The concept of Rails involves setting up a framework or a set of guidelines that the LLM must follow while generating responses. These guidelines are typically defined using a modeling language or templates known as Canonical Forms, which standardize the way natural language sentences are structured and delivered.

Rails can be designed for various purposes, depending on the specific needs of the application:
\begin{itemize}
    \item \textbf{Topical Rails:} Ensure that the LLM sticks to a particular topic or domain.
    \item \textbf{Fact-Checking Rails:} Aimed at minimizing the generation of false or misleading information.
    \item \textbf{Jailbreaking Rails:} Prevent the LLM from generating responses that attempt to bypass its own operational constraints or guidelines.
\end{itemize}

\subsubsection{Automatic Prompt Engineering (APE)}
Automatic Prompt Engineering (APE) \cite{zhou2023large} focuses on automating the process of prompt creation for Large Language Models (LLMs). APE seeks to streamline and optimize the prompt design process, leveraging the capabilities of LLMs themselves to generate and evaluate prompts. APE involves using LLMs in a self-referential manner where the model is employed to generate, score, and refine prompts. This recursive use of LLMs enables the creation of high-quality prompts that are more likely to elicit the desired response or outcome. 

The methodology of APE can be broken down into several key steps:
\begin{itemize}
    \item \textbf{Prompt Generation:} The LLM generates a range of potential prompts based on a given task or objective.
    \item \textbf{Prompt Scoring:} Each generated prompt is then evaluated for its effectiveness, often using criteria like clarity, specificity, and likelihood of eliciting the desired response.
    \item \textbf{Refinement and Iteration:} Based on these evaluations, prompts can be refined and iterated upon, further enhancing their quality and effectiveness.
\end{itemize}

\subsection{\textbf{Augmenting LLMs through external knowledge - RAG}}\label{sub:RAG}
\begin{figure*}
    \centering
    \includegraphics[scale=0.5]{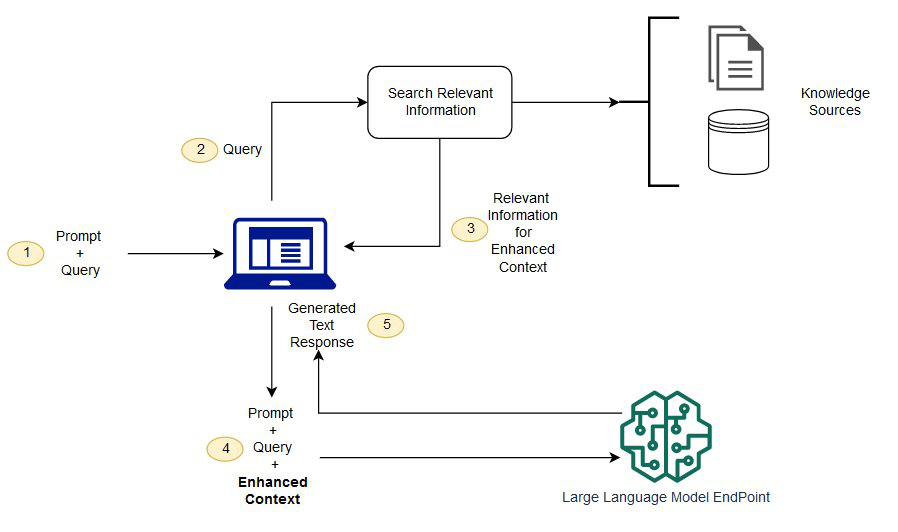}
    \caption{An example of synthesizing RAG with LLMs for question answering application \cite{aws-qa-rag-sagemaker}.}
    \label{fig:rag}
\end{figure*}

\begin{figure}
    \centering
    \includegraphics[scale=0.5]{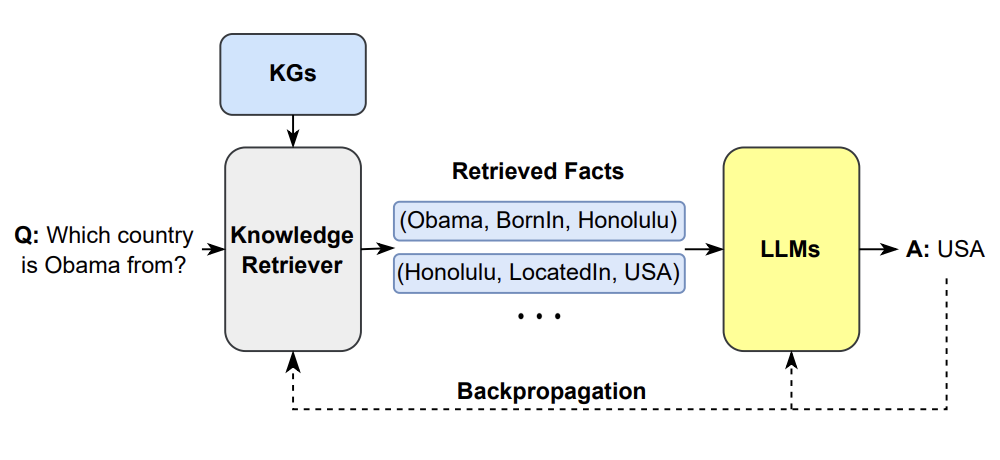}
    \caption{This is one example of synthesizing the KG as a retriever with LLMs \cite{pan2023unifying}.}
    \label{fig:rag_kg}
\end{figure}

One of the main limitations of pre-trained LLMs is their lack of up-to-date knowledge or access to private or use-case-specific information. This is where retrieval augmented generation (RAG) comes into the picture \cite{DBLP:journals/corr/abs-2005-11401}. RAG, illustrated in figure \ref{fig:rag}, involves extracting a query from the input prompt and using that query to retrieve relevant information from an external knowledge source (e.g. a search engine or a knowledge graph, see figure \ref{fig:rag_kg} ). The relevant information is then added to the original prompt and fed to the LLM in order for the model to generate the final response. A RAG system includes three important components: Retrieval, Generation, Augmentation \cite{gao2023retrieval}. 

\paragraph{RAG-aware prompting techniques}

Because of the importance of RAG to build advanced LLM systems, several RAG-aware prompting techniques have been developed recently. One such technique is Forward-looking Active Retrieval Augmented Generation (FLARE)

Forward-looking Active Retrieval Augmented Generation (FLARE) \cite{jiang2023active} enhances the capabilities of Large Language Models (LLMs) by iteratively combining prediction and information retrieval. FLARE represents an evolution in the use of retrieval-augmented generation, aimed at improving the accuracy and relevance of LLM responses.

FLARE involves an iterative process where the LLM actively predicts upcoming content and uses these predictions as queries to retrieve relevant information. This method contrasts with traditional retrieval-augmented models that typically retrieve information once and then proceed with generation. In FLARE, this process is dynamic and ongoing throughout the generation phase. In FLARE, each sentence or segment generated by the LLM is evaluated for confidence. If the confidence level is below a certain threshold, the model uses the generated content as a query to retrieve relevant information, which is then used to regenerate or refine the sentence. This iterative process ensures that each part of the response is informed by the most relevant and current information available.

For more details on RAG framework and its relevant works, we refer the readers to this survey of retrieval augmented generations 
 \cite{gao2023retrieval}.

\subsection{\textbf{Using External Tools}}

Retrieving information from an external knowledge source as described above is only one of the potential ways to augment an LLM. More generally, an LLM can access any number of external tools (e.g. an API to a service) to augment its functionality. In that regards, RAG can be seen as a specific instance of the broader category of the so called "tools".

Tools in this context are external functions or services that LLMs can utilize. These tools extend the range of tasks an LLM can perform, from basic information retrieval to complex interactions with external databases or APIs.

In the paper "Toolformer: Language Models Can Teach Themselves to Use Tools" 
\cite{schick2023toolformer}, the authors go beyond simple tool usage by training an LLM to decide what tool to use when, and even what parameters the API needs. Tools include two different search engines, or a calculator. In the following examples, the LLM decides to call an external Q\&A tool, a calculator, and a Wikipedia Search Engine More recently, researchers at Berkeley have trained a new LLM called Gorilla \cite{patil2023gorilla} that beats GPT-4 at the use of APIs, a specific but quite general tool.

\paragraph{Tool-aware prompting techniques}

Similarly to what was described with RAG, several tool-aware prompting approaches have been developed to make usage of tools more scalable. A popular technique is the so called 
Automatic Multi-step Reasoning and Tool-use (ART).

Automatic Multi-step Reasoning and Tool-use (ART) \cite{paranjape2023art} is a prompt engineering technique that combines automated chain of thought prompting with the use of external tools. ART represents a convergence of multiple prompt engineering strategies, enhancing the ability of Large Language Models (LLMs) to handle complex tasks that require both reasoning and interaction with external data sources or tools.

ART involves a systematic approach where, given a task and input, the system first identifies similar tasks from a task library. These tasks are then used as examples in the prompt, guiding the LLM on how to approach and execute the current task. This method is particularly effective when tasks require a combination of internal reasoning and external data processing or retrieval.

\subsection{\textbf{LLM Agents}}

The idea of AI agents has been well-explored in the history of AI. An agent is typically an autonomous entity that can perceive the environment using its sensors, make a judgment based on the state it currently is, and accordingly act based on the actions that are available to it. 

In the context of LLMs, an agent refers to a system based on a specialized instantiation of an (augmented) LLM that is capable of performing specific tasks autonomously. These agents are designed to interact with users and environment to
make decisions based on the input and the intended goal of the interaction. Agents are based on LLMs equipped with the ability to access and use tools, and to make decisions based on the given input. They are designed to handle tasks that require a degree of autonomy and decision-making, typically beyond simple response generation.  

The functionalities of a generic LLM-based agent include:
\begin{itemize}
    \item {Tool Access and Utilization:} Agents have the capability to access external tools and services, and to utilize these resources effectively to accomplish tasks.
    \item {Decision Making:} They can make decisions based on the input, context, and the tools available to them, often employing complex reasoning processes.
\end{itemize}

As an example, an LLM that has access to a function (or an API) such as weather API, can answer any question related to the weather of the specific place. In other words, it can use APIs to solve problems. Furthermore, if that LLM has access to an API that allows to make purchases, a purchasing agent can be built to not only have capabilities to read information from the external world, but also act on it \cite{shen2023hugginggpt}.

\begin{figure*}
    \centering
    \includegraphics[scale=0.28]{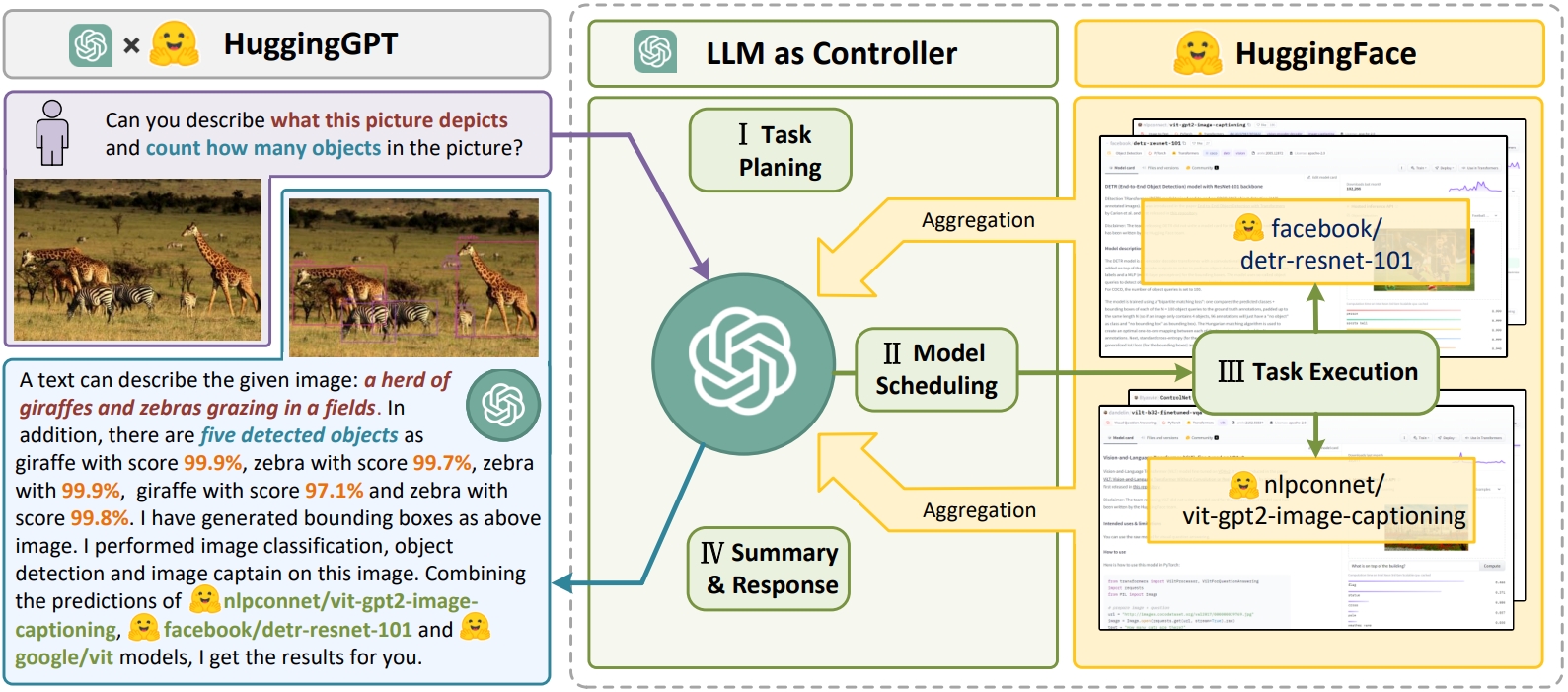}
    \caption{HuggingGPT: An agent-based approach to use tools and planning [image courtesy of \cite{shen2023hugginggpt}]}
    \label{fig:enter-label}
\end{figure*}

\begin{figure}[h]
\begin{center}
    \includegraphics [scale=0.95] {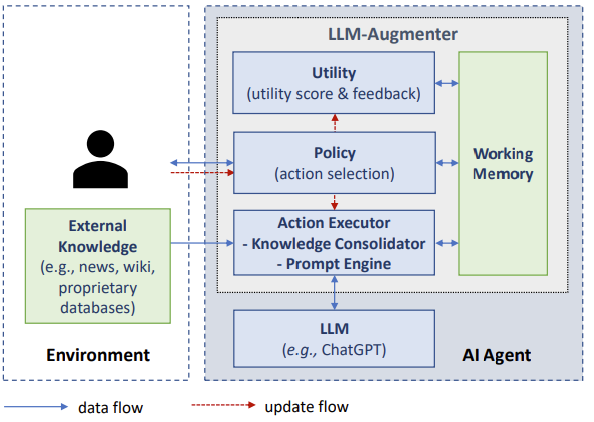}
\end{center}
  \caption{A LLM-based agent for conversational information seeking. Courtesy of \cite{peng2023check}.}
\label{fig:llm_augmenter}
\end{figure}

Fig. \ref{fig:llm_augmenter} shows another example of LLM-based agents for conversational information seeking \cite{peng2023check}, where an LLM is augmented with a set of plug-and-play modules, including a
\emph{working memory} that tracks the dialog state, a
\emph{policy} that makes an execution plan for the task and selects next system action, an
\emph{action executor} that performs an action selected by the policy (consolidating evidence from external knowledge, or prompting the LLM to generate responses), and a
\emph{utility} that accesses the alignment of the LLM’s responses with user expectations or specific business requirements, and generate feedback to improve agent performance.

For more details on LLM-based AI agents see recent survey \cite{xi2023rise, wang2023survey, durante2024agent}. 

\paragraph{Prompt engineering techniques for agents}
Like RAG and Tools, prompt engineering techniques that specifically address the needs of LLM-based agents have been developed. Three such examples are Reasoning without Observation (ReWOO), Reason and Act (ReAct), and Dialog-Enabled Resolving Agents (DERA).

Reasoning without Observation (ReWOO) \cite{xu2023rewoo} aims to decouple reasoning from direct observations. ReWOO operates by enabling LLMs to formulate comprehensive reasoning plans or meta-plans without immediate reliance on external data or tools. This approach allows the agent to create a structured framework for reasoning that can be executed once the necessary data or observations are available. In ReWOO, the LLM initially develops a plan (a series of steps) that outlines how to approach and solve a given problem. This meta-planning phase is crucial as it sets the stage for the agent to process information once it becomes available. The execution phase then involves integrating actual data or observations into the pre-specified plan, leading to coherent and contextually relevant responses. ReWOO offers significant advantages in terms of token efficiency and robustness to tool failure. It enables LLMs to handle tasks where immediate access to external data is not available, relying instead on a well-structured reasoning framework. This method is particularly advantageous in scenarios where data retrieval is costly, slow, or uncertain, allowing the LLM-based agent to maintain a high level of performance and reliability.

Reason and Act (ReAct)\cite{yao2023react} prompts LLMs to generate not only verbal reasoning but also actionable steps, thus enhancing the model's dynamic problem-solving capabilities. ReAct is grounded in the principle of integrating reasoning with action. In this approach, the LLM is prompted to alternate between generating reasoning traces (explanations) and taking actions (steps or commands) in an interleaved manner. This approach allows the model to dynamically reason about a problem, and propose and take concrete actions simultaneously.

Dialog-Enabled Resolving Agents (DERA) \cite{nair2023dera} are specialized AI agents that can engage in dialogue, resolve queries, and make decisions based on interactive exchanges. DERA is developed based on the idea of utilizing multiple agents within a dialog context, each with specific roles and functions. These agents can include Researchers, who gather and analyze information, and Deciders, who make final judgments based on the information provided. This division of roles allows for a well-organized and efficient approach to problem-solving and decision-making. DERA is particularly advantageous in scenarios requiring complex decision-making and problem-solving, such as those in medical diagnostics or customer service. The collaborative and interactive nature of DERA agents allows them to handle intricate queries with a level of depth and nuance that single-agent systems might struggle with. Moreover, this approach aligns well with human decision-making processes, making AI reasoning more relatable and trustworthy.

\section{Popular Datasets for LLMs} 
\label{sec:llm_datasets}
Large language models exhibit promising accomplishments, but the main question that arises is how effectively they function and how their performance can be assessed in specific tasks or applications.

The evaluation of LLMs poses particular challenges due to the evolving landscape of their applications. The original intent behind developing LLMs was to boost the performance of NLP tasks such as translation, summarization, question-answering, and so on \cite{chang2023survey}. However, it is evident today that these models are finding utility across diverse domains including code generation and finance. Moreover, the evaluation of LLMs encompasses several critical considerations such as fairness and bias, fact-checking, and reasoning.  In this section, we outline the commonly used benchmarks for assessing LLMs. These benchmarks are categorized based on training or evaluating the LLM Capabilities.

\begin{figure*}
 \begin{center}
    \includegraphics [scale=0.48] {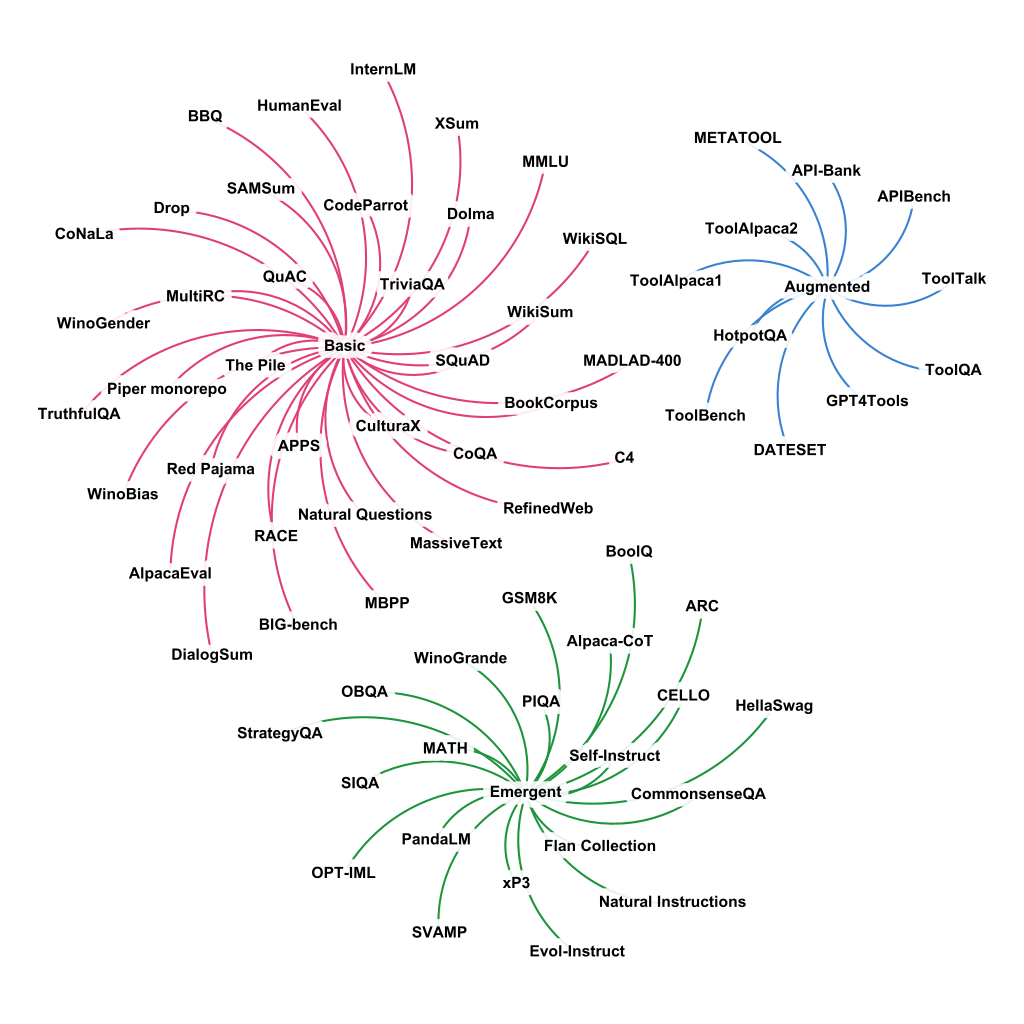}
\end{center}
  \caption{Dataset applications.}
\label{fig:dataset}   
\end{figure*}

\begin{figure*}
 \begin{center}
    \includegraphics [scale=0.8] {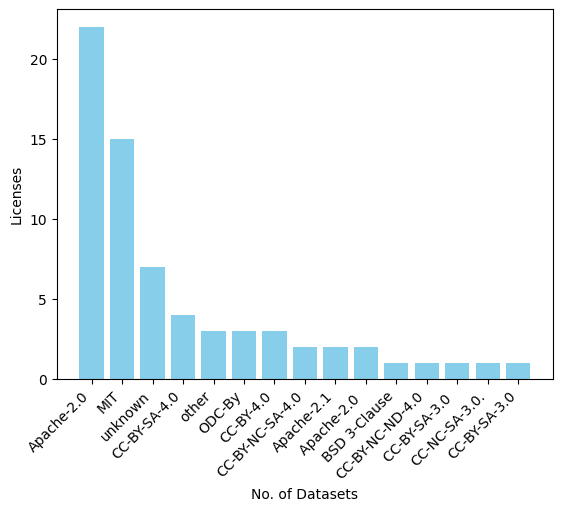}
\end{center}
  \caption{Datasets licensed under different licenses.}
\label{fig:dataset_license}   
\end{figure*}

\subsection{Datasets for Basic Tasks: language modeling/understanding/generation} 

This section provides an overview of the benchmarks and datasets suited to evaluate the basic abilities of LLMs.

\begin{itemize}
    
    \item  \textbf{Natural Questions} \cite{kwiatkowski-etal-2019-natural} is a QA dataset that consists of real anonymized, aggregated queries submitted to the Google search engine as questions. An annotator is presented with a question along with a Wikipedia page from the top $5$ search results, and annotates a long answer (typically a paragraph) and a short answer (one or more entities) if present on the page, or marks null if no long/short answer is present.

    \item \textbf{MMLU} \cite{hendrycks2021measuring} is intended to evaluate the knowledge gained in zero-shot and few-shot scenarios. That means that MMLU assesses both the general knowledge and problem-solving ability of a model. It covers 57 subjects in STEM, humanities, social sciences, and other areas. The benchmark varies in complexity, ranging from elementary to advanced professional.  It is worth mentioning that the main contribution of this dataset is for multi-task language understanding, question answering, and arithmetic reasoning. 
    
    \item  \textbf{MBPP} \cite{austin2021program} stands for ``Mostly Basic Python Problems'' and provides a benchmark for evaluating the performance of models designed for code generation. The benchmark encompasses $974$ short Python programs including a wide range of topics, including fundamental programming concepts and standard library usage, and more. Each challenge comprises a task description, a code solution, and three automated test cases. 
    
    \item \textbf{HumanEval} \cite{choi-etal-2018-quac} is a dataset for code generation task. This dataset consists of $164$ hand-crafted programming challenges.  Each challenge is accompanied by a function signature, docstring, code body, and multiple unit tests.  The main intuition behind developing this dataset is to guarantee the exclusion of its contents from training datasets for code generation models.
    
     \item \textbf{APPS} \cite{hendrycksapps2021} is designed for code generation task focusing on the Python programming language. The APPS dataset contains a collection of $232,444$ Python programs.   Each program in the dataset has an average of $18$ lines of Python code. Additionally, APPS offers access to a repository of $10,000$ unique programming exercises, each with text-based problem descriptions.   The final aspect to highlight is that the it includes test cases. 
     
     \item \textbf{WikiSQL} \cite{zhong2017seq2sql} is crafted for code generation task and it has 87,726 carefully labeled pairs of SQL queries and corresponding natural language questions from Wikipedia tables. The SQL queries comprise three subsets: test sets ($17,284$ examples), development ($9,145$ examples), and training ($61,297$ examples).

    \item \textbf{TriviaQA} \cite{joshi-etal-2017-triviaqa} is designed for QA task. This dataset comprises more than $650,000$ question-answer-evidence triples. There are $95,000$ question-answer pairs in this dataset, each authored by trivia enthusiasts and supported by an average of six independently sourced evidence documents. These documents are automatically acquired from Wikipedia or broader web search results. The dataset is categorized into two segments, including those with authentic answers from Wikipedia and web domains, and verified sets embody the accurately answered questions along with their associated documents from both Wikipedia and online.

    \item \textbf{RACE} \cite{lai-etal-2017-race} suits for reading comprehension task. This dataset is based on English tests completed by Chinese students from middle school and high school, aged $12$ to $18$, and it contains roughly $28,000$ texts and $100,000$ questions rigorously prepared by human specialists, primarily English instructors. This dataset contains a wide range of subjects that were purposefully chosen to assess students' comprehension and reasoning abilities.  This dataset is available in three subgroups: RACE-M, RACE-H, and RACE. RACE-M refers to the middle school examinations, whereas RACE-H denotes the high school tests. Finally, RACE is the synthesis of RACE-M and RACE-H.
    
    \item \textbf{SQuAD} \cite{rajpurkar-etal-2016-squad} stands for ``Stanford Question Answering Dataset'' and is a crowdsourced reading comprehension dataset based on Wikipedia articles. It has approximately $100,000$ question-answer pairs connected to more than $500$ articles. The answers to these questions are typically text fragments or spans taken from the corresponding reading passages. The questions may be unanswerable in some cases. The dataset is divided into three sets: an $80\%$ training set, a $10\%$ development set, and a $10\%$ hidden test set. 

    \item \textbf{BoolQ} \cite{DBLP:journals/corr/abs-1905-10044} is a yes/no question-answering dataset where the goal is reading comprehension task. BoolQ includes $15,942$ examples.  Each example is a triplet that includes a question, a relevant paragraph, and the solution. Although the main intuition behind this dataset is for reading comprehension, it can be used for reasoning, natural language inference, and question-answering tasks. 

    \item \textbf{MultiRC} \cite{MultiRC2018} is another dataset that fits reading comprehension task. MultiRC contains brief paragraphs as well as multi-sentence questions that can be answered using the information in the paragraph.  The paragraphs in this dataset come from a variety of sources, including news, fiction, historical texts, Wikipedia articles, discussions on society and law, elementary school science textbooks, and 9/11 reports.  Each question has many response choices, with one or more of them being correct. Answering the questions requires reasoning across several sentences. MultiRC dataset encompasses around $6,000$ multi-sentence questions gathered from over 800 paragraphs. On average, each question offers about two valid answer alternatives out of a total of five.
\end{itemize}

\subsection{Datasets for Emergent: ICL, reasoning (CoT), instruction following}
This section centers on the benchmarks and datasets employed to evaluate the emergent abilities of LLMs.

\begin{itemize}
    \item \textbf{GSM8K} \cite{DBLP:journals/corr/abs-2110-14168}  is designed to evaluate the model's ability for multi-step mathematical reasoning. GSM8K includes 8.5K linguistically diverse grade school math word problems written by humans.  The dataset is split into two sets: a training set with $7.5K$ problems,  and a test set with $1$K problems. These problems need $2$ to $8$ steps to be solved.  Solutions mainly are a series of elementary calculations using basic arithmetic operations. 
    
    \item \textbf{MATH} \cite{DBLP:journals/corr/abs-2103-03874} enables to assess how well models can solve math problems. MATH dataset hast $12$, $500$ problems from high school math competitions. Each problem in the dataset has a step-by-step solution and a final answer enclosed in a box. The problems cover a wide range of topics and have different levels of complexity. There are seven subjects in total. Furthermore, the difficulty of each problem is rated based on the  AoPS standards on a scale from $'1'$ to $'5'$. A $'1'$ shows the easiest problems in a subject, while $'5'$ represents the most difficult.  In terms of formatting, all problems and solutions are presented using LATEX and the Asymptote vector graphics language.

    \item \textbf{HellaSwag} \cite{zellers2019hellaswag} is designed to assess commonsense reasoning in LLMs. This benchmark includes $70,000$ multiple-choice questions. Each question is derived from one of two domains: ActivityNet or WikiHow, and presents four answer choices regarding what might happen in the following situation. The correct answer provides an actual statement describing the upcoming event, but the three wrong answers are created to confuse machines. 
    
    \item \textbf{AI2 Reasoning Challenge (ARC)} \cite{DBLP:journals/corr/abs-1803-05457} is used for commonsense reasoning. This benchmark encompasses $7,787$ science examination questions. These questions are in English, and most of them are set up in a multiple-choice format. The questions have been divided into two groups: a Challenge Set with $2,590$ difficult questions and an Easy Set with 5,197 questions. Each collection has also been pre-divided into Train, Development, and Test subsets.
    
    \item \textbf{PIQA} \cite{DBLP:journals/corr/abs-1911-11641} is intended to evaluate the language representations on their knowledge of physical commonsense. In this dataset,  the focus is on everyday situations with a preference for uncommon solutions.  The central task is a multiple-choice question answering, where a question $(q)$ is provided along with two potential solutions $(s1, s2)$. Then,  the best solution is chosen by whether a model or a human. For each question, only one of the solutions is the correct answer.

    \item \textbf{SIQA} \cite{DBLP:journals/corr/abs-1904-09728} provides a framework for evaluating models' ability for commonsense reasoning about social situations. SIQA dataset has $38,000$ multiple-choice questions designed to assess emotional and social intelligence in everyday circumstances. This dataset covers a wide variety of social scenarios. In SIQA, the potential answers is a mixture of human-selected responses and machine-generated ones that have been filtered through adversarial processes.

    \item \textbf{OpenBookQA (OBQA)} \cite{DBLP:journals/corr/abs-1809-02789} is a new kind of question-answering dataset where answering its questions requires additional common and commonsense knowledge not contained in the book and rich text comprehension. This dataset includes around 6,000 multiple-choice questions. Each question is linked to one core fact, as well as an additional collection of over $6000$ facts. The questions were developed using a multi-stage crowdsourcing and expert filtering procedure. OpenBookQA questions are difficult because they need multi-hop reasoning with limited background.

    \item  \textbf{TruthfulQA} \cite{lin2021truthfulqa} is designed specifically to evaluate the truthfulness of language models in generating answers to questions. This dataset includes 817 questions, written by authors, from $38$ different categories, including health, law, finance, and politics. These questions are purposefully designed to challenge human responders, as they may contain common misunderstandings that lead to incorrect answers.

    \item  \textbf{OPT-IML Bench} \cite{iyer2022opt} is a comprehensive benchmark for Instruction Meta-Learning.  It covers  2000  NLP tasks from 8 existing benchmarks. The OPT-IML Bench consists of a training set with 17.9 M examples, a dev set with 145K samples, and a test set with 321K samples.
\end{itemize}

\subsection{Datasets for Augmented: using external knowledge/tools}
This section focuses on datasets designed for the augmented abilities of LLMs.

\begin{itemize}
    \item \textbf{HotpotQA} \cite{DBLP:journals/corr/abs-1809-09600} is designed to cover a diverse and explainable question-answering dataset that necessitates multi-hop reasoning.  This dataset is derived from the English Wikipedia. It consists of roughly $113,000$ questions. Each question in the dataset comes with two paragraphs, called gold paragraphs, from two Wikipedia articles. Also, there is a list of sentences in those paragraphs that crowdworkers have picked as important for answering the question.  

    \item \textbf{ToolQA} \cite{zhuang2023toolqa} is a question answering benchmark to evaluate LLMs' ability to use external tools for answering questions. 

    \item \textbf{GPT4Tools} serves as an instructional dataset, generated by instructing advanced teachers (such as ChatGPT), with instructions conditioned on visual content and tool descriptions. This process results in the generation of instructions related to the use of tools. There are three versions of this dataset. The first version comprises 71,000 instruction-following data points utilized to fine-tune the GPT4Tools model. The next version consists of manually cleaned instruction data used for validation, covering instructions related to the tools from the first version. The last version is cleaned instruction data used for testing and includes instructions related to some tools that are not present in the first version.
\end{itemize}

\begin{table*}
    \centering
    \caption{LLM Datasets Overview.} 
    \label{table:dataset_overview}
    \begin{tabular}{|m{2.5cm}|m{4cm}|m{2cm}|m{2cm}|m{2cm}|}
    \hline
        \textbf{Benchmark Name} & \textbf{Evaluation Metric} & \textbf{Leaderboard} & \textbf{Source} & \textbf{paperswithcode} \\ \hline
        HumanEval & PASS@k & \href{https://llm-leaderboard.streamlit.app}{Link} & \href{https://github.com/openai/human-eval}{Link} & \href{https://paperswithcode.com/sota/code-generation-on-humaneval}{Link} \\ \hline
        MBPP &  PASS@k, Accuracy  &  - & \href{https://github.com/google-research/google-research/tree/master/mbpp}{Link} & \href{https://paperswithcode.com/sota/code-generation-on-mbpp}{Link} \\ \hline 
        APPS &  PASS@k, Accuracy  &  - & \href{https://github.com/hendrycks/apps}{Link}& \href{https://paperswithcode.com/sota/code-generation-on-apps}{Link} \\ \hline
        WikiSQL & Accuracy &  - & \href{https://github.com/IBM/SQL-to-Text}{Link} & \href{https://paperswithcode.com/sota/code-generation-on-wikisql}{Link}  \\ \hline 
        CoNaLa & BLEU &  ~ & \href{https://conala-corpus.github.io/ }{Link}& \href{https://paperswithcode.com/sota/code-generation-on-conala}{Link}  \\ \hline
        CodeParrot & PASS@k &  - & \href{https://github.com/huggingface/blog/blob/main/codeparrot.md}{Link} & - \\ \hline
        HellaSwag & Accuracy &  \href{https://huggingface.co/spaces/HuggingFaceH4/open\_llm\_leaderboard}{Link}  & \href{https://rowanzellers.com/hellaswag}{Link} & \href{https://paperswithcode.com/sota/sentence-completion-on-hellaswag}{Link} \\ \hline
        AI2 Reasoning Challenge (ARC) & Accuracy &  \href{https://huggingface.co/spaces/HuggingFaceH4/open\_llm\_leaderboard}{Link} & \href{https://github.com/fchollet/ARC/}{Link} & \href{https://paperswithcode.com/sota/common-sense-reasoning-on-arc-challenge}{Link} \\ \hline
        BoolQ & Accuracy &  - & \href{https://github.com/google-research-datasets/boolean-questions}{Link} & \href{https://paperswithcode.com/sota/question-answering-on-boolq}{Link} \\ \hline
        MultiRC &  F1-score, Accuracy  & - & \href{https://cogcomp.seas.upenn.edu/multirc/}{Link} & \href{https://paperswithcode.com/sota/question-answering-on-multirc}{Link} \\ \hline
        CNN/Daily Mail \cite{chen2016thorough} & Accuracy & - & \href{https://github.com/danqi/rc-cnn-dailymail}{Link}  & - \\ \hline
        SQuAD &  F1-score, EM  & \href{https://rajpurkar.github.io/SQuAD-explorer/ }{Link} & \href{https://rajpurkar.github.io/SQuAD-explorer/ }{Link} &  \href{https://paperswithcode.com/dataset/squad}{Link}  \\ \hline
        RACE & Accuracy &  - & \href{https://www.cs.cmu.edu/\~glai1/data/race/}{Link} & \href{https://paperswithcode.com/sota/reading-comprehension-on-race}{Link}  \\ \hline
        CNN/Daily Mail \cite{nallapati-etal-2016-abstractive} & ROUGE &  - & \href{https://github.com/abisee/cnn-dailymail}{Link}  & \href{https://paperswithcode.com/sota/abstractive-text-summarization-on-cnn-daily}{Link} \\ \hline
        Drop &  F1-score, EM  &  \href{https://leaderboard.allenai.org/drop/submissions/public}{Link}  & \href{https://allenai.org/data/drop}{Link}  & \href{https://paperswithcode.com/sota/question-answering-on-drop-test}{Link}  \\ \hline
        QuAC &  F1-score, HEQ-Q, HEQ-D  &  \href{https://quac.ai/}{Link}  & \href{https://quac.ai/}{Link}  & \href{https://paperswithcode.com/sota/question-answering-on-quac}{Link}\\ \hline
        TriviaQA &  EM, F1-score, Accuracy  &  \href{https://competitions.codalab.org/competitions/17208\#results}{Link} &\href{https://nlp.cs.washington.edu/triviaqa/}{Link}  & \href{https://paperswithcode.com/sota/question-answering-on-triviaqa}{Link}  \\ \hline
        Natural Questions &  EM, F1-score, Accuracy  &  \href{https://ai.google.com/research/NaturalQuestions}{Link}  & \href{https://ai.google.com/research/NaturalQuestions}{Link} & \href{https://paperswithcode.com/sota/question-answering-on-natural-questions}{Link}  \\ \hline
        StrategyQA &  Accuracy, Recall@10, SARI  &  \href{https://leaderboard.allenai.org/strategyqa/submissions/public}{Link} & \href{https://allenai.org/data/strategyqa}{Link} & \href{https://paperswithcode.com/sota/question-answering-on-strategyqa}{Link} \\ \hline
        CoQA & F1-score &  \href{https://stanfordnlp.github.io/coqa/}{Link} & \href{https://stanfordnlp.github.io/coqa/}{Link}  & \href{https://paperswithcode.com/dataset/coqa}{Link} \\ \hline
        XSum & ROUGE &  - & \href{https://github.com/EdinburghNLP/XSum}{Link}  & \href{https://paperswithcode.com/sota/text-summarization-on-x-sum}{Link}  \\ \hline
        SAMSum & ROUGE &  - & - & \href{https://paperswithcode.com/dataset/samsum-corpus}{Link}  \\ \hline
        WikiSum & ROUGE &  - & \href{https://github.com/tensorflow/tensor2tensor/tree/master/tensor2tensor/data_generators/wikisum}{Link} & - \\ \hline
        DialogSum & ROUGE & - & \href{https://github.com/cylnlp/dialogsum}{Link}  & \href{https://paperswithcode.com/dataset/dialogsum}{Link} \\ \hline
        TruthfulQA &  MC1 , MC2, \% true, \% info, BLEURT  & \href{https://huggingface.co/spaces/HuggingFaceH4/open\_llm\_leaderboard}{Link}  & \href{https://github.com/sylinrl/TruthfulQA}{Link}  & \href{https://paperswithcode.com/sota/question-answering-on-truthfulqa}{Link}  \\ \hline
        MMLU & Accuracy &  \href{https://huggingface.co/spaces/HuggingFaceH4/open\_llm\_leaderboard}{Link} & \href{https://github.com/hendrycks/test}{Link} & \href{https://paperswithcode.com/dataset/mmlu}{Link}\\ \hline
        GSM8K & Accuracy &  \href{https://opencompass.org.cn/dataset-detail/GSM8K}{Link}  & \href{https://github.com/openai/grade-school-math}{Link}  & \href{https://paperswithcode.com/sota/arithmetic-reasoning-on-gsm8k}{Link}  \\ \hline
        PIQA & Accuracy &  \href{https://leaderboard.allenai.org/physicaliqa/submissions/public}{Link}  & \href{https://leaderboard.allenai.org/physicaliqa/submissions/about}{Link} & \href{https://paperswithcode.com/sota/question-answering-on-piqa}{Link}  \\ \hline
        SIQA & Accuracy &  \href{https://leaderboard.allenai.org/socialiqa/submissions/public}{Link}  & \href{https://leaderboard.allenai.org/socialiqa/submissions/about}{Link} & \href{https://paperswithcode.com/sota/question-answering-on-social-iqa}{Link}  \\ \hline
        OpenBookQA (OBQA) & Accuracy &  \href{https://leaderboard.allenai.org/open\_book\_qa/submissions/public}{Link}  & \href{https://github.com/allenai/OpenBookQA}{Link} & \href{https://paperswithcode.com/sota/question-answering-on-openbookqa}{Link}  \\ \hline
        HotpotQA &  EM, F1-score, Joint EM, Joint F1-score,   &  \href{https://hotpotqa.github.io/}{Link}  & \href{https://hotpotqa.github.io/}{Link}  & \href{https://paperswithcode.com/sota/question-answering-on-hotpotqa}{Link}  \\ \hline
        MATH & Accuracy &  - & \href{https://github.com/hendrycks/math}{Link}  & \href{https://paperswithcode.com/sota/math-word-problem-solving-on-math}{Link} \\ \hline
        CommonsenseQA & Accuracy & \href{https://www.tau-nlp.sites.tau.ac.il/csqa-leaderboard2}{Link}& \href{https://www.tau-nlp.sites.tau.ac.il/commonsenseqa}{Link}& \href{https://paperswithcode.com/sota/common-sense-reasoning-on-commonsenseqa}{Link} \\ \hline
        Natural Instructions & ROUGE-L, Human & \href{https://leaderboard.allenai.org/natural-instructions/submissions/public}{Link}& \href{https://instructions.apps.allenai.org/}{Link}& \href{https://paperswithcode.com/dataset/natural-instructions}{Link} \\ \hline
        BIG-bench & Accuracy, Average& - & \href{https://github.com/google/BIG-bench}{Link} & \href{https://paperswithcode.com/dataset/big-bench}{Link} \\ \hline
      ToolTalk & Success rate, Precision, Recall, Incorrect action rate, Percent of failing error types & - & \href{https://github.com/microsoft/ToolTalk}{Link} &
        \href{https://paperswithcode.com/paper/tooltlk-evaluating-tool-usage-in-a}{Link} \\ \hline
    MetaTool & Accuracy, Precision, Recall, F1-score& - & \href{https://github.com/HowieHwong/MetaTool?tab=readme-ov-file}{Link} & \href{https://paperswithcode.com/paper/metatool-benchmark-deciding-whether-to-use}{Link}  \\ \hline
     GPT4Tools & Successful Rate of Thought, Successful Rate of Action, Successful Rate of Arguments, Success Rate & - & \href{https://github.com/AILab-CVC/GPT4Tools}{Link} & \href{https://paperswithcode.com/paper/gpt4tools-teaching-large-language-model-to}{Link}  \\ \hline
     API-Bank & Correctness, ROUGE, Error(API Hallucination, Has Exception, Invalid Input Parameters, False API Call Format, API Call, Miss Input Parameters) & - & \href{https://github.com/AlibabaResearch/DAMO-ConvAI/tree/main/api-bank}{Link} &  \href{https://paperswithcode.com/paper/api-bank-a-benchmark-for-tool-augmented-llms}{Link}\\ \hline
     Alpaca-CoT & - & - & \href{https://github.com/PhoebusSi/alpaca-CoT}{Link}  & \href{https://paperswithcode.com/paper/an-empirical-study-of-instruction-tuning}{Link}\\ \hline
    \end{tabular}
\end{table*}

\section{Prominent LLMs' Performance on Benchmarks} 
\label{sec:llm_performance}
In this section we first provide an overview of some of popular metrics used for evaluating the performance of LLMs under different scenarios. We then look at the performance of prominent large language models on some of the popular datasets and benchmarks.

\subsection{Popular Metrics for Evaluating LLMs}
Evaluating the performance of generative language models  depends on the underlying task they are going to be used for. Tasks that are mostly about selecting a choice out of given ones (such as sentiment analysis), can be seen as simple as classification and their performance can be evaluated using classification metrics. Metrics such as accuracy, precision, recall, F1, etc are applicable in this case. It is also important to note that the answers generated by the model for specific tasks such as multi-choice question answering are always either True or False. If the answer is not in a set of options, it can be seen as False as well.

However, some tasks that are purely open-ended text generation cannot be evaluated in the same way as for categorization. Different metrics are required for the specific purpose of the evaluation. Code generation is a very different case in open-ended generative evaluations. The generated code must pass the test suite but on the other hand, it is also important to understand if a model is capable of generating different solutions as a code, what is the probability of selecting the correct one among them. Pass@k is a very good metric in this case. It works in this manner that given a problem, different solutions as code are generated. They are tested for correctness using different functionality tests. Afterward, from generated n solutions, and the respective c number of them being correct equation \ref{eq:passk} provides the final value.

\begin{equation}\label{eq:passk}
    \text{pass@$k$} := \mathop{\mathbb{E}}_{\text{Problems}} \left[ 1 - \frac{{\binom{n-c}{k}}} {\binom{n}{k}} \right]
\end{equation}

Exact match (EM) is another metric that is mostly concerned with exact matches from (pre-defined) answers. It counts a prediction as correct if it exactly matches one of more than one desired reference text token by token. In some cases, it can be the same as accuracy and the equation \ref{eq:em} shows the mathematical definition. Here M is total number of correct answers and N is the total number of questions \cite{bai2021more}.

\begin{equation}\label{eq:em}
    EM = \frac{M}{N}
\end{equation}

Human equivalence score (HEQ) on the other hand, is an alternative to F1 score \cite{huang2018flowqa}. HEQ-Q represents the precision of individual questions, wherein an answer is deemed correct if the model's F1 score surpasses the average human F1 score. Likewise, HEQ-D denotes the precision of each dialogue; it is deemed accurate when all questions within the dialogue meet the criteria of HEQ \cite{choi-etal-2018-quac}.

\begin{table*}[ht]
\centering
\caption{LLM categories and respective definitions.} \label{table:llm_categories}
\begin{tabular}{lll}
\hline
\multicolumn{1}{|l|}{Classification}        & \multicolumn{1}{l|}{Category} & \multicolumn{1}{l|}{Description} \\ \hline
\multicolumn{1}{|l|}{\multirow{4}{*}{Size}} & \multicolumn{1}{l|}{Small}    & \multicolumn{1}{l|}{Number of parameters $\leq$ 1B}            \\ \cline{2-3} 
\multicolumn{1}{|l|}{}                      & \multicolumn{1}{l|}{Medium}   & \multicolumn{1}{l|}{1B $<$ Number of parameters $\leq$ 10B}            \\ \cline{2-3} 
\multicolumn{1}{|l|}{}                      & \multicolumn{1}{l|}{Large}    & \multicolumn{1}{l|}{10B $<$ Number of parameters $\leq$ 100B}            \\ \cline{2-3} 
\multicolumn{1}{|l|}{}                      & \multicolumn{1}{l|}{Very Large} & \multicolumn{1}{l|}{100B $<$ Number of parameters}            \\ \hline
\multicolumn{1}{|l|}{\multirow{3}{*}{Type}} & \multicolumn{1}{l|}{Foundation model}    & \multicolumn{1}{l|}{Pretrained language model}            \\ \cline{2-3} 
\multicolumn{1}{|l|}{}                      & \multicolumn{1}{l|}{Instruction model}   & \multicolumn{1}{l|}{Pretrained and instruction fine-tuned language model}            \\ \cline{2-3} 
\multicolumn{1}{|l|}{}                      & \multicolumn{1}{l|}{Chat model}    & \multicolumn{1}{l|}{Pretrained, instruction fine-tuned, and chat fine-tuned language model}            \\ \hline
\multicolumn{1}{|l|}{\multirow{2}{*}{Origin}} & \multicolumn{1}{l|}{Original model}    & \multicolumn{1}{l|}{An original model released with either Foundation, Instruction, or Chat model}            \\ \cline{2-3}  
\multicolumn{1}{|l|}{}                      & \multicolumn{1}{l|}{Tuned model}    & \multicolumn{1}{l|}{Fine-tuned version of an original model}            \\ \hline
\multicolumn{1}{|l|}{\multirow{2}{*}{Availability}} & \multicolumn{1}{l|}{Publicly available}    & \multicolumn{1}{l|}{Model and weights are available due to request to without request}            \\ \cline{2-3}  
\multicolumn{1}{|l|}{}                      & \multicolumn{1}{l|}{Publicly unavailable}    & \multicolumn{1}{l|}{Model and weights are not publicly available}            \\
\hline
\end{tabular}
\end{table*}

\begin{table*}[!ht]
    \centering
    \caption{Different LLM categorization.}
    \label{tab:categorized}
    \begin{tabular}{|l|l|l|l|l|l|}
    \hline
        Model & Size & \#Params (B)& Type & Availability & Origin \\ \hline
        Davinci-002 & \cellcolor{red!25}Very Large & 175 & \cellcolor{blue!25}Instruction &  \cellcolor{yellow!25}Unavailable & \cellcolor{pink!25}Tuned \\ \hline
        Davinci-003 & \cellcolor{red!25}Very Large & 175 &\cellcolor{blue!25}Instruction &  \cellcolor{yellow!25}Unavailable & \cellcolor{pink!25}Tuned \\ \hline
        GPT 3.5-turbo & \cellcolor{yellow!25}Large & 20 &\cellcolor{purple!25}Chat &  \cellcolor{yellow!25}Unavailable & \cellcolor{pink!25}Tuned \\ \hline
        Falcon 7B & \cellcolor{green!25}Medium & 7 &\cellcolor{gray!25}Foundation & \cellcolor{green!25}Public & \cellcolor{teal!25}Original \\ \hline
        Alpaca & \cellcolor{yellow!25}Large & 13 &\cellcolor{purple!25}Chat & \cellcolor{green!25}Public & \cellcolor{pink!25}Tuned \\ \hline
        Pythia 7B & \cellcolor{green!25}Medium & 7 & \cellcolor{gray!25}Foundation & \cellcolor{green!25}Public & \cellcolor{teal!25}Original \\ \hline
        Pythia 12B & \cellcolor{yellow!25}Large & 12 &\cellcolor{gray!25}Foundation & \cellcolor{green!25}Public & \cellcolor{teal!25}Original \\ \hline
        LLAMA 7B & \cellcolor{green!25}Medium & 7 &\cellcolor{purple!25}Chat & \cellcolor{green!25}Public & \cellcolor{teal!25}Original \\ \hline
        LLAMA 2 7B & \cellcolor{green!25}Medium & 7 &\cellcolor{purple!25}Chat & \cellcolor{green!25}Public & \cellcolor{pink!25}Tuned \\ \hline
        LLAMA 2 7B & \cellcolor{green!25}Medium & 7 &\cellcolor{gray!25}Foundation & \cellcolor{green!25}Public &  \cellcolor{teal!25}Original\\ \hline
        Vicuna 13B & \cellcolor{yellow!25}Large & 13 & \cellcolor{gray!25}Foundation & \cellcolor{green!25}Public & \cellcolor{pink!25}Tuned \\ \hline
        Vicuna 7B & \cellcolor{green!25}Medium & 7 &\cellcolor{gray!25}Foundation & \cellcolor{green!25}Public & \cellcolor{pink!25}Tuned \\ \hline
        Claude & \cellcolor{yellow!25}Large & 93 &\cellcolor{purple!25}Chat &  \cellcolor{yellow!25}Unavailable & \cellcolor{teal!25}Original \\ \hline
        Claude 2 & \cellcolor{red!25}Very Large & 137 & \cellcolor{purple!25}Chat &  \cellcolor{yellow!25}Unavailable & \cellcolor{teal!25}Original \\ \hline
    \end{tabular}
\end{table*}

Evaluation of other generative tasks such as machine translation are based on metrics such as Rouge and BLEU. These scores work well when there is a reference text as ground truth (such as translation) and a hypothesis that is generated by the generative model, in our case the LLM. These scores are mostly used for cases where the goal is to detect the similarity of the answer and ground truth in a computation manner. In a computation manner, it meant that nothing more than N-Grams would be used. However, metrics such as BERT-Score are also good for these cases but they are also heavily erroneous because another model is used to judge. Still, even today, evaluating purely generated content is very hard and no completely fitting metric is not found, metrics are either looking for simplistic features such as N-Gram, SkipGram, etc, or they are models with unknown accuracy and preciseness \cite{lee2023survey}.

Generative evaluation metrics are also another type of evaluation metric for LLMs that use another LLM for evaluating the answer. However, depending on the task itself, evaluation can be possible in this way or not. Another dependency that makes generative evaluation error-prone is reliance on the prompt itself. \hyperlink{https://docs.ragas.io/en/stable/}{RAGAS} is one of the good examples that incorporate the usage of generative evaluation. 

Various benchmarks and leaderboards have been proposed to address the most challenging question in the world of large language models: Which one is better? However not a simple answer can address this question. The answer depends on various aspects of large language models. Section \ref{sec:llm_datasets} shows the categorical presentation of different tasks and the most important datasets in each category. We will follow the same categorization and provide a comparison based on each category. After providing comparison for each category, we will provide a broad overview of aggregated performance  by averaging the reported performance metric on different tasks.

\begin{figure*}[h]
    \centering
    \includegraphics[scale=0.95]{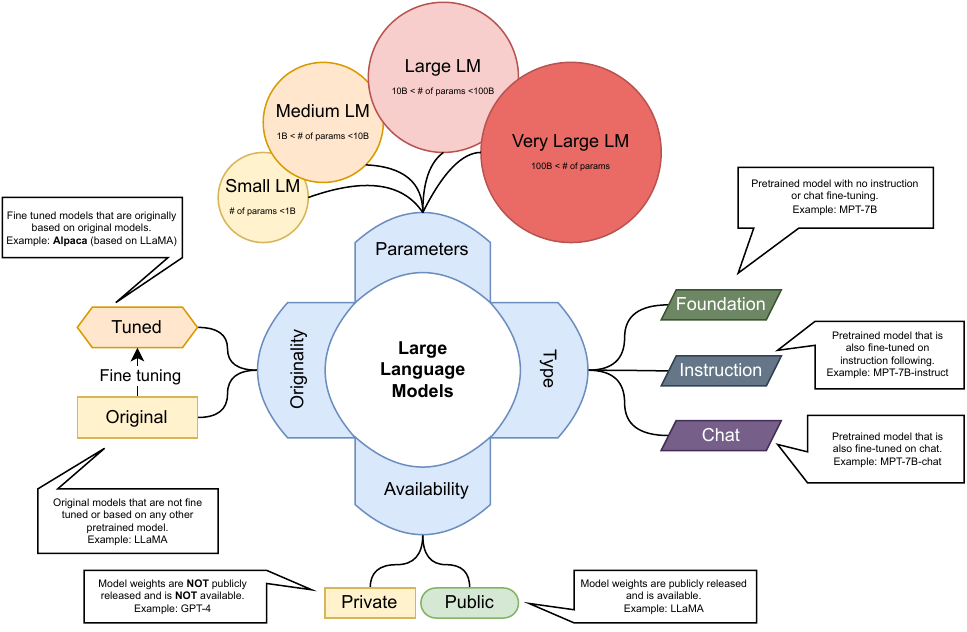}
    \caption{LLM categorizations.}
    \label{fig:llm_types}
\end{figure*}

Evaluating different LLMs can be seen also from different perspectives. For example, a LLM with a drastically fewer number of parameters is not completely comparable to one with a larger number of parameters. From this perspective, we will categorize LLMs in four categories as well: \textbf{small} (less than or equal to 1 billion parameters), \textbf{medium} (between 1 and 10 billion), \textbf{large} (between 10 and 100 billion), and \textbf{very large} (more than 100 billion). Another classification for the LLMs we use is their primary use case. We consider each LLM to be either: \textbf{Foundation} model (pretrained language model with no instruction fine-tuning and chat fine-tuning), \textbf{Instruction} model (pretrained language model with only instruction fine-tuning), and \textbf{Chat} model (pretrained language model with instruction and chat fine-tuning). Apart from all the categorization described, another category is required to distinguish between original models and tuned ones. \textbf{Original} models are those that have been released as a foundation model or a fine-tuned one. \textbf{Tuned} models are those that grasped the original model and tuned it with different datasets or even different training approaches. It is also good to note that original models are usually foundation models that have been fine-tuned on specific datasets or even different approaches. Availability of the model weights regardless of the license is another category in our classification. Models that have their weights publicly available (even through request) are noted as \textbf{Public} models while others are noted as \textbf{Private}. Table \ref{table:llm_categories} shows all of these definitions and abbreviations used in the rest of the article. 
Figure \ref{fig:llm_types} illustrate these visually.

According to the provided categorizations, we can categorize and label each notable LLM as shown in table \ref{tab:categorized}. As can be seen from this table, models categorized as very large are also unavailable as well.

\subsection{LLMs' Performance on Different Tasks}
Commonsense reasoning is one of the important capabilities each model can obtain. This capability denotes the ability of the model to use prior knowledge in combination with reasoning skills. In the case of HellaSwag for example, finding the continuation of text is challenging because the given text contains a partial part of the story while the given choices as continuation are tricky to select, and without having prior knowledge about the world it is not possible. This specific kind of reasoning deserves high attention because it is related to utilizing previous knowledge with open text-described scenes or facts. As can be seen from table \ref{table:commonsense_reasoning} not just Unavailable models but also Public ones can achieve good results on various tests.

\begin{table}[H]
    \centering
    \caption{Commonsense reasoning comparison.} \label{table:commonsense_reasoning}
    \begin{tabular}{|l|l|l|l|l|}
    \hline
        Model            & OBQA & HellaSwag \\ \hline
Davinci-003      & \textbf{51}   & 83.4      \\ \hline
Falcon 7B        & 44.4 & 76.3      \\ \hline
Alpaca           & 43.4 & 73.9      \\ \hline
Pythia 7B        & 37.2 & 64        \\ \hline
Pythia 12B       & 43.2 & 68.1      \\ \hline
LLAMA 7B         & 42.4 & 73        \\ \hline
Dolly 6B         & 41.2 & 67.6      \\ \hline
Dolly 12B        & 40.4 & 71        \\ \hline
Alpaca 7B        & 43.4 & 73.9      \\ \hline
Alpaca Lora 7B   & 42.6 & 74        \\ \hline
GPT-J 6.7B       & 38.2 & 66.2      \\ \hline
LLama 7B         & 42.4 & 73        \\ \hline
LLama 13B        & 42.2 & 76.2      \\ \hline
Pythia 6.7B      & 37.2 & 64        \\ \hline
Pythia 12B       & 38   & 67.3      \\ \hline
StableLM Tuned   & 33.4 & 53.6      \\ \hline
Koala 13B        & 42.8 & 72.6      \\ \hline
Mosaic mpt-7B    & 42.6 & 76.3      \\ \hline \hline
LLAMA 2 70B      & -    & 87.33     \\ \hline
LLAMA 65B        & -    & 86.09     \\ \hline
Falcon 40B       & -    & 85.3      \\ \hline
Falcon 180B      & -    & 88.86     \\ \hline
MPT Instruct 30B & -    & 84.31     \\ \hline
MPT Instruct 7B  & -    & 77.91     \\ \hline
Yi 6B            & -    & 76.42     \\ \hline
Yi 34B           & -    & 85.69     \\ \hline
GPT-4             & -    & \textbf{95.3}      \\ \hline
Gemini Ultra     & -    & 87.8      \\ \hline
    \end{tabular}
\end{table}

From the results presented in Table \ref{table:commonsense_reasoning} it is clear that GPT-4 achieves best results for HellaSwag while Davinci-003 is best model for OBQA. It is also good to note that results for OBQA are not reported for all of the models and possibly davinci-003 is not the best model achieving highest results on OBQA.

Not all models report their performance on all datasets, and because of that, the number of models for which performance is reported in different tables varies.

\begin{table}[!ht]
    \centering
    \caption{Symbolic reasoning comparison.}
    \begin{tabular}{|l|l|l|}
    \hline
        Model & Cobjects & Penguins \\ \hline
        GPT-NeoX & 26 & 33.56\\ \hline
OPT 66B & 31.2 & 28.08\\ \hline
Bloomberg GPT & 34.8 & 37.67\\ \hline
BLOOM 176B & 36.8 & 40.41\\ \hline
PaLM 540B & 38 & 44.5\\ \hline
Gopher-280B & 49.2 & 40.6\\ \hline
Chinchilla-70B & 59.7 & 48.7\\ \hline
PaLM 2 & 61.2 & 65.8\\ \hline
    \end{tabular}
\end{table}



World knowledge is mostly about general knowledge questions, for example, in Wikifact dataset questions such as "Who is the author of a specific well-known book" can be found and references are also provided. Table \ref{tab:world_knowledge} shows the results.

\begin{table}[H]
    \centering
    \caption{World knowledge comparison.}\label{tab:world_knowledge}
    \begin{tabular}{|l|l|l|l|l|}
    \hline
        Model & TriviaQA & NaturalQ & WebQ & ARC \\ \hline
BLOOM & - & - & - & 32.9\\ \hline
BLOOM 176B & - & - & - & 50.85\\ \hline
Bloomberg GPT & - & - & - & 48.63\\ \hline
Chinchilla & - & 35.5 & - & -\\ \hline
Codex + REPLUG & 76.8 & 44.7 & - & -\\ \hline
GAL 120B & - & - & - & 67.9\\ \hline
GLaM 62B/64E & 75.8 & 32.5 & 15.5 & 50.3\\ \hline
Gopher & - & 28.2 & - & -\\ \hline
GPT-3 175B & 71.2 & 29.9 & 41.5 & 85.2\\ \hline
GPT-4 & - & - & - & 96.4\\ \hline
GPT-NeoX & - & - & - & 45.39\\ \hline
LLaMA 13B & - & - & - & 52.7\\ \hline
LLaMA 2 70B & 85 & 33 & - & -\\ \hline
LLaMA 33B & - & 24.9 & - & 57.8\\ \hline
LLaMA 65B & 72.6 & 39.9 & - & -\\ \hline
LLaMA 7B & - & - & - & 47.6\\ \hline
Mistral 7B & 69.9 & 28.8 & - & 55.5\\ \hline
Neo-6B & - & 13.7 & - & -\\ \hline
OPT & - & - & - & 31.1\\ \hline
OPT 66B & - & - & - & 44.54\\ \hline
OPT-175B & - & - & - & 43.94\\ \hline
OPT-175B & - & - & - & 25.6\\ \hline
PaLM 2-L & 86.1 & 37.5 & 28.2 & 95.1\\ \hline
PaLM 2-M & 81.7 & 32 & 26.9 & 64.9\\ \hline
PaLM 2-S & 75.2 & 25.3 & 21.8 & 59.6\\ \hline
PaLM-540B & 81.4 & 39.6 & 43.5 & 87.1\\ \hline
phi-1.5-web 1.3B & - & - & - & 44.9\\ \hline
SparseGPT & - & - & - & 38.99\\ \hline
SparseGPT & - & - & - & 39.85\\ \hline
SparseGPT & - & - & - & 41.3\\ \hline
    \end{tabular}
\end{table}

For some specific use-case models, it is highly demanded to have coding and code-generation capability. Table \ref{tab:coding} shows the results of different models on coding capability.

\begin{table}[H]
    \centering
    \caption{Coding capability comparison.} \label{tab:coding}
    \begin{tabular}{|l|l|}
    \hline
        Model & HumanEval \\ \hline
Gemini Ultra & 74.4 \\ \hline
Gemini Pro & 67.7 \\ \hline
GPT-4 & 67 \\ \hline
WizardCoder 15B & 57.3 \\ \hline
phi-1 1.3B & 50.6 \\ \hline
Code Llama & 48.8 \\ \hline
GPT-3.5 & 48.1 \\ \hline
OctoCoder & 46.2 \\ \hline
phi-1-small & 45 \\ \hline
PaLM 2-S & 37.6 \\ \hline
InstructCodeT5+ 16B & 35 \\ \hline
Mistral 7B & 30.5 \\ \hline
LLaMA 2 & 29.9 \\ \hline
phi-1-base & 29 \\ \hline
Codex-12B & 28.81 \\ \hline
PaLM 540B & 26.2 \\ \hline
CodeT5+ 2B & 24.2 \\ \hline
LLaMA 65B & 23.7 \\ \hline
LLaMA 33B & 21.7 \\ \hline
PaLM 62B & 15.9 \\ \hline
LLaMA 13B & 15.8 \\ \hline
LaMDA 137B & 14 \\ \hline
MIM-350M & 13.7 \\ \hline
LLaMA 7B & 10.5 \\ \hline
PaLM 8B & 3.6 \\ \hline
    \end{tabular}
\end{table}

Arithmetic reasoning is another challenging reasoning capability to achieve. GSM8K for example contains grade school mathematical questions with respect to their answers. Table \ref{tab:arithmetic} provides an insight for different model comparisons.

\begin{table}[H]
    \centering
    \caption{Arithmetic reasoning comparison.} \label{tab:arithmetic}
    \begin{tabular}{|l|l|l|}
    \hline
        Model & GSM8k & MATH \\ \hline
Gemini Ultra & 94.4 & 53.2 \\ \hline
GPT-4 & 87.1 & 42.5 \\ \hline
Gemini Pro & 86.5 & 32.6 \\ \hline
ToRA 70B & 84.3 & 49.7 \\ \hline
MathCoder-L-70B & 83.9 & - \\ \hline
MetaMath 70B & 82.3 & 26 \\ \hline
MuggleMATH 70B & 82.3 & - \\ \hline
MathCoder-CL-34B & 81.7 & 45.2 \\ \hline
ToRA-Code 34B & 80.7 & 50.8 \\ \hline
MetaMath-Mistral-7B & 77.7 & - \\ \hline
Arithmo2-Mistral-7B & 76.4 & - \\ \hline
ToRA-Code 13B & 75.8 & 48.1 \\ \hline
Arithmo-Mistral-7B & 74.7 & - \\ \hline
MathCoder-CL-13B & 74.1 & 35.9 \\ \hline
MuggleMATH 13B & 74 & - \\ \hline
CodeT5+ & 73.8 & - \\ \hline
KwaiYiiMath 13B & 73.3 & - \\ \hline
ToRA-Code 7B & 72.6 & 44.6 \\ \hline
MathCoder-L-13B & 72.6 & 29.9 \\ \hline
MetaMath 13B & 71 & 22.5 \\ \hline
LLaMA 65B & 69.7 & 10.6 \\ \hline
MuggleMATH 7B & 68.4 & - \\ \hline
MathCoder-CL-7B & 67.8 & 23.3 \\ \hline
MetaMath 7B & 66.4 & 19.4 \\ \hline
RFT 70B & 64.8 & - \\ \hline
MathCoder-L-7B & 64.2 & - \\ \hline
Orca 2-13B & 59.14 & - \\ \hline
U-PaLM & 58.5 & - \\ \hline
PaLM-540B & 58.1 & 8.8 \\ \hline
LLaMA 2 70B & 56.8 & - \\ \hline
RFT 13B & 55.3 & - \\ \hline
LLaMA 33B & 53.1 & 7.1 \\ \hline
Mistral 7B & 52.2 & 13.1 \\ \hline
RFT 7B & 51.2 & - \\ \hline
LLaMA 65B & 50.9 & 20.5 \\ \hline
Orca 2-7B & 47.23 & - \\ \hline
Text-davinci-002 & 40.7 & 19.1 \\ \hline
LLaMA 33B & 35.6 & 3.9 \\ \hline
GPT-Neo-2.7B & 19.5 & - \\ \hline
LLaMA 7B & 18.1 & 2.9 \\ \hline
PaLM 540B & 17.9 & 8.8 \\ \hline
LLaMA 13B & 17.8 & 3.9 \\ \hline
LLaMA 7B & 11 & 2.9 \\ \hline
GPT-Neo-125M & 7.5 & - \\ \hline
PaLM 8B & 4.1 & 1.5 \\ \hline
GPT-2 & - & 5.4 \\ \hline
GPT-3 175B & - & 5.2 \\ \hline
PaLM 62B & - & 4.4 \\ \hline
GPT-3-13B & - & 3 \\ \hline
LLaMA 7B & 11 & 2.9 \\ \hline
PaLM 8B & - & 1.5 \\ \hline
    \end{tabular}
\end{table}

Large language models in some cases are hallucinating answers simply because they are next-token prediction machines. Hallucination is one of the important factors in measuring how much a large language model is trustworthy and reliable. Measuring hallucination on the other hand is also not easy as it seems because each fact can be written in different styles and even the smallest changes in writing make it hard to detect. It is fair to assume if any particular LLM is more capable to detect hallucination of false information in text, it is also more trustworthy. HaluEval is one of the datasets that aims to measure hallucination in this field \cite{li2023halueval}. Evaluation can also be performed by another model judging the response with regard to the actual answer \cite{Simon2024}. Table \ref{tab:hallucination} shows the evaluation of different models based on these datasets.

\begin{table*}[ht]
    \centering
    \caption{Hallucination evaluation} \label{tab:hallucination}
    \begin{tabular}{|l|l|l|l|l|l|}
    \hline
        Model & HHEM & HaluEval QA & HaluEval Dialogue & HaluEval Sum. & HaluEval General \\ \hline
GPT 4 & 97 & - & - & - & - \\ \hline
GPT 4 Turbo & 97 & - & - & - & - \\ \hline
GPT 3.5 Turbo & 96.5 & 62.59 & 72.4 & 58.53 & 79.44 \\ \hline
Davinci002 & - & 60.05 & 60.81 & 47.77 & 80.42 \\ \hline
Davinci003 & - & 49.65 & 68.37 & 48.07 & 80.4 \\ \hline
GPT-3 & - & 49.21 & 50.02 & 51.23 & 72.72 \\ \hline
Google Gemini Pro & 95.2 & - & - & - & - \\ \hline
Llama 2 70B & 94.9 & - & - & - & - \\ \hline
Llama 2 7B & 94.4 & 49.6 & 43.99 & 49.55 & 20.46 \\ \hline
Llama 2 13B & 94.1 & - & - & - & - \\ \hline
Cohere-Chat & 92.5 & - & - & - & - \\ \hline
Cohere & 91.5 & - & - & - & - \\ \hline
Claude 2 & 91.5 & 69.78 & 64.73 & 57.75 & 75 \\ \hline
Claude 1 & & 67.6 & 64.83 & 53.76 & 73.88 \\ \hline
Microsoft Phi 2 & 91.5 & - & - & - & - \\ \hline
Google Palm 2 (beta) & 91.4 & - & - & - & - \\ \hline
Mixtral 8x7B & 90.7 & - & - & - & - \\ \hline
Amazon Titan Express & 90.6 & - & - & - & - \\ \hline
Mistral 7B & 90.6 & - & - & - & - \\ \hline
Google Palm 2 Chat (beta) & 90 & - & - & - & - \\ \hline
Google Palm 2 & 87.9 & - & - & - & - \\ \hline
Google Palm 2 Chat & 72.8 & - & - & - & - \\ \hline
ChatGLM & - & 47.93 & 44.41 & 48.57 & 30.92 \\ \hline
Falcon & - & 39.66 & 29.08 & 42.71 & 18.98 \\ \hline
Vicuna & - & 60.34 & 46.35 & 45.62 & 19.48 \\ \hline
Alpaca & - & 6.68 & 17.55 & 20.63 & 9.54 \\ \hline
    \end{tabular}
\end{table*}

\section{Challenges and Future Directions}
\label{sec:LLM_challenges}
As we have seen in the previous sections, large language models have achieved impressive results in the past 1-2 years. At the same time this is still a new and extremely active research area where the pace of innovation is increasing rather than slowing down. As in any other evolving area though, there are still numerous challenges ahead. 
Here we briefly mention some of the challenges and main active areas which are known so far.
It is worth noting that LLM challenges are discussed in details in a work by Kaddour et al. \cite{kaddour2023challenges}.

\subsection{Smaller and more efficient Language Models}

This is a survey on \emph{large} language models, and there has been an initial push towards "larger is better" that has clearly been rewarded with ever larger models like GPT-4 getting better accuracy and performance in benchmarks. However, those large models are costly and inefficient in several dimensions (e.g. high latency). In response to all of this, there is a current research trend to come up with Small Language Models (SLMs) as a cost-effective alternative to LLMs, particularly when used on specific tasks that might not require the full generality of larger models.
Prominent works in this direction include {Phi-1} \cite{gunasekar2023textbooks}, {Phi-1.5} \cite{li2023textbooks}, and {Phi-2} from Microsoft.

More generally, we should expect many research efforts in this area of how to train smaller and more efficient models. Techniques such as parameter-efficient fine-tuning (PEFT), teacher/student, and other forms of distillation -- see section \ref{sub:cost-effective} -- will continue to be used to build a smaller model out of larger ones.

\subsection{New Post-attention Architectural Paradigms}
Transformer blocks have been a crucial and constant part of most of current LLM frameworks, and it’s a big question mark how much longer this architecture will be in vogue, and what will be the next big architectural break-through in the field of deep learning (and NLP).
Since AlexNet in 2012, we have seen many architectures go in and out of fashion, including LSTM, GRU, seq2seq, but Transformers have been the dominant approach since its inception. As described earlier, attention is the main mechanism driving transformers. More recently, there has been promising research in alternative approaches that are being labelled as post-attention.

An important class of such class of post-attention models are the so called State Space Models (SSMs). While the notion of State Space Models has a long history in machine learning, it should be noted that in the context of language models, SSM is usually used in reference to the newer Structure State Space Model architecture or S4 for short (see Gu et al. \cite{gu2022S4}). Some recent models in this category are Mamba \cite{gu2023mamba}, Hyena \cite{poli2023hyena}, and Striped Hyena \cite{stripedhyena}.

While all of those models are very competitive in terms of performance in leaderboards and efficiency, they also address an important challenge in more traditional attention-based architectures: \emph{the lack of support for larger context windows}.

Having a good answer to many prompts requires context. For example, the response to "Recommend some good movies for me" requires a lot of context about "me" as well as what movies are available and which ones I have not watched. Context length is especially important for RAG, where large portions of text might be retrieved and injected into the prompt for generation (see section \ref{sub:RAG}.

The longer the context length, the more tokens we can squeeze into the context. The more information the model has access to, the better its response will be. But on the other hand, with very long context, it would be hard for the model to remember everything and efficiently process all the information. Attention-based models are highly inefficient for longer contexts and that is why we should expect more research in different mechanisms that enable processing longer contexts and generally come up with more efficient architectures.

That being said, new architectures might not only propose alternatives for the attention mechanism but rather rethink the whole Transformer architecture. As an early example of this, Monarch Mixer \cite{fu2023monarch} proposes a new architecture that uses the same sub-quadratic primitive that achieves high hardware efficiency on GPUs -- Monarch matrices -- along both sequence length and model dimension.

On the other end of the spectrum, it is worth mentioning that there are some attention-compatible architectural mechanisms that have been recently gaining steam and proving their value in creating better and more powerful LLMs. Probably the best example of such mechanism is Mixture of Experts (MoE). MoEs have been around in machine learning for years, even before the Deep Learning Era \cite{mclachlan2019finite}, but they have been gaining popularity since then, and particularly in the context of Transformer models and LLMs. 

In LLMs, MoEs allow to train an extremely large model than is then only partially instantiated during inference when some of the experts are turned off wherever the gating/weighting function has a low weight assigned to them. As an example, the GLaM model has 1.2 trillion parameters, but during inference only 2 out of the 64 experts are used \cite{du2022glam}.

MoEs are nowadays an important component of the so-called frontier LLMs (i.e. the most advanced and capable models). GPT-4 itself is rumored to be based on a MoE architecture, and some of the best performing LLMs such as Mixtral \cite{mixtral}, are basically an MoE version of pre-existing LLMs. 

Finally, it is important to note that MoEs can be used as a component of any architecture regardless of whether it is based on attention or not. In fact, MoEs have also been applied to SSM-based LLMs like Mamba\ cite{pioro2024moemamba}. We should continue to see MoE-driven improvements in the future regardless of the underlying architecture.

\subsection{Multi-modal Models}
Future LLMs are expected to be multi-modal and handle a variety of data types, such as text, images, and videos, audio, in a unified manner. 
This opens up possibilities for more diverse applications in fields like question answering, content generation, creative arts, and healthcare, robotics, and beyond.
There are already several prominent multi-modal LLMs out there, including: LLAVA 
\cite{liu2023visual}, LLAVA-Plus \cite{liu2023llava}, GPT-4 \cite{gpt4}, Qwen-vl \cite{bai2023qwen}, Next-GPT \cite{wu2023next}, but the trend is expected to be continued. Evaluation of these models also is a new research topic, especially conversational generative vision models \cite{khasmakhi2023convgenvismo}.
Multi-modal LLMs can unlock huge potentials in a variety of tasks, and there has already been a descent progress in this direction, which needs a dedicated paper to discuss all its details.

\subsection{Improved LLM Usage and Augmentation techniques}

As we described in section\ref{sec:LLM_used}, many of the shortcomings and limitations of LLMs such as \emph{hallucination} can be addressed through advanced prompt engineering, use of tools, or other augmentation techniques. We should expect not only continued, but accelerated research in this area. 
It is worth mentioning that, in the specific case of software engineering, some works (\cite{alshahwan2024automated}) tried to automatically  eliminate this issue from the overall software engineering workflow

LLM-based systems are already starting to replace machine learning systems that were until recently using other approaches. As a clear example of this, LLMs are now being deployed to better understand people preference and interests, and provide more personalized interactions, whether in customer service, content recommendation, or other applications. This involves better understanding of user preferences, and analyzing their past interactions and using them as the context. We will continue to see research in the application and usage of LLMs for not only \emph{personalization and recommendations}, but many other application areas using other machine learning techniques.

Finally, another important area of research we expect to gather increased attention is that of \emph{LLM-based agents and multi-agent systems} \cite{xi2023rise,wang2023survey,durante2024agent}. The development of LLM systems with access to external tools and decision-making capabilities is both exciting and challenging. We will see continued research and progress in this important area that some argue could lead to Artificial General Intelligence (AGI).

\subsection{Security and Ethical/Responsible AI}
Ensuring the robustness and security of LLMs against adversarial attacks and other vulnerabilities is a critical area of research \cite{sun2024trustllm}. As LLMs are increasingly deployed in real-world applications, they need to be protected from potential threats, to prevent them being used to manipulate people or spread mis-information.
Improving the reasoning capabilities of these model \cite{josifoski2023flows}, would help them to better detect potential adversarial attacks.

Addressing ethical concerns and biases in LLMs is another active area of research. Efforts are being made to ensure that LLMs are fair, unbiased, and capable of handling sensitive information responsibly. As LLMs are being used more and more by a large number of people on a daily basis, making sure they are unbiased and behave responsibly is crucial.

\section{Conclusion}
This paper present a survey of LLMs developed in the past few years.
We first provide an overview of early pre-trained language models (e.g., as BERT), then review three popular LLM families (GPT, LLaMA, PaLM), and other representative LLMs. 
We then survey methods and techniques of building, augmenting, and using LLMs.
We review popular LLM datasets and benchmarks, and compare performance of a set of prominent models on public benchmarks.
Finally, we present open challenges and future research directions.

\bibliographystyle{IEEEtran}
\bibliography{ref}

\appendix
\textbf{1. Open Source Toolkits For LLM Development and Deployment}
\label{sec:LLM_tools}

There are various frameworks and libraries developed for LLM training, evaluation, and deployment, and covering every single framework is out of this paper's scope. But we try to provide a brief introduction of some of the most popular ones, grouped into different categories.

\subsection{LLM Training/Inference Frameworks}
Some of the popular frameworks which are useful for LLM training includes (note that some of them can be used beyond LLM training too):

\textbf{DeepSpeed} 
\cite{deepspeed} is a deep learning optimization library that makes distributed training and inference easy, efficient, and effective.
DeepSpeed enables world's most powerful language models like MT-530B and BLOOM. It is an easy-to-use deep learning optimization software suite that powers unprecedented scale and speed for both training and inference. With DeepSpeed you can:

\textbf{Transformers}
\cite{transformers} is library by HuggingFace which provides thousands of pretrained models to perform tasks on different modalities such as text, vision, and audio. Using pretrained models one can reduce compute costs, carbon footprint, and save the time and resources required to train a model from scratch.

\textbf{Megatron-LM}
\cite{MegatronLM} is a large, powerful transformer developed by the Applied Deep Learning Research team at NVIDIA.
It contains efficient, model-parallel (tensor, sequence, and pipeline), and multi-node pre-training of transformer based models such as GPT, BERT, and T5 using mixed precision.

\textbf{BMTrain}
\cite{BMTrain} is an efficient large model training toolkit that can be used to train large models with tens of billions of parameters. It can train models in a distributed manner while keeping the code as simple as stand-alone training.

\textbf{GPT-NeoX}
\cite{gpt-neox} leverages many of the same features and technologies as the popular Megatron-DeepSpeed library but with substantially increased usability and novel optimizations.

\textbf{LoRA}
\cite{lora} library provides the support for Low-Rank Adaptation of Large Language Models. It reduces the number of trainable parameters by learning pairs of rank-decompostion matrices while freezing the original weights. This vastly reduces the storage requirement for large language models adapted to specific tasks and enables efficient task-switching during deployment all without introducing inference latency. LoRA also outperforms several other adaptation methods including adapter, prefix-tuning, and fine-tuning.

\textbf{ColossalAI}
library \cite{ColossalAI} provides a collection of parallel components. It aims to support developers to write their distributed deep learning models just like how they write their model on their laptop. They provide user-friendly tools to kickstart distributed training and inference in a few lines.
In terms of Parallelism strategies, they support: Data Parallelism, Pipeline Parallelism, Sequence Parallelism, Zero Redundancy Optimizer (ZeRO) \cite{rajbhandari2020zero}, and Auto-Parallelism.

\subsection{Deployment Tools}
We provide an overview of some of the most popular LLM deployment tools here.

\textbf{FastChat} \cite{FastChat} is an open platform for training, serving, and evaluating large language model based chatbots. 
FastChat's core features include:
The training and evaluation code for state-of-the-art models (e.g., Vicuna, MT-Bench), and a distributed multi-model serving system with web UI and OpenAI-compatible RESTful APIs.

\textbf{Skypilot} \cite{skypilot} is a framework for running LLMs, AI, and batch jobs on any cloud, offering maximum cost savings, highest GPU availability, and managed execution.

\textbf{vLLM} \cite{vllm} is a fast and easy-to-use library for LLM inference and serving. 
vLLM seamlessly supports many Hugging Face models, including the following architectures: Aquila, Baichuan, BLOOM, ChatGLM, DeciLM, Falcon, GPT BigCode, LLaMA, LLaMA 2, Mistral, Mixtral, MPT, OPT, Qwen, Yi, and many more.

\textbf{text-generation-inference} \cite{text-generation-inference} is a toolkit for deploying and serving Large Language Models (LLMs). TGI enables high-performance text generation for the most popular open-source LLMs, including Llama, Falcon, StarCoder, BLOOM, GPT-NeoX, and more. 

\textbf{LangChain} \cite{langchain} is a framework for developing applications powered by language models. It enables applications that:
\begin{itemize}
    \item Are context-aware: connect a language model to sources of context (prompt instructions, few shot examples, content to ground its response in, etc.)
    \item Reason: rely on a language model to reason (about how to answer based on provided context, what actions to take, etc.)
\end{itemize}

\textbf{OpenLLM} \cite{OpenLLM} is an open-source platform designed to facilitate the deployment and operation of large language models (LLMs) in real-world applications. With OpenLLM, you can run inference on any open-source LLM, deploy them on the cloud or on-premises, and build powerful AI applications.

\textbf{Embedchain} \cite{embedchain} is an Open Source RAG Framework that makes it easy to create and deploy AI apps.
Embedchain streamlines the creation of RAG applications, offering a seamless process for managing various types of unstructured data. It efficiently segments data into manageable chunks, generates relevant embeddings, and stores them in a vector database for optimized retrieval.

\textbf{Autogen} \cite{autogen} is a framework that enables the development of LLM applications using multiple agents that can converse with each other to solve tasks. AutoGen agents are customizable, conversable, and seamlessly allow human participation. They can operate in various modes that employ combinations of LLMs, human inputs, and tools.

\textbf{BabyAGI} \cite{babyagi} is an autonomous Artificial Intelligence agent, that is designed to generate and execute tasks based on given objectives. It harnesses cutting-edge technologies from OpenAI, Pinecone, LangChain, and Chroma to automate tasks and achieve specific goals. In this blog post, we will dive into the unique features of BabyAGI and explore how it can streamline task automation.


\subsection{Prompting Libraries}

\textbf{Guidance} \cite{guidance} is a programming paradigm that offers superior control and efficiency compared to conventional prompting and chaining. It allows users to constrain generation (e.g. with regex and CFGs) as well as to interleave control (conditional, loops) and generation seamlessly.

\textbf{PromptTools} \cite{prompttools} offers a set of open-source, self-hostable tools for experimenting with, testing, and evaluating LLMs, vector databases, and prompts. The core idea is to enable developers to evaluate using familiar interfaces like code, notebooks, and a local playground.

\textbf{PromptBench} \cite{promptbench}  is a Pytorch-based Python package for Evaluation of Large Language Models (LLMs). It provides user-friendly APIs for researchers to conduct evaluation on LLMs.

\textbf{Promptfoo} \cite{promptfoo} is a tool for testing and evaluating LLM output quality. It systematically test prompts, models, and RAGs with predefined test cases.

\subsection{VectorDB}
\textbf{Faiss} \cite{faiss} is a library developed by Facebook AI Research that provides efficient similarity search and clustering of dense vectors. It is designed for use with large-scale, high-dimensional data and supports several index types and algorithms for various use cases.

\textbf{Milvus} \cite{milvus} is an open-source vector database built to power embedding similarity search and AI applications. Milvus makes unstructured data search more accessible, and provides a consistent user experience regardless of the deployment environment.

\textbf{Qdrant} \cite{qdrant} is a vector similarity search engine and vector database. It provides a production-ready service with a convenient API to store, search, and manage points—vectors with an additional payload Qdrant is tailored to extended filtering support. environment.

\textbf{Weaviate} \cite{weaviate} is an open-source, GraphQL-based vector search engine that enables similarity search on high-dimensional data. While it is open-source, the commercial version offers additional features, support, and managed services.

Some of the other popular options includes \textbf{LlamaIndex} \cite{llama-index} and \textbf{Pinecone}.

\end{document}